\documentclass[10pt,twocolumn,letterpaper]{article}

\usepackage{iccv}
\usepackage{times}
\usepackage{epsfig}
\usepackage{graphicx}
\usepackage{amsmath}
\usepackage{amssymb}
\usepackage{booktabs}
\usepackage{algpseudocode}
\usepackage{algorithm}
\usepackage{multirow}
\usepackage{multicol}
\usepackage{mathtools}
\newcommand{\comment}[1]{}
\usepackage{xcolor}
\usepackage{ragged2e}

\newcommand{\net}{\mathcal{G}_\theta}

\DeclarePairedDelimiter\norm{\lVert}{\rVert}

\usepackage[pagebackref=true,breaklinks=true,letterpaper=true,colorlinks,bookmarks=false]{hyperref}

\iccvfinalcopy 

\def\iccvPaperID{32} 

\begin{document}

\title{On the unreasonable vulnerability of transformers for image restoration \\-- and an easy fix}




\author{%
  Shashank Agnihotri\\
  Visual Computing, University of Siegen\\
  Germany \\ 
 \and
  Kanchana Vaishnavi Gandikota\\  
  Computer Vision, University of Siegen\\
  Germany \\
  \and
  Julia Grabinski\\
  Fraunhofer ITWM, Kaiserslautern\\
  IMLA, University of Offenburg \\
  Visual Computing, University of Siegen\\
  Germany\\
  \and  
  Paramanand Chandramouli\\  
  Computer Graphics, University of Siegen\\
  Germany\\
  \and
  Margret Keuper\\
  Visual Computing, University of Siegen, and\\
  Max Planck Institute for Informatics, Saarland\\
  Germany\\
}
\maketitle
\ificcvfinal\thispagestyle{empty}\fi

\begin{abstract}
Following their success in  visual recognition tasks, Vision Transformers(ViTs) are being increasingly employed for image restoration. 
As a few recent works claim that ViTs for image classification also have better robustness properties, we investigate whether the improved adversarial robustness of ViTs extends to image restoration. 
We consider the recently proposed Restormer model, as well as NAFNet and the ``Baseline network" which are both simplified versions of a Restormer. 
We use Projected Gradient Descent (PGD) and CosPGD, a recently proposed adversarial attack tailored to pixel-wise prediction tasks for our robustness evaluation. 
Our experiments are  performed on real-world images from the GoPro dataset for image deblurring.
Our analysis indicates that contrary to as advocated by ViTs in image classification works, these models are highly susceptible to adversarial attacks. 
We attempt to improve their robustness through adversarial training. 
While this yields a significant increase in robustness for Restormer, results on other networks are less promising. 
Interestingly, the design choices in NAFNet and Baselines, which were based on \emph{iid} performance, and not on robust generalization, seem to be at odds with the model robustness. 
Thus, we investigate this further and find a fix. 
\end{abstract}

\section{Introduction}
\label{sec:intro}
\begin{figure}[htb]
    \centering 
    \begin{tabular}{@{}c@{\hspace{0.1cm}}c@{}}
    Ground Truth & Restormer \\
    \includegraphics[width=0.49\linewidth]{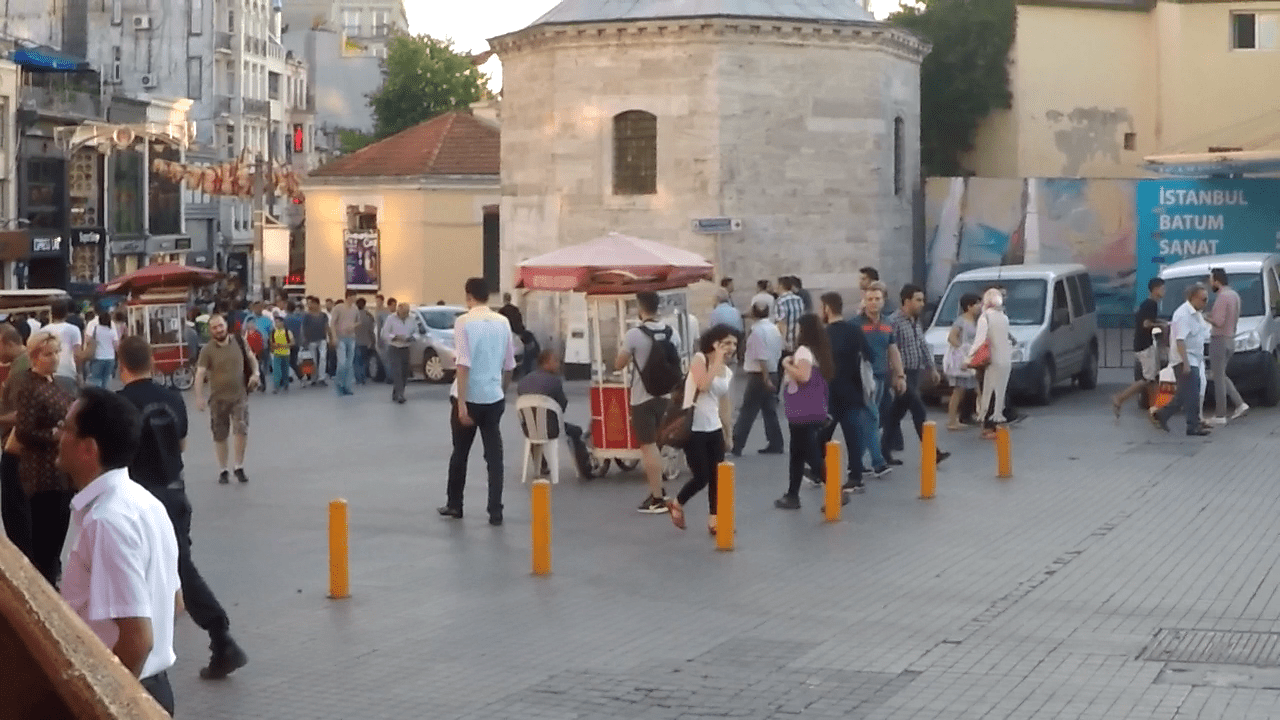} &
    \includegraphics[width=0.49\linewidth]{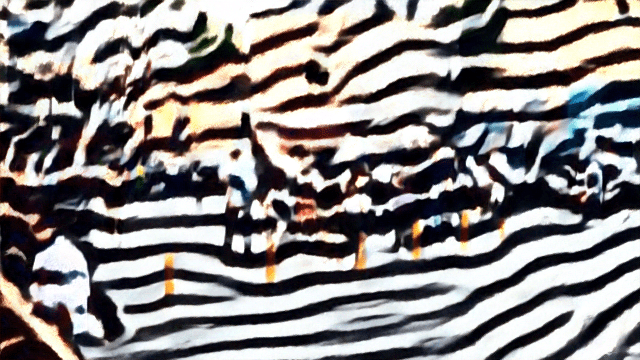} \\
    Baseline network & NAFNet \\
    \includegraphics[width=0.49\linewidth]{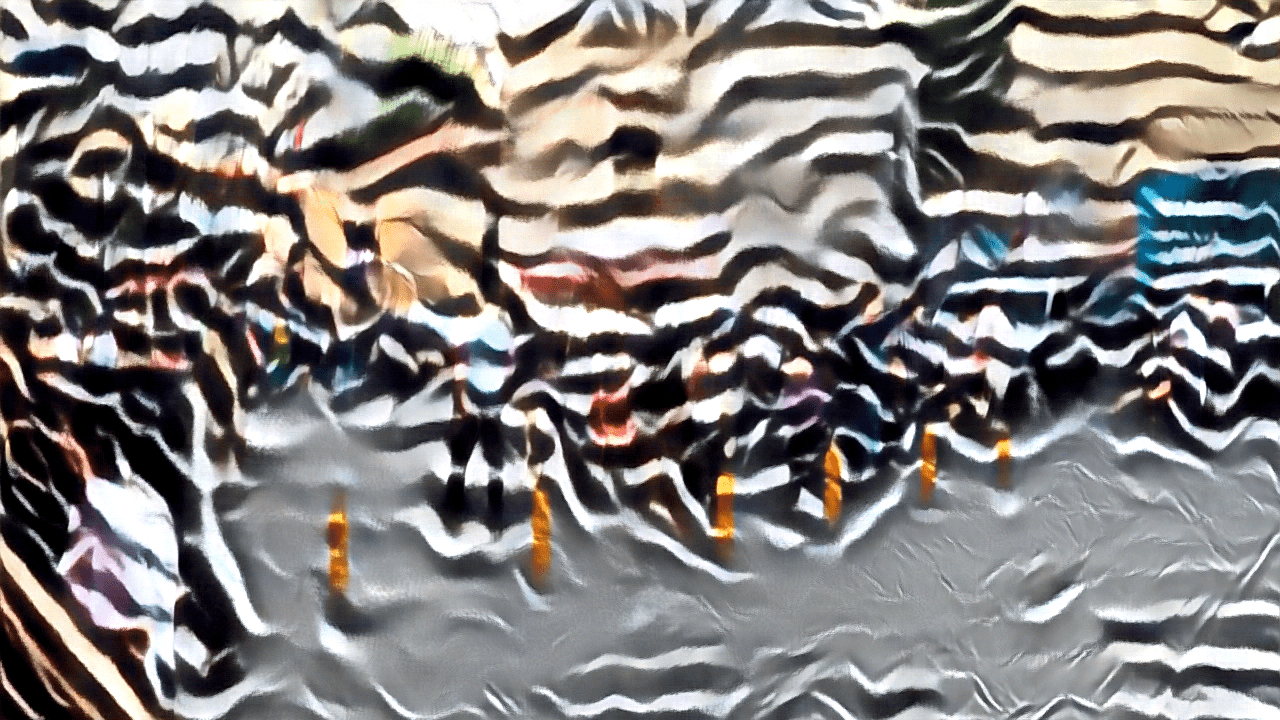} &
    \includegraphics[width=0.49\linewidth]{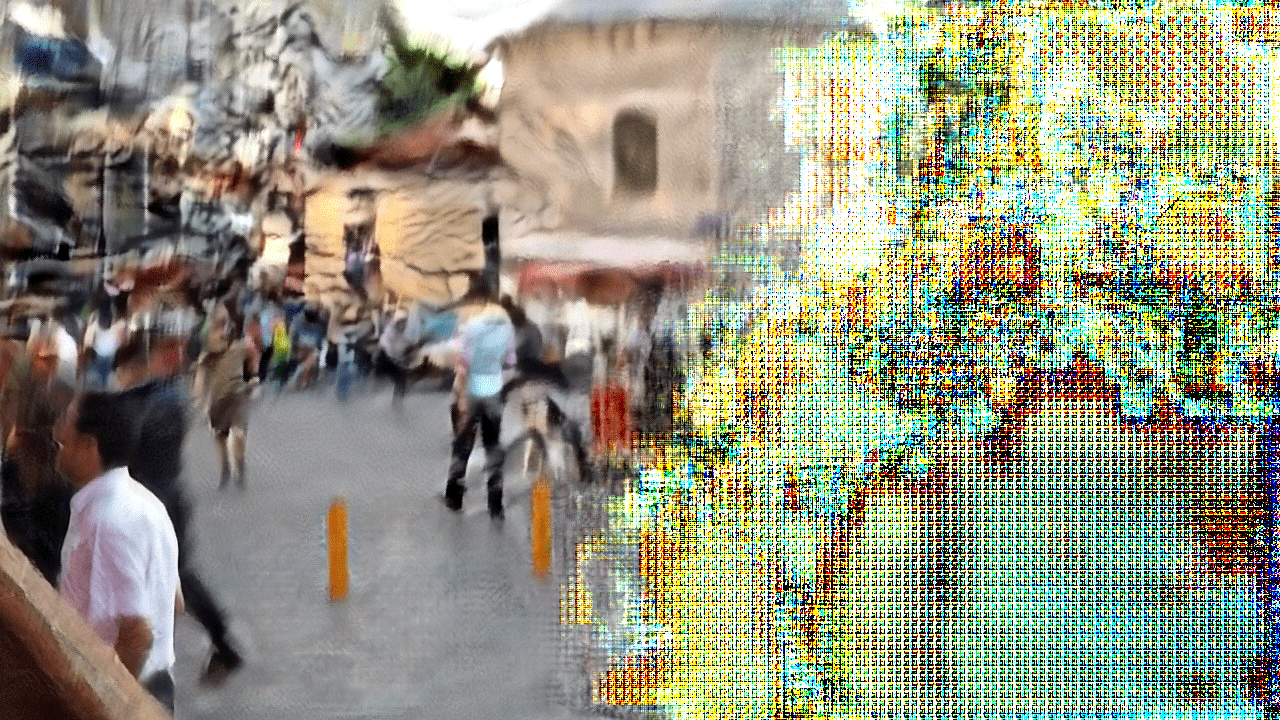}
\end{tabular}
\caption{Comparing images reconstructed by all considered models after 5 iterations of CosPGD attack
. We observe strong spectral artifacts in the reconstructed images.}
\label{fig:teaser}
\end{figure}
The goal of image restoration is to recover high-quality images from degraded observations. The degradation could be due to a variety of factors such as noise, blur, artifacts due to jpeg compression, raindrops, haze, and other factors.  Earlier methods for image restoration \cite{richardson1972bayesian,nonlocalmean,Combettes2004,bm3d,Babacan2012} employed carefully chosen priors and degradation models to derive degradation-specific restoration algorithms. Yet, such methods are limited by the strength of the image prior and the accuracy in modeling or estimating the degradation operator. 
 The past decade saw a  large-scale adoption of deep learning methods to image restoration \cite{su2022survey}, which outperformed the classical approaches. 
 Recent approaches \cite{zamir2022restormer,wang2022uformer,tu2022maxim} successfully adopt novel architectures such as Transformers \cite{vaswani2017attention,dosovitskiy2021an} and MLP-mixers \cite{tolstikhin2021mlp} for restoration.

Yet, CNNs, MLP-mixers as well as Transformer have been shown to be vulnerable to 
carefully crafted adversarial examples \cite{pgd,fgsm}. 
Recent work \cite{agnihotri2023cospgd, estimators_robustness,ijcai2022p211,ga2022deblurring} also confirms the existence of such vulnerabilities in deep learning-based image restoration. Yet, existing works mainly analyze the robustness of CNN-based restoration methods.
Conversely, with the introduction of novel network architectures such as vision Transformers \cite{liu2021swin,dosovitskiy2021an}, MLP mixers \cite{tolstikhin2021mlp}, and improved convolutional architectures \cite{liu2022convnet,bit2020} which outperform the earlier networks such as ResNets \cite{he2016deep}, there have been several studies on the robustness of these new architectures \cite{bhojanapalli2021understanding,shao2022on,tang2021robustart,croce2022interplay, agnihotri2023cospgd}. To the best of our knowledge, very limited works~\cite{croce2022interplay,NEURIPS2021_e19347e1} investigate the effect of architectural components and training recipes.
Existing works focus on image classification and do not study restoration. 
Thus to bridge this gap we investigate the adversarial robustness of recent Transformers specialized to image restoration. 

In this work we study the adversarial robustness of Transformer based restoration networks, Restormer \cite{zamir2022restormer}, and two architectures introduced in \cite{chen2022simple} the \emph{Baseline network} and the \emph{Non-linear Activation free Network~(NAFNet)}, both obtained by simplifying the original Restormer, with modifications to the channel attention and activation functions.
Further, to better understand the architectural design choices made by \cite{chen2022simple}, we include an \emph{Intermediate network} also considered by \cite{chen2022simple} which serves as a step between the Baseline network and NAFNet.
This study is particularly interesting as recent works \cite{xie2020smooth,NEURIPS2021_e19347e1} indicate that the choice of activation function significantly impacts adversarial robustness. 
We study the network robustness under standard and adversarial attacks, by considering $\ell_\infty$ perturbations crafted using PGD attack \cite{pgd} and  CosPGD attack proposed in \cite{agnihotri2023cospgd} for dense prediction tasks. 
We conduct our experiments on dynamic deblurring using the Go-Pro dataset \cite{gopro}.

Our experiments reveal that under standard training settings, Transformer based restoration networks are not robust to adversarial attacks in general.
As shown in Figure~\ref{fig:teaser}, the networks also exhibit distinct artifacts in the reconstructions under attack. 
The images from the Baseline network and the Restormer exhibit severe ringing artifacts~\cite{ringing_artifacts}, whereas the NAFNet reconstructs images with very strong grid and color artifacts under adversarial attack. 
We find that adversarial training can largely reduce the artifacts and significantly improve the robustness of all three networks.
However, the recently proposed NAFNet and Baseline network fail to rival the performance of Restormer, which leads us to contemplate the importance of the architectural components necessary to achieve  robust generalization.

The main contributions of this work can be summarized as follows:
\begin{itemize}
    \item We investigate the robustness of recently proposed Transformer based architectures for image restoration, namely image deblurring.
    \item We analyze the quality of the restored images and the spectral artifacts introduced by models under the aforementioned adversarial attacks.
    \item We understand the effects of defense strategy against adversarial attacks that consequently reduce the spectral artifacts in reconstructed images.
    \item Lastly, we study the effect of certain architectural design choices in the recently proposed \emph{state-of-the-art} image restoration model, NAFNet, to improve their robustness.
\end{itemize}

\section{Related Work}
\label{sec:related}
\paragraph{Transformers for Image Restoration} 
The past decade saw significant improvements in  image restoration, largely owing to the adoption of  deep networks trained on large datasets of clean and degraded images. 
While earlier restoration networks largely adopted CNN-based architectures, subsequent works also explored the use of attention mechanisms inside CNNs~\cite{zhou2020cross,niu2020single,Suin_2020_CVPR}. 
We refer \cite{su2022survey} for a  detailed survey on deep learning approaches to restoration. More recently, vision Transformers \cite{liu2021swin,dosovitskiy2021an} are increasingly adopted for several restoration tasks. 
While~\cite{liang2021swinir,zamir2022restormer,wang2022uformer,conde2022swin2sr,pmlr-v202-xiao23a} adopt Transformers for generic restoration tasks, a few works focus on specific restoration tasks by including such as deblurring~\cite{tsai2022stripformer}, deraining~\cite{liang2022drt}, dehazing~\cite{guo2022image,song2023vision}, removing degradations due to adverse weather conditions~\cite{valanarasu2022transweather}.  
These networks typically employ  encoder-decoder-based architectures with Transformer blocks combined with convolutions.

\paragraph{Adversarial Robustness of Image Restoration. } While the adversarial robustness of deep networks for image recognition is extensively studied, a few works also study the robustness of image restoration networks to adversarial attacks.
\cite{estimators_robustness,Choi_2020_ACCV,Yue21RobustSR}  evaluate adversarial robustness of deep learning-based image super-resolution. 
While \cite{Choi_2020_ACCV} propose adversarial regularization, \cite{Yue21RobustSR} propose frequency domain adversarial example detection, combined with random frequency masking to improve robustness.
\cite{ga2022deblurring} evaluate adversarial robustness of deblurring networks with and without the knowledge of the blur operator, and introduce targeted attacks on restoration.
In \cite{image-to-image}, the adversarial robustness of image-to-image translation models is studied, including restoration tasks, and adversarial training and different transformation-based defenses are evaluated.
Yan et al.~\cite{ijcai2022p211} investigate the robustness of image denoising to zero-mean adversarial perturbations and propose training with clean and adversarial samples to improve robustness.
Yu et al.~\cite{yu2022towards} investigate adversarial robustness of deep learning-based rain removal, and study the effect of architecture and training choices on robustness. Yet, these works do not focus on the more recent Transformer based restoration networks.
With the notable exception of \cite{agnihotri2023cospgd}, where they simply benchmark the adversarial performance of the image restoration networks recently proposed by \cite{chen2022simple}.

\paragraph{Robustness of Transformers \& other modern architectures. }
Recently, Vision Transformers (ViTs) \cite{dosovitskiy2021an,liu2021swin} have been successfully applied to  image recognition, outperforming the older ResNets. Follow-up works modified training schemes and architectures leading to more performant CNN architectures such as ConvNext \cite{liu2022convnet}, and hybrid models combining components of ViTs and CNNs \cite{ali2021xcit}. 
Following the introduction of these novel architectures, several works examined  the robustness properties of these models.
\cite{shao2021adversarial,bhojanapalli2021understanding,shao2022on,paul2022vision} suggest Transformers have better adversarial robustness than CNNs. 
However, \cite{Mahmood_2021_ICCV} shows that vision Transformers are also as vulnerable as CNNs under strong attacks.
\cite{NEURIPS2021_e19347e1} show that CNNs can achieve similar adversarial robustness as Transformers when trained using similar training recipes, yet Transformers still outperform CNNs on out-of-distribution generalization.
\cite{tang2021robustart} benchmark for robustness dependent on the network architecture. 
They find that Transformers are best suited against adversarial attacks while being extremely vulnerable to common corruptions~\cite{hendrycks2019benchmarking} and system noise.
Conversely, CNNs are more robust against common corruptions and system noise while being weakest against adversarial attacks. 
Further, they show that MLP-Mixers are not the best and also not the worst for both cases.

In their work, \cite{eccv2022_cnn_vs_Transformer} benchmark the robustness of state-of-the-art Transformers and CNN architectures and show that CNNs using ConvNext  architecture can be at least as robust as  Transformers for image recognition. 
Meanwhile \cite{croce2022interplay} analyzes the effect of different architectural components such as patches, convolution, activation, and attention, and demonstrates that ConvNexts have better adversarial robustness than ResNets. \cite{xie2020smooth} observe that smooth activation functions improve adversarial training as they enable better gradient updates to compute harder adversarial examples. Subsequent works  \cite{NEURIPS2021_e19347e1,croce2022interplay} also confirm improvement in robustness when GELU~\cite{gelu} activation functions are used in adversarial training.
While \cite{NEURIPS2021_e19347e1} attribute significant robustness gains in Transformers to the self-attention mechanism, \cite{wang2023can} identify other architectural components, including, the use of patches, larger kernels, reducing activation and normalization layers which when incorporated into CNNs lead to out of distribution robustness at least on par with Transformers without the use of attention.

In contrast, our work focuses on the investigation of the robustness of several recent Transformer based restoration models and shows interesting effects of adversarial attacks that can be attributed to different building modules of such models.
\begin{figure*}[ht]
    \centering
    \includegraphics[width=0.95\textwidth]{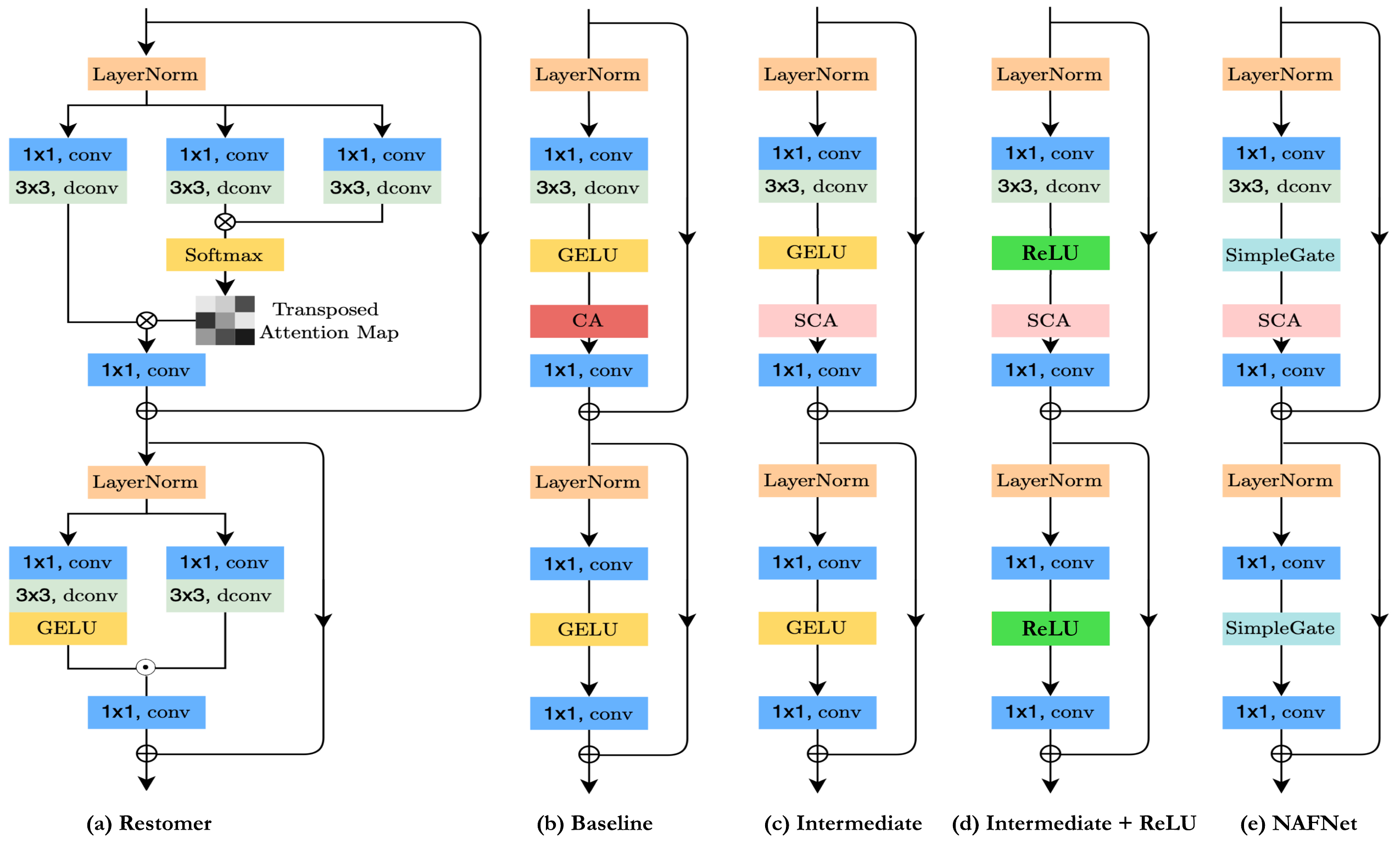}
    \caption{Modified visualization of repeating blocks of the architectures from \cite{chen2022simple} the considered \emph{Intermediate network} from \cite{chen2022simple} (please refer to (c)) and \emph{Intermediate~+~ReLU network} (please refer to (d)).}
    \label{fig:arch_blocks}
\end{figure*}
\section{Methodology}
\label{sec:method}
Following, we describe the attack framework used and the defense strategy used to combat the vulnerabilities of the architectures exposed by the adversarial attacks.

\subsection{Attack Framework} 
\label{subsec:method:attack}

Let $\mathbf{x}$ denote the ground-truth image, which is corrupted by a possibly non-linear degradation operator $\mathbf{A}$, resulting in an observation $\mathbf{y}^{\mathrm{clean}}$, which can be expressed as
\begin{equation}
    \mathbf{y}^{\mathrm{clean}} = \mathbf{A}(\mathbf{x}).
    \label{eq:measurement}
\end{equation}
Let  $\net$ be a (Transformer-based) neural network parameterized by $\theta$ trained to recover $\mathbf{x}$ from $\mathbf{y}^{\mathrm{clean}}$. 
In this work, we are interested in studying the stability of $\net$ to adversarial attacks that aim to degrade its performance through visually imperceptible changes to the inputs \cite{fgsm, pgd}. 
We evaluate the robustness to attacks using additive perturbations $\delta$ with $\ell_p$-norm constraints.
We generate the adversarial perturbations based on two powerful attack methods CosPGD \cite{agnihotri2023cospgd} developed for dense prediction tasks, and PGD attack \cite{pgd}, both of which we detail in the following.
The objective of the attack is to maximize the deviation of the network output from the ground truth as measured by a loss function $L$, subject to $\ell_p$ norm constraints on the perturbation:
\begin{equation}
 \underset{\delta}{\mathrm{maximize}}~L(\net(\mathbf{y}^{\mathrm{clean}} + \delta) ,~ \mathbf{x} )~ \text{  s.t.  }  \norm{\delta}_p \leq \epsilon.
 \end{equation}
 
\paragraph{PGD. } PGD is an iterative adversarial attack, where each sample is perturbed for a 
fixed amount of attack iterations (steps) with the intention of maximizing the loss further with each attack step.
 A single attack step in the PGD attack \cite{pgd} is given as follows,
\begin{align}
    \mathbf{y}^{\mathrm{adv}_{t+1}} &= \mathbf{y}^{\mathrm{adv}_t}+\alpha \cdot \mathrm{sign}\nabla_{\mathbf{y}^{\mathrm{adv}_t}}L(\net(\mathbf{y}^{\mathrm{adv}_t}), \mathbf{x})\\
    \delta &= \phi^{\epsilon}(\mathbf{y}^{\mathrm{adv}_{t+1}} - \mathbf{y}^{\mathrm{clean}})\nonumber\\
    \mathbf{y}^{\mathrm{adv}_{t+1}} &= \phi^{r}(\mathbf{y}^{\mathrm{clean}}+ \delta)\nonumber
    \label{eq:pgd}
\end{align}

where the adversarial example $\mathbf{y}^{\mathrm{adv}_{t+1}}$ at step $t+1$,  is updated using the adversarial example from the previous step $\mathbf{y}^{\mathrm{adv}_{t}}$, $\nabla$ represents the gradient operation, $\alpha$ is the step size for the perturbation, $\phi^{\epsilon}$ is denotes projection onto the appropriate $\ell_p$-norm ball of radius $\epsilon$, depending on the  $\ell_p$ norm constraints on $\delta$, and 
$\phi^{r}$ clips the adversarial example to lie in the valid intensity range of images (between [0, 1]).
Prior works evaluating the adversarial robustness of image restoration networks consider $L$ to be the reconstruction loss (MSE loss) to obtain adversarial examples  maximizing  the reconstruction error.
\paragraph{CosPGD. } Instead of directly utilizing the averaged pixel-wise losses in PGD attack steps, \cite{agnihotri2023cospgd} propose to weigh the pixel-wise losses using the cosine similarity between the network output and the ground truth (both scaled by softmax), to reduce the importance of the pixels which already have a large error in the previous iterations, and enable the attack to focus on the pixels with low error. 
For the task of restoration (a regression task),
CosPGD attack steps for an untargeted attack are given as:
\begin{align}
\mathbf{x}^{\mathrm{adv}_{t}} &= \net(\mathbf{y}^{\mathrm{adv}_{t}})\\
L_{\mathrm{cos}}&=\sum\mathrm{cossim}(\Psi(\mathbf{x}^{\mathrm{adv}_{t}}), \Psi(\mathbf{x}))\odot L(\mathbf{x}^{\mathrm{adv}_{t}}, \mathbf{x})\nonumber\\
\mathbf{y}^{\mathrm{adv}_{t+1}} &= \mathbf{y}^{\mathrm{adv}_{t}} + \alpha \cdot \mathrm{sign}\nabla_{\mathbf{y}^{\mathrm{adv}_{t}}}L_{\mathrm{cos}}\nonumber\\
\delta &= \phi^{\epsilon}(\mathbf{y}^{\mathrm{adv}_{t+1}} - \mathbf{y}^{\mathrm{clean}})\nonumber\\
    \mathbf{y}^{\mathrm{adv}_{t+1}} &= \phi^{r}(\mathbf{y}^{\mathrm{clean}}+ \delta),\nonumber
\end{align}
where $\Psi$ is the softmax function, $\odot$ denotes  point-wise multiplication,  and the cosine similarity (cossim) is given by
\begin{equation}
\label{eq:cossim}
    \mathrm{cossim}(\overrightarrow{\mathbf{u}},\overrightarrow{\mathbf{v}})=\dfrac{\overrightarrow{\mathbf{u}} \cdot \overrightarrow{\mathbf{v}}}
{||\overrightarrow{\mathbf{u}}|| \cdot  ||\overrightarrow{\mathbf{v}}|| }
\end{equation}
\cite{agnihotri2023cospgd} demonstrate that this approach results in a stronger attack for pixel-wise regression tasks than a PGD attack.
We use both PGD and CosPGD in our robustness evaluation.
\subsection{Architectures: from Restormer to NAFNet}
\label{subsec:method:arch}

We evaluate the adversarial robustness of  \emph{Restormer} \cite{zamir2022restormer}, a Transformer based architecture for image restoration  and two architectures introduced in \cite{chen2022simple} by  modifying the Restormer architecture. 
Restormer \cite{zamir2022restormer} has a UNet  \cite{ronneberger2015u} like encoder-decoder architecture, using multi-head channel-wise attention modules, gated linear units \cite{dauphin2017language} and  depth-wise convolutions in the feed-forward network. 
This network achieved state-of-the-art performance in image restoration at the time of its publication. 
The authors in \cite{chen2022simple} investigate whether it is possible to retain the performance of Restormer, with a simplified architecture. 
After a thorough ablation study, they propose a simplified \emph{Baseline} network that improved upon the SOTA performance. 
The Baseline network utilizes GELU activations \cite{gelu} and replaces multi-headed self-attention in \cite{zamir2022restormer} with a channel attention module \cite{hu2018squeeze}. 
Without loss in i.i.d. performance, they further simplify this architecture by removing activation functions altogether, replacing GELU with a \emph{simple gate} which performs element-wise product of feature maps, and replacing the channel attention by a \emph{simplified channel attention} without activation functions. 
The resulting network is referred to as a Nonlinear Activation-Free Network (NAFNet).
In contrast to \cite{chen2022simple} who focus on performance with clean inputs, we analyze the adversarial robustness of these networks, which also allows us to evaluate the effect of different activation functions and attention mechanisms on the robustness of restoration transformers.
In Figure \ref{fig:teaser}, we observe that NAFNet has significantly different artifacts in the reconstructed images compared to Restormer and the Baseline network.
One might simply hypothesize that these strange artifacts which appear to be the cumulative effect of aliasing and color mixing are due to the use of `Simple Gate' in place of a non-linear activation function like GELU. 
To confirm this hypothesis we additionally consider an \emph{Intermediate network}, from \cite{chen2022simple}.
In this \emph{Intermediate network} we replace the \emph{channel attention} in the baseline network with the \emph{simplified channel attention} but retain the GELU activation.
Additionally, to better understand the role of non-linear activation functions in this context, we consider an architecture the same as the \emph{Intermediate network} but with ReLU activations instead of GELU.
In Figure~\ref{fig:arch_blocks}, we modify the visualization by \cite{chen2022simple}, to present the repeating blocks of all the considered architectures in our work.

\subsection{Defenses}
As discussed in Section~\ref{sec:intro}, we observe in Figure~\ref{fig:teaser} that all considered architectures are vulnerable to adversarial attacks.
Prior work \cite{fgsm, pgd, gu2022segpgd} has shown that adversarial training is an effective defense against adversarial attacks.
Thus we use adversarial training as a defense strategy.
\paragraph{Adversarial Training. } We use the FGSM attack as proposed by \cite{fgsm} to generate adversarial samples during training.
Adversarial training can be hypothesized as a min-max problem, where we try to find perturbations for the samples such that the loss is maximized while training the network on these samples to minimize the loss of the model over training iterations. 
PGD attack is essentially a multi-step extension of FGSM attack, and thus the loss that FGSM attack attempts to maximize remains the same.
Additionally, the attack step of FGSM is also the same as described in Section~\ref{subsec:method:attack}, with one notable difference being that in the case of an FGSM attack, the attack step size $\alpha$ is equal to the permissible perturbation size of $\epsilon$.

While training, to avoid overfitting to adversarial samples, and enable the model to make reasonable reconstructions on unperturbed samples we use the training regime similar to \cite{gu2022segpgd} and use only 50\% of the sample in the training batch to generate perturbed adversarial samples and use the other 50\% samples unperturbed.
Thus, the effective learning objective is as described by Equation~\ref{eq:adv_train}.
\begin{equation}
 \underset{\theta}{\mathrm{minimize}}~\sum_i L(\net(\mathbf{y}^{\mathrm{clean}_i}) ,~ \mathbf{x}_i)+ \sum_j L(\net(\mathbf{y}^{\mathrm{adv}_j}) ,~ \mathbf{x}_j)
\label{eq:adv_train}
 \end{equation}
where the indices $i$ and $j$ correspond to the examples from the clean and adversarial batch splits, and  FGSM adversarial examples are generated as:
\begin{equation}
\mathbf{y}^{\mathrm{adv}_{j}} = \phi^{r}(\mathbf{y}^{\mathrm{clean}_j}+ \phi^{\epsilon}( \epsilon \cdot \mathrm{sign}\nabla_{\mathbf{y}_j}L(\net(\mathbf{y}_j), \mathbf{x}_j)))
\label{eq:fgsm}
\end{equation}

\section{Experiments}
\label{sec:exp}
In this work on image restoration, we focus on reconstructing deblurred images using a few recently proposed image restoration networks.

\subsection{Experimental Setup}
\label{subsec:exp:setup}

\noindent\textbf{Networks. } We consider Restormer proposed by \cite{zamir2022restormer}, and Baseline network and NAFNet proposed by \cite{chen2022simple} with width 32.
For understanding the design choices that lead to NAFNet producing reconstructed images with significantly different spectral artifacts than the other considered networks, we also consider an \emph{Intermediate network} and \emph{Intermediate + ReLU}.
This \emph{Intermediate network} with width 32 has also been considered by \cite{chen2022simple} when discussing design choices to arrive from the Baseline network to NAFNet.
These networks are similar to the Baseline, except it has the ``simplified channel attention" as proposed by \cite{chen2022simple} rather than the ``channel attention" used in the Baseline network. 
We visualize all the considered architectures in Figure \ref{fig:arch_blocks}.

\noindent\textbf{Dataset. } For our experiments we use the GoPro image deblurring dataset\cite{gopro}.
This dataset consists of 3~214 real-world images with realistic blur and their corresponding ground truth (deblurred images) captured using a high-speed camera.
The dataset is split into 2~103 training images and 1~111 test images.

\noindent\textbf{Metrics. } We report the PSNR and SSIM scores of the reconstructed images w.r.t. to the ground truth images, averaged over all images.
PSNR stands for Peak Signal-to-Noise ratio, a higher PSNR indicates a better quality image or an image closer to the image to which it is being compared.
SSIM stands for Structural similarity\cite{ssim}.
A higher SSIM score corresponds to better higher similarity between the reconstruction and the ground-truth image.

\noindent\textbf{Training Regimes. } For Restormer and its adversarial training counterpart (`+ADV') we follow the training procedure used by \cite{zamir2022restormer} except due to computational limitations we do not train on the last recommended patch size 384.
For the Baseline network, NAFNet, and its counterparts we follow the training regime used by \cite{chen2022simple}.

\noindent\textbf{Adversarial Training. } We used FGSM \cite{fgsm} adversarial training for efficiency. The maximum allowed perturbation for the adversaries is set to $\epsilon = \frac{8}{255}$. 
We use `+ADV' after the model name to denote that the model has been trained with FGSM adversarial training.

\noindent\textbf{Adversarial Attacks. } We consider PGD and CosPGD attacks. 
Following the procedure by \cite{agnihotri2023cospgd}, we use $\epsilon\approx\frac{8}{255}$, $\alpha$(attack step size)$=0.01$.
We consider attack iterations $\in$ \{5, 10, 20\} for our attacks.
We use MSE loss for generating adversarial samples for all networks.

\subsection{Results}
\label{subsec:exp:results}
The good performance of image restoration models on unperturbed samples is indubitably essential for possible real-world applications. 
However, the generalization ability of these models to perturbed samples has to be better understood for their reliability in safety-critical applications such as medical imaging, autonomous driving, etc. 
To this effect, we study the performance of the considered networks on both clean~(unperturbed) and adversarial~(perturbed) samples.
Further, to overcome the observed shortcomings of these models, we harden them using adversarial training. 
\begin{table}[t]
    \centering
    \caption{Performance of the different considered networks and their counterparts on clean~(unperturbed) GoPro test images. While NAFNet has highest PSNR value, Restormer is slightly better in terms of SSIM. All models slightly suffer from adversarial training when evaluated on clean data, which is to be expected.}
    \small
    \begin{tabular}{l@{\hspace{0.5cm}}|c@{\hspace{0.5cm}}c}
    \toprule
    
    Architecture & PSNR & SSIM \\
    \toprule
         Restormer & 31.99 & \textbf{0.9635} \\
          ~~~~ + ADV & 30.25 & 0.9453 \\
    \midrule
        Baseline  & 32.48 & 0.9575 \\
           ~~~~ + ADV & 30.37 & 0.9355 \\
    \midrule
        NAFNet  & \textbf{32.87} & 0.9606 \\
           ~~~~ + ADV & 29.91 & 0.9291 \\ 
    \bottomrule
    \end{tabular}
    
    \label{tab:clean_perf}
\end{table}

As observed in Figure~\ref{fig:teaser}, under adversarial attack both Restormer and Baseline network induce ringing-like artifacts in the restored images. 
However, NAFNet introduces aliasing like grid artifacts and color mixing in the restored images.

We report the performance of three networks along with adversarial training over clean images in Table~\ref{tab:clean_perf}.

\begin{table*}[h]
    \centering
    \caption{Comparison of performance of the considered models against CosPGD and PGD attacks with various attack strengths. Attack strength increases with the number of attack iterations~(itrs). Note that \textit{Intermediate + ReLU }achieves reasonably robust results entirely without adversarial training.}
    \scalebox{0.85}{
    \begin{tabular}{@{}l|cc|cc|cc|cc|cc|cc@{}}
    \toprule
    \multirow{3}{*}{Architecture} & \multicolumn{6}{c|}{CosPGD} & \multicolumn{6}{c}{PGD} \\
    & \multicolumn{2}{c|}{5 attack itrs } & \multicolumn{2}{c|}{10 attack itrs } & \multicolumn{2}{c|}{20 attack itrs } & \multicolumn{2}{c|}{5 attack itrs } & \multicolumn{2}{c|}{10 attack itrs } & \multicolumn{2}{c}{20 attack itrs } \\
    & PSNR & SSIM & PSNR & SSIM & PSNR & SSIM & PSNR & SSIM & PSNR & SSIM & PSNR & SSIM \\
    \toprule
         \textbf{Restormer} & 11.36 & 0.3236 & 9.05 & 0.2242 & 7.59 & 0.1548 & 11.41 & 0.3256 & 9.04 & 0.2234 & 7.58 & 0.1543 \\
         ~~~~ + \textbf{ADV} & \textbf{24.49} & 0.81 & \textbf{23.48} & \textbf{0.78} & 21.58 & 0.7317 & \textbf{24.5} & 0.8079 & \textbf{23.5} & \textbf{0.7815} & 21.58  & 0.7315 \\
          \midrule
        Baseline &  10.15 & 0.2745  &  8.71 & 0.2095  &  7.85 & 0.1685  &  10.15 & 0.2745 & 8.71 & 0.2094 & 7.85 & 0.1693 \\
          ~~~~ + ADV & 15.47 & 0.5216  &  13.75 & 0.4593  &  12.25 & 0.4032  & 15.47 & 0.5215  &  13.75 & 0.4592  & 12.24 & 0.4026   \\
          \midrule
        NAFNet & 8.67 & 0.2264 & 6.68 & 0.1127  & 5.81 & 0.0617 & 10.27 & 0.3179  & 8.66 & 0.2282  &  5.95 & 0.0714\\
          ~~~~ + ADV & 17.33 & 0.6046 & 14.68 & 0.509 &  12.30 & 0.4046  & 15.76 & 0.5228  & 13.91 & 0.4445  & 12.73 & 0.3859 \\
        \midrule
        \textbf{Intermediate} & 6.0224 & 0.0509 & 5.8166 & 0.0366 & 5.7199 & 0.0315 & 6.0225 & 0.0509 & 5.8158 & 0.0365 & 5.7173 & 0.0314 \\
        \textbf{~~~~ + ADV} & 24.02 & \textbf{0.8213} & 22.01 & 0.7775 & 20.15 & 0.7286 & 24.02 & \textbf{0.8213} & 21.98 & 0.7770 & 20.15 & 0.7286 \\  
        \midrule
         \textbf{Intermediate + ReLU} & 13.87 & 0.4093 & 11.63 & 0.3128 & 10.29 & 0.2538 & 13.87 & 0.4094 & 11.62 & 0.3127 & 10.29 & 0.2542 \\  
         \textbf{~~~~ + ADV} & 23.90 & 0.8046 & 22.46 & 0.7637 & \textbf{21.85} & \textbf{0.7484} & 23.91 & 0.8046 & 22.47 & 0.7638 & \textbf{21.84} & \textbf{0.7481} \\ 

    \bottomrule
    \end{tabular}    
    }
    \label{tab:adv_perf}
\end{table*}
Further, to study the generalization ability of these networks we adversarially attack the networks and report the findings in Table~\ref{tab:adv_perf}.

With standard training protocol, Restormer is marginally more robust in comparison to the Baseline network with fewer attack iterations, however, this difference reduces as the number of attack iterations increases. 
With adversarial training using FGSM adversarial examples, we observe improvement in the robustness of all three networks.
Interestingly,  the gain in performance of Restormer when trained with FGSM is significantly better than that of the Baseline network and NAFNet.
This indicates that Restormer has a much higher potential of being generalizable than both the Baseline network and NAFNet.
This raises doubts over the claims by \cite{chen2022simple} regarding the Baseline network and NAFNet having ``comparable or better performance" to the recent \textit{state-of-the-art} image restoration models.
Their claim holds true for clean samples, however with just slight perturbation ($\epsilon=\frac{8}{255}$), the performance of their proposed models drops significantly.
Contrary to this, \emph{Intermediate+ReLU} is significantly more robust, across attack iterations. 
We discuss this further in Section~\ref{subsec:discuss:intermediate}.
\begin{figure*}[htb]
    \centering 
    \scalebox{0.87}{
    \begin{tabular}{@{}c@{}c@{\hspace{0.1cm}}c@{\hspace{0.1cm}}c@{\hspace{0.1cm}}c@{\hspace{0.1cm}}c@{}}
    \multicolumn{2}{c}{MODEL} & NO ATTACK & 5 iterations & 10 iterations & 20 iterations\\
  \rotatebox{90}{\textbf{Restormer}} & & 
  \includegraphics[width=0.215\textwidth]{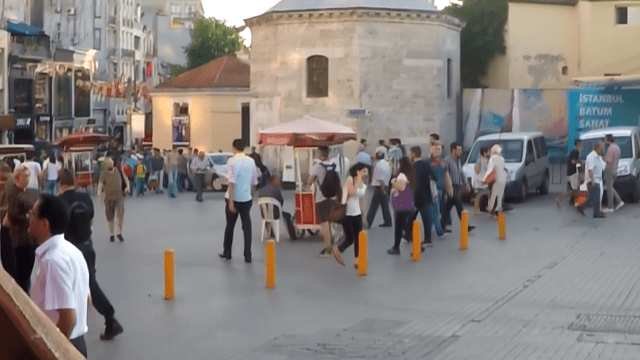}&
  \includegraphics[width=0.215\textwidth]{restormer_cospgd_5_GOPR0384_11_00-000002.png}&
  \includegraphics[width=0.215\textwidth]{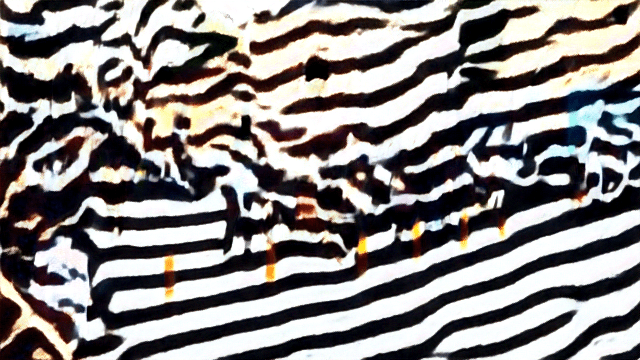}&
  \includegraphics[width=0.215\textwidth]{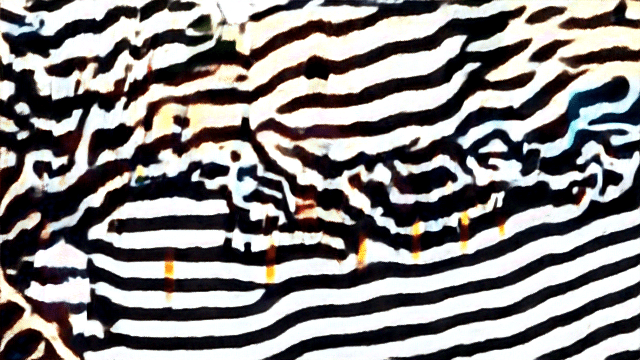}\\
  
  \rotatebox{90}{\textbf{Baseline}} & & 
  \includegraphics[width=0.215\textwidth]{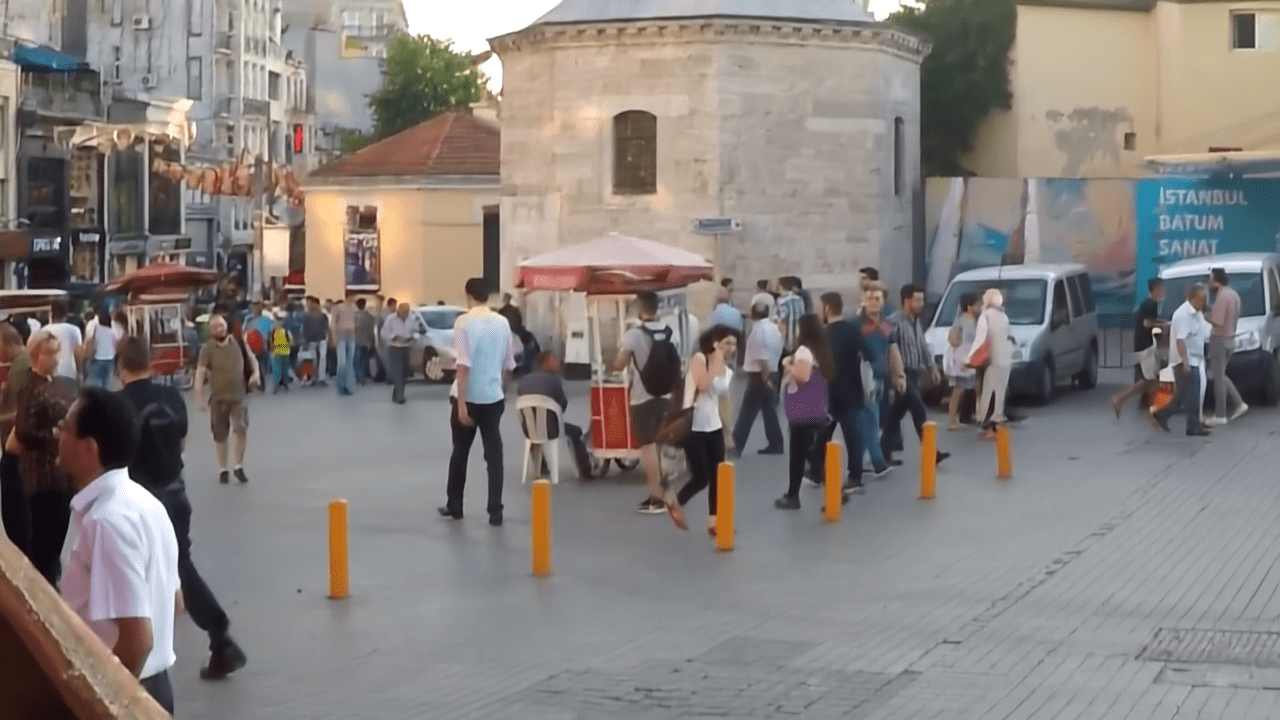}&
  \includegraphics[width=0.215\textwidth]{baseline_cospgd_5_GOPR0384_11_00-000002.png}&
  \includegraphics[width=0.215\textwidth]{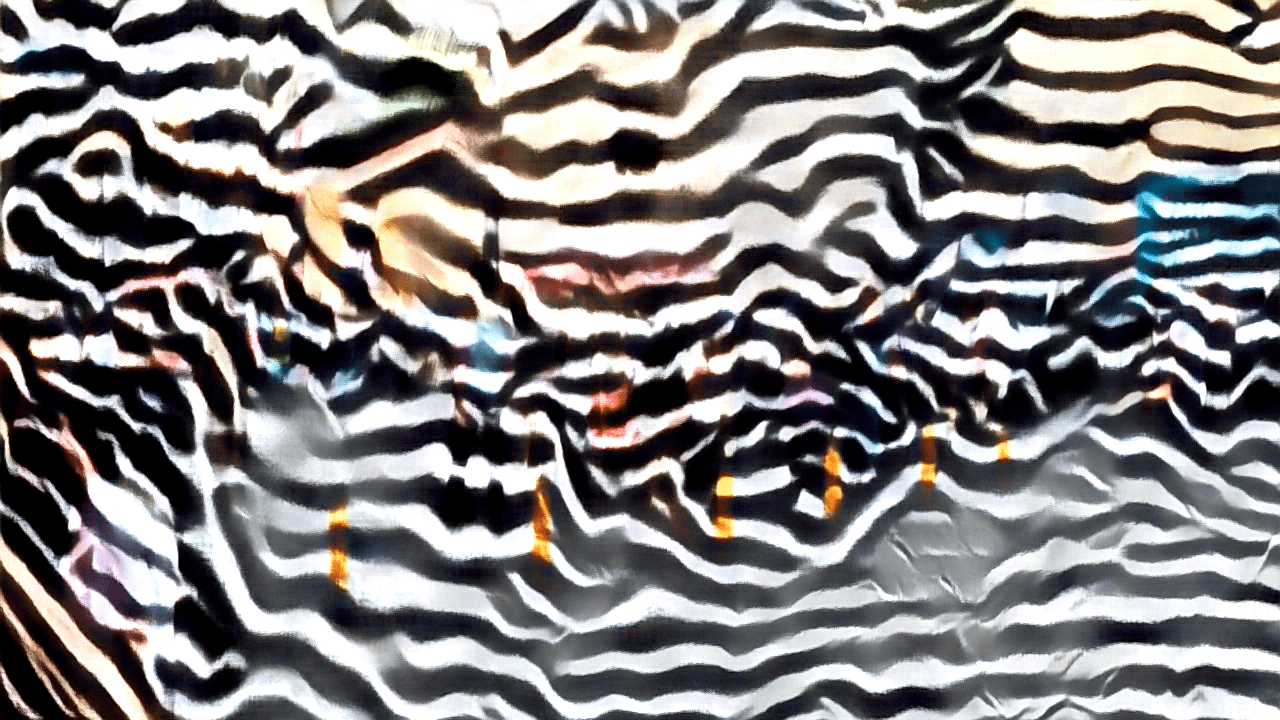}&
  \includegraphics[width=0.215\textwidth]{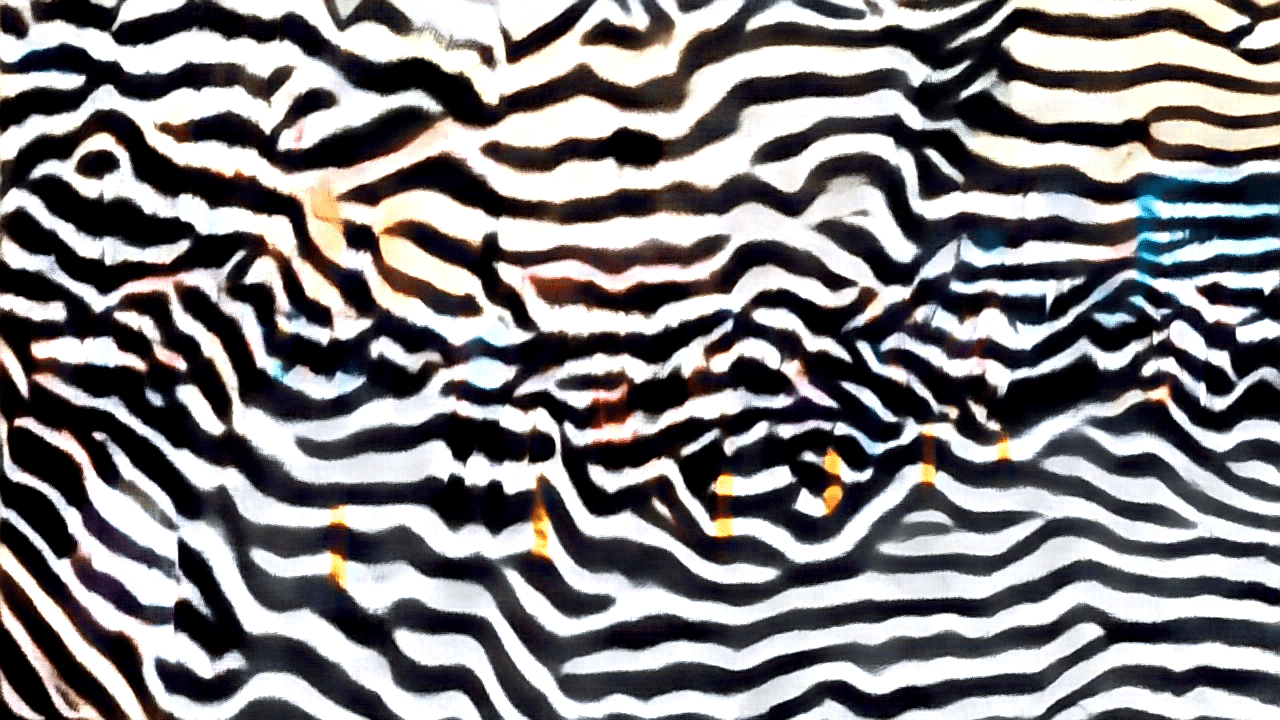}\\
  
  \rotatebox{90}{\textbf{NAFNet}} & & 
  \includegraphics[width=0.215\textwidth]{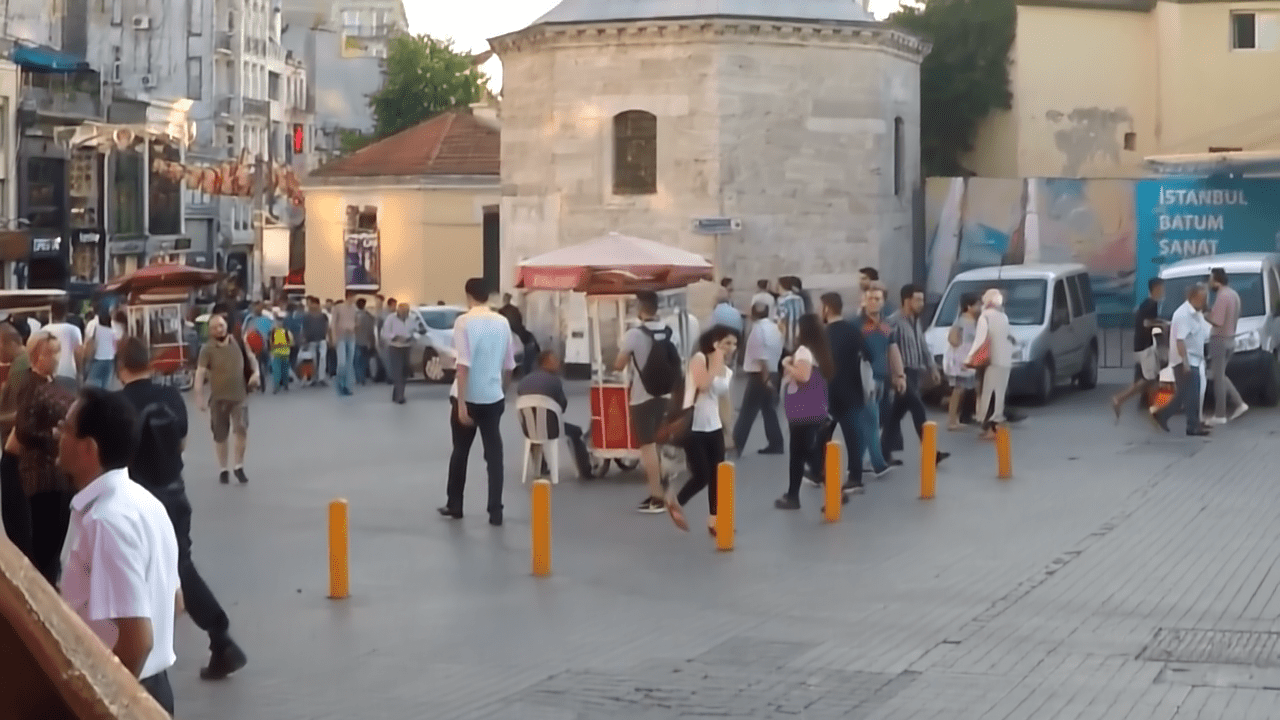}&
  \includegraphics[width=0.215\textwidth]{nafnet_cospgd_5_GOPR0384_11_00-000002.png}&
  \includegraphics[width=0.215\textwidth]{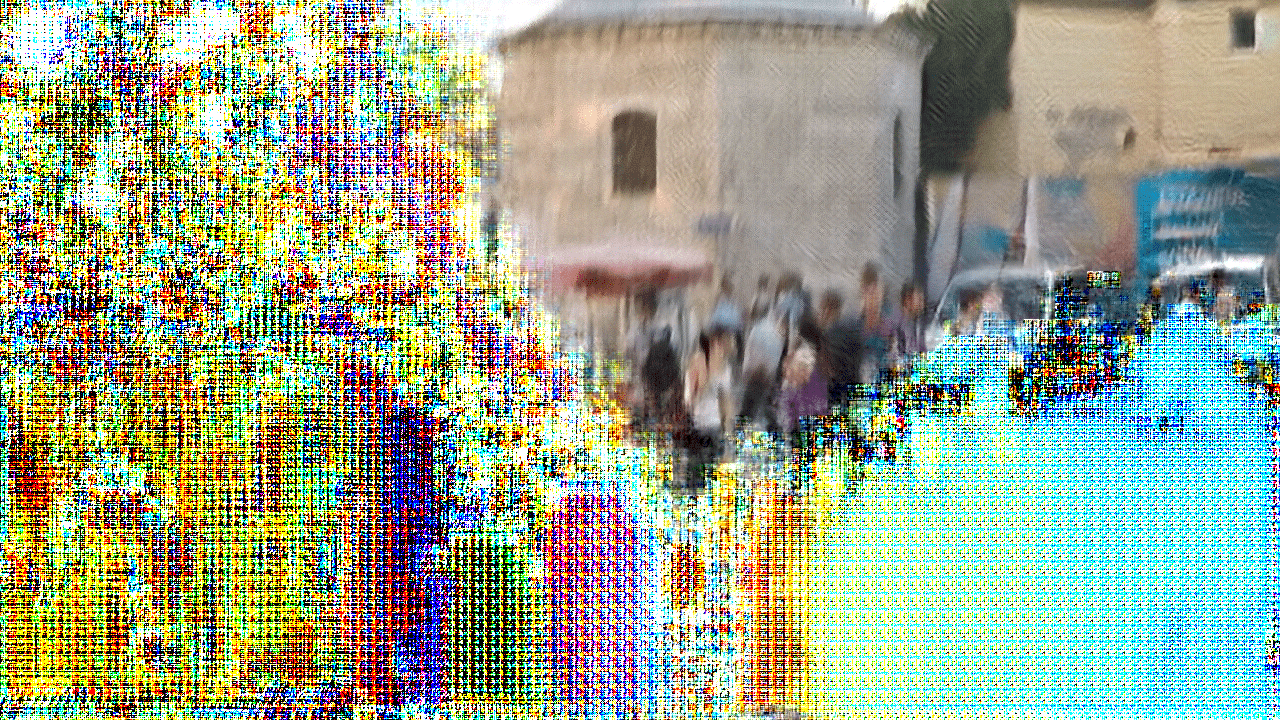}&
  \includegraphics[width=0.215\textwidth]{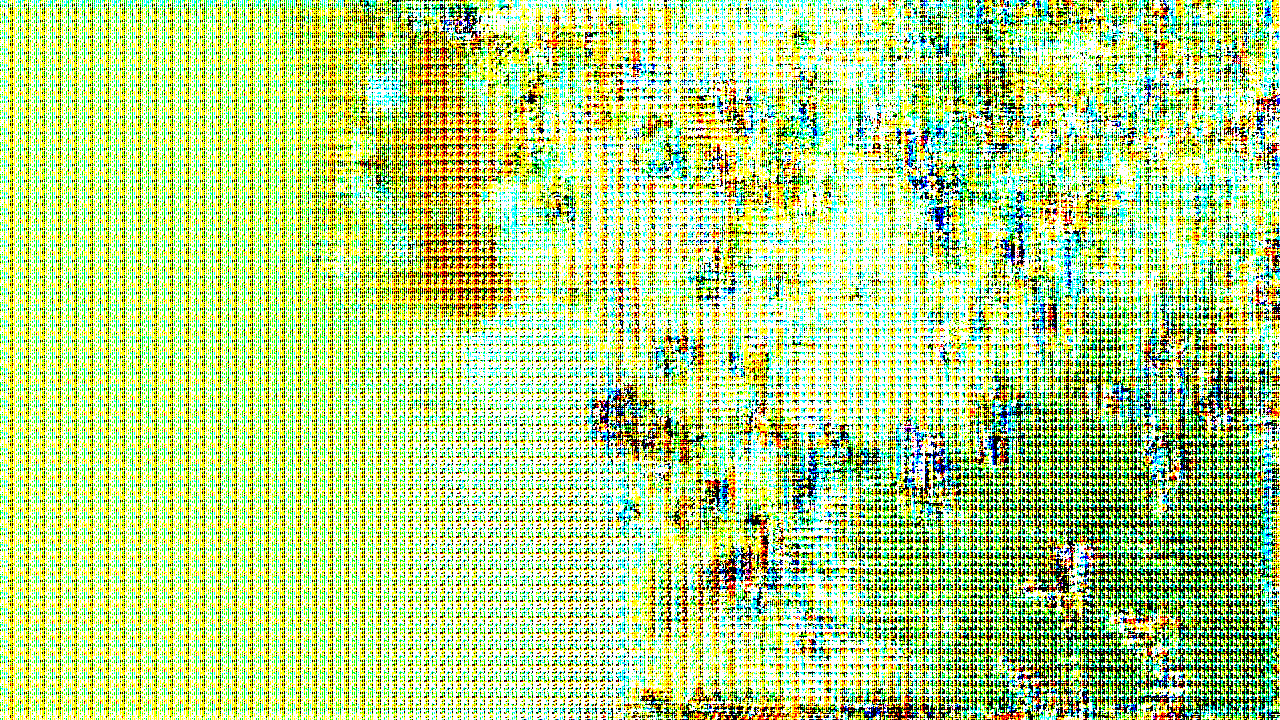}\\

  \rotatebox{90}{\textbf{Restormer}} & \rotatebox{90}{\textbf{~~~~+ADV}} & 
  \includegraphics[width=0.215\textwidth]{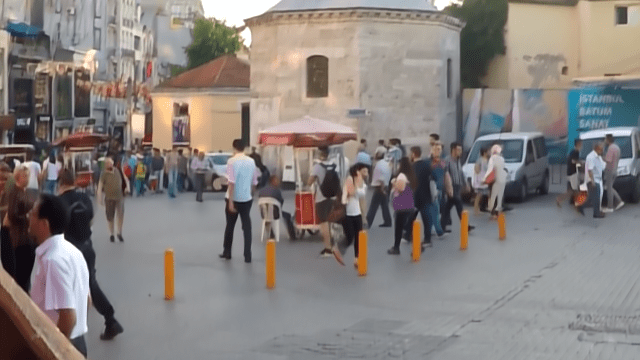}&
  \includegraphics[width=0.215\textwidth]{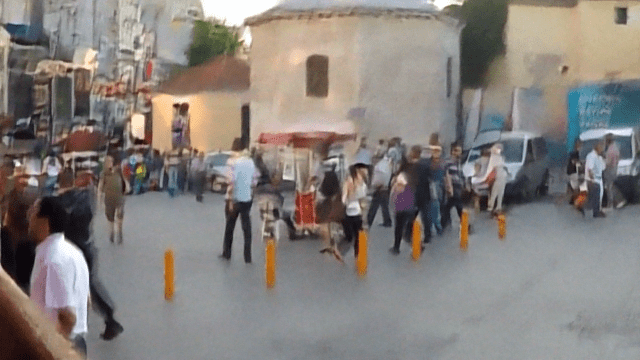}&
  \includegraphics[width=0.215\textwidth]{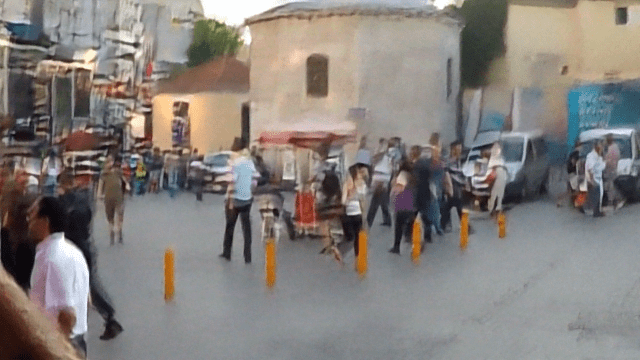}&
  \includegraphics[width=0.215\textwidth]{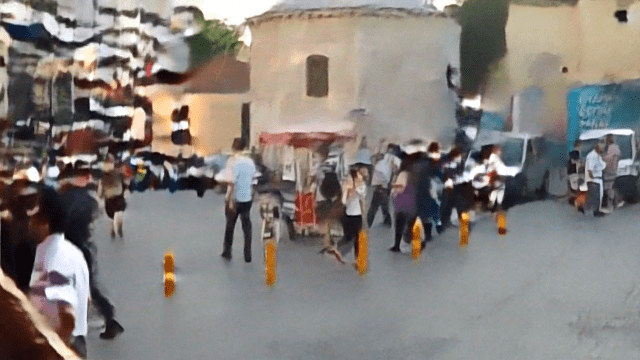}\\
  
  \rotatebox{90}{\textbf{Baseline}} & \rotatebox{90}{\textbf{~~~~+ADV}} & 
  \includegraphics[width=0.215\textwidth]{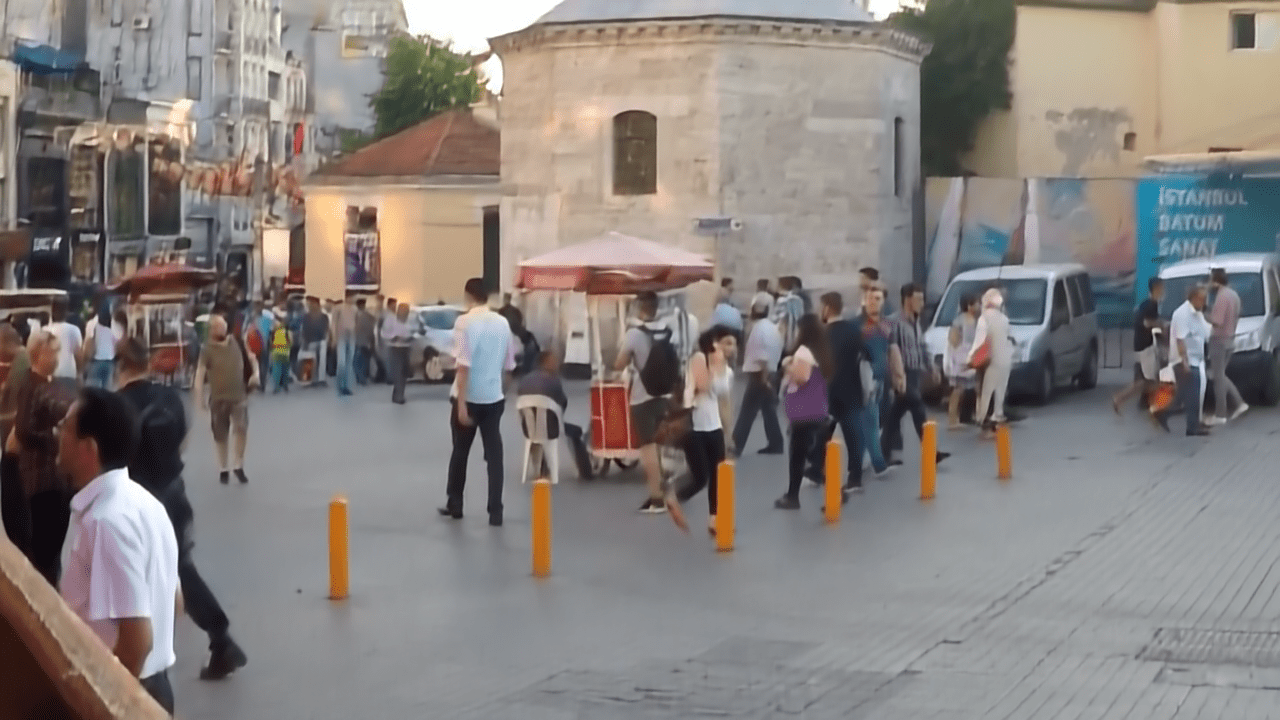}&
  \includegraphics[width=0.215\textwidth]{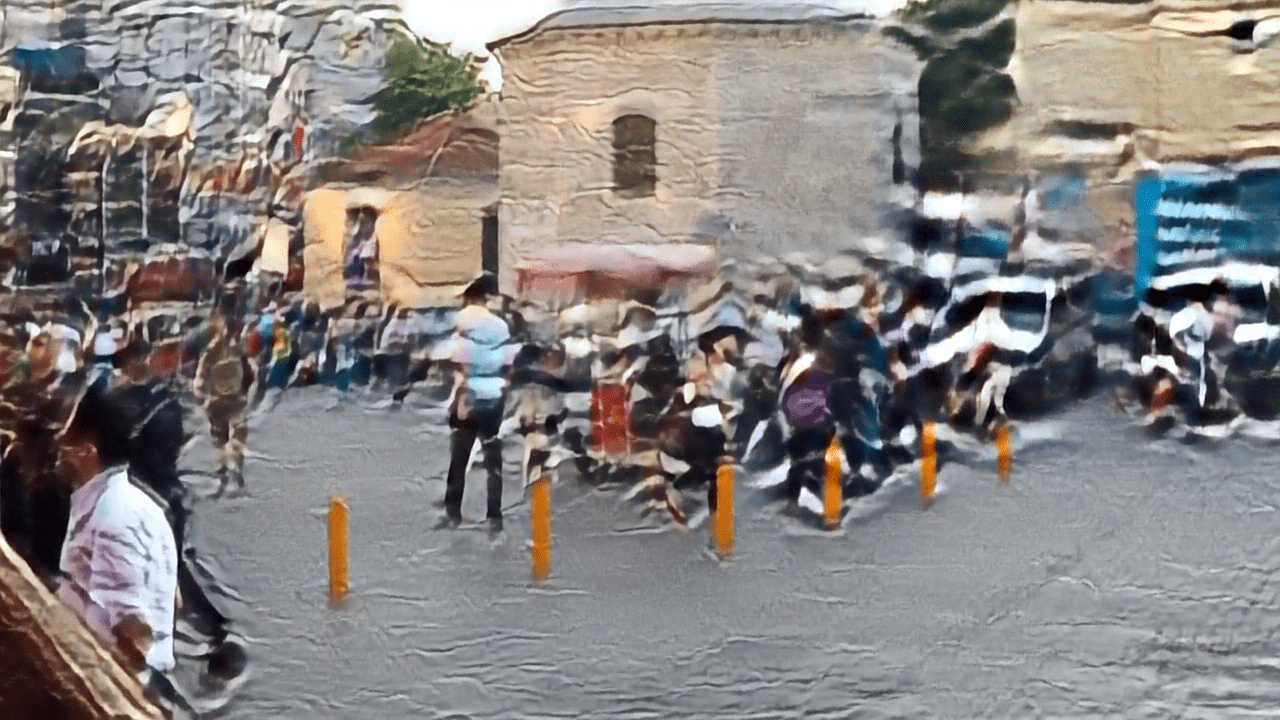}&
  \includegraphics[width=0.215\textwidth]{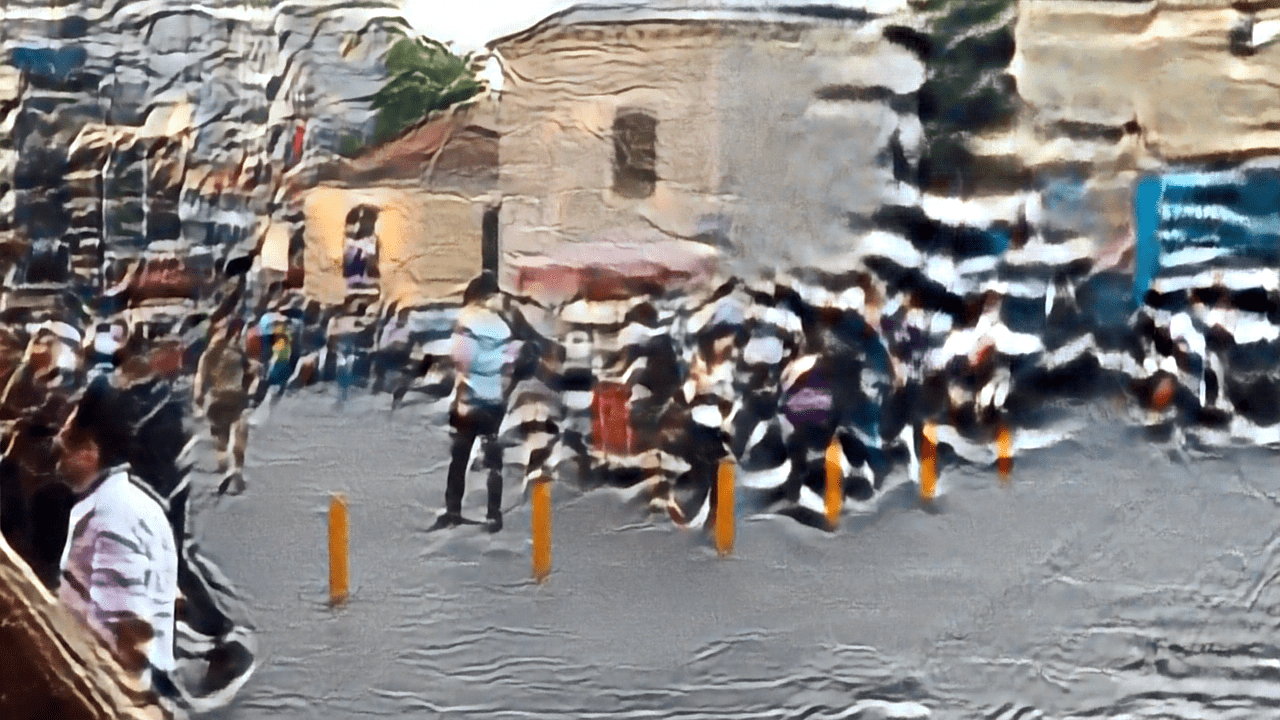}&
  \includegraphics[width=0.215\textwidth]{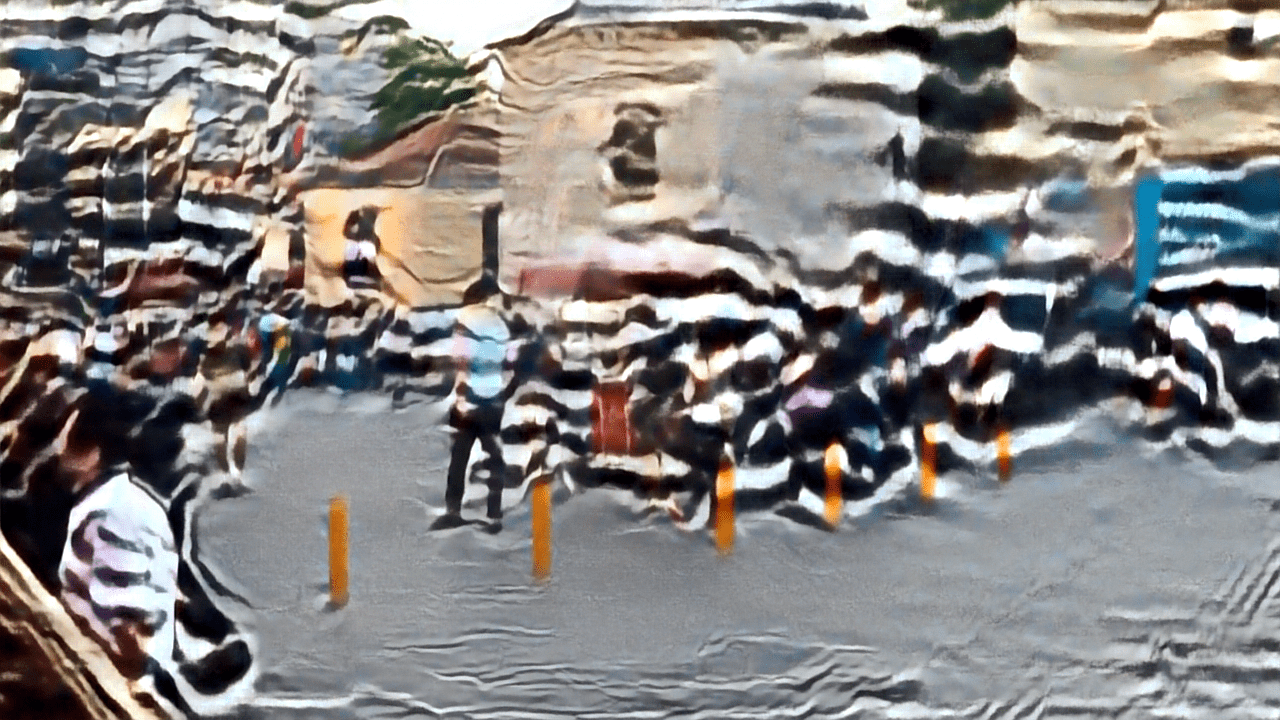}\\

  \rotatebox{90}{\textbf{NAFNet}} & \rotatebox{90}{\textbf{~~~~+ADV}} & 
  \includegraphics[width=0.215\textwidth]{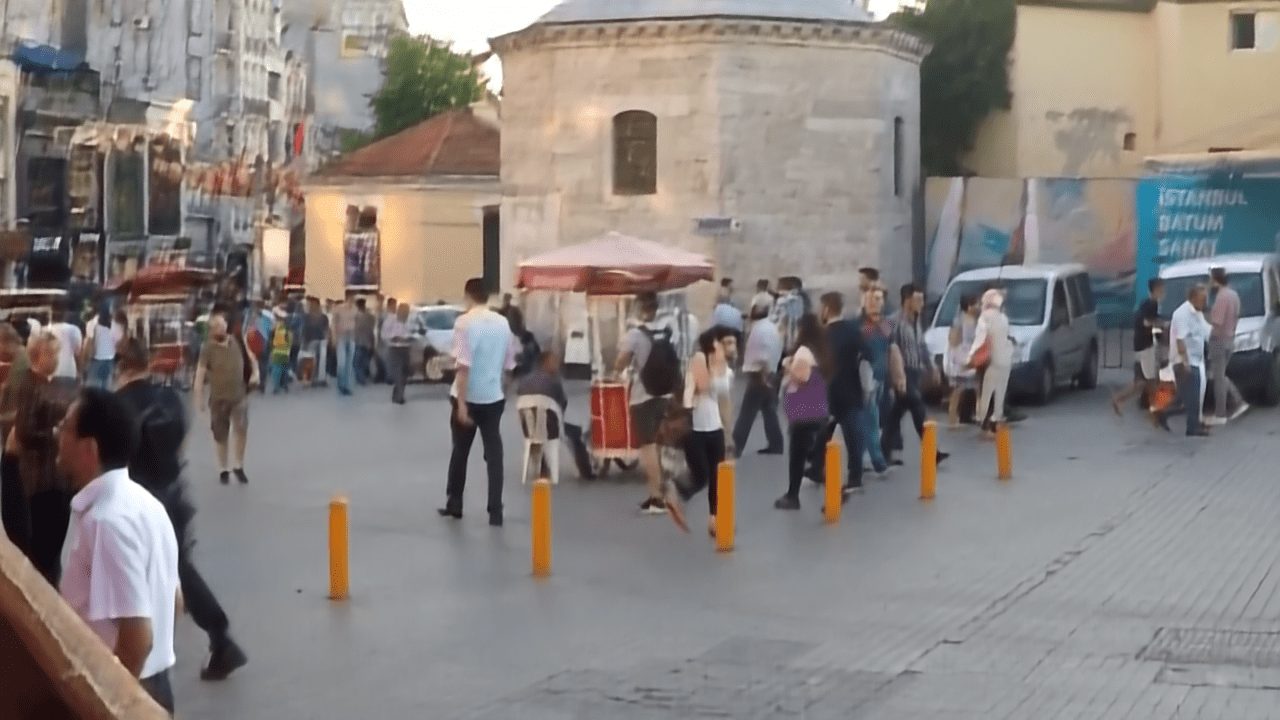}&
  \includegraphics[width=0.215\textwidth]{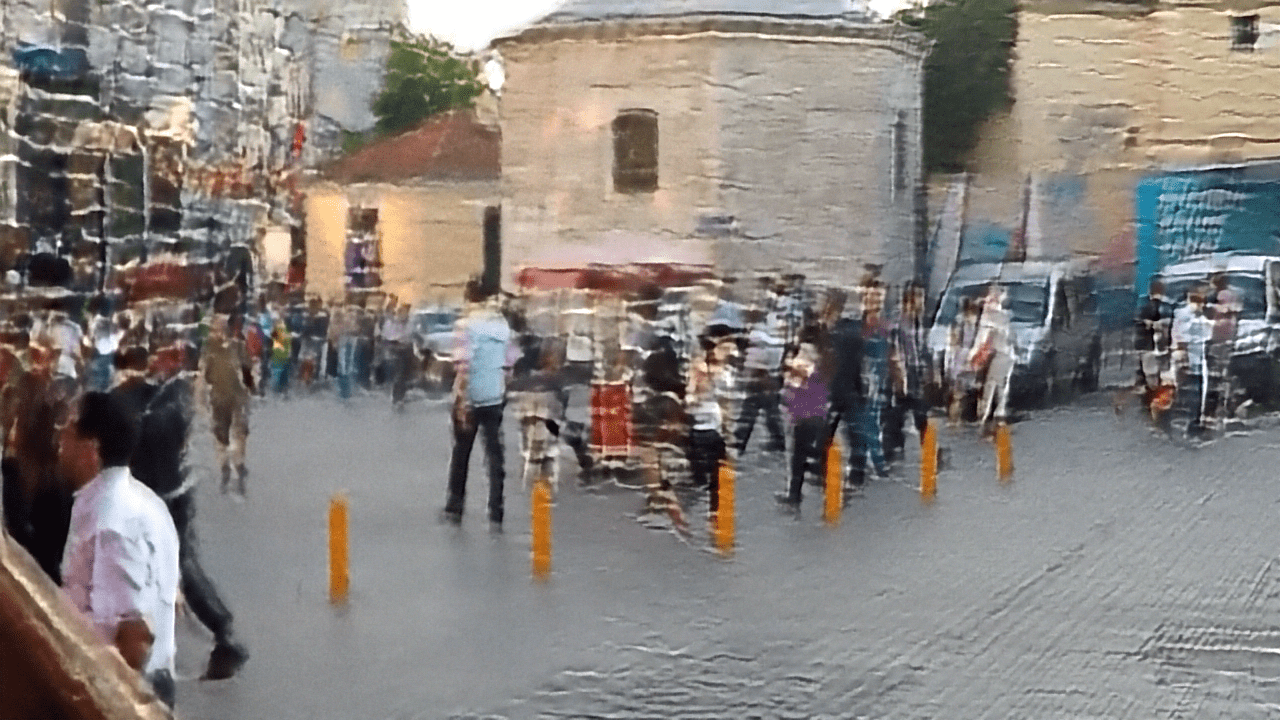}&
  \includegraphics[width=0.215\textwidth]{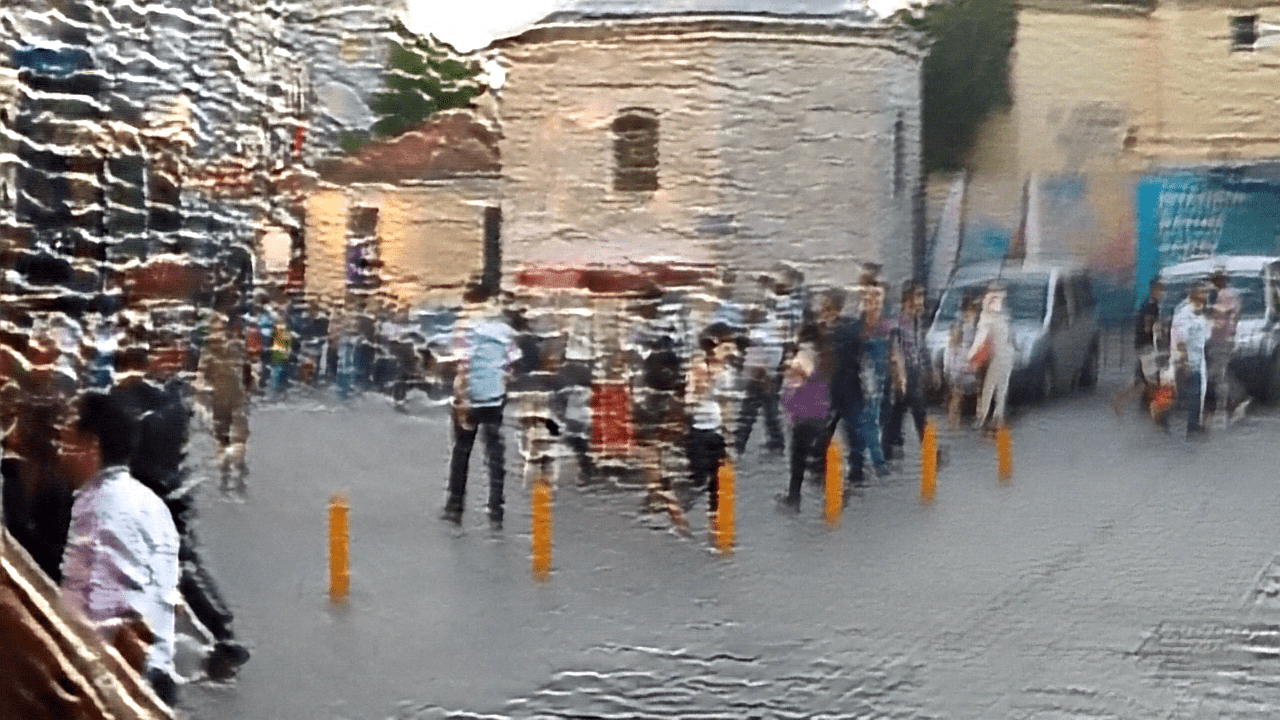}&
  \includegraphics[width=0.215\textwidth]{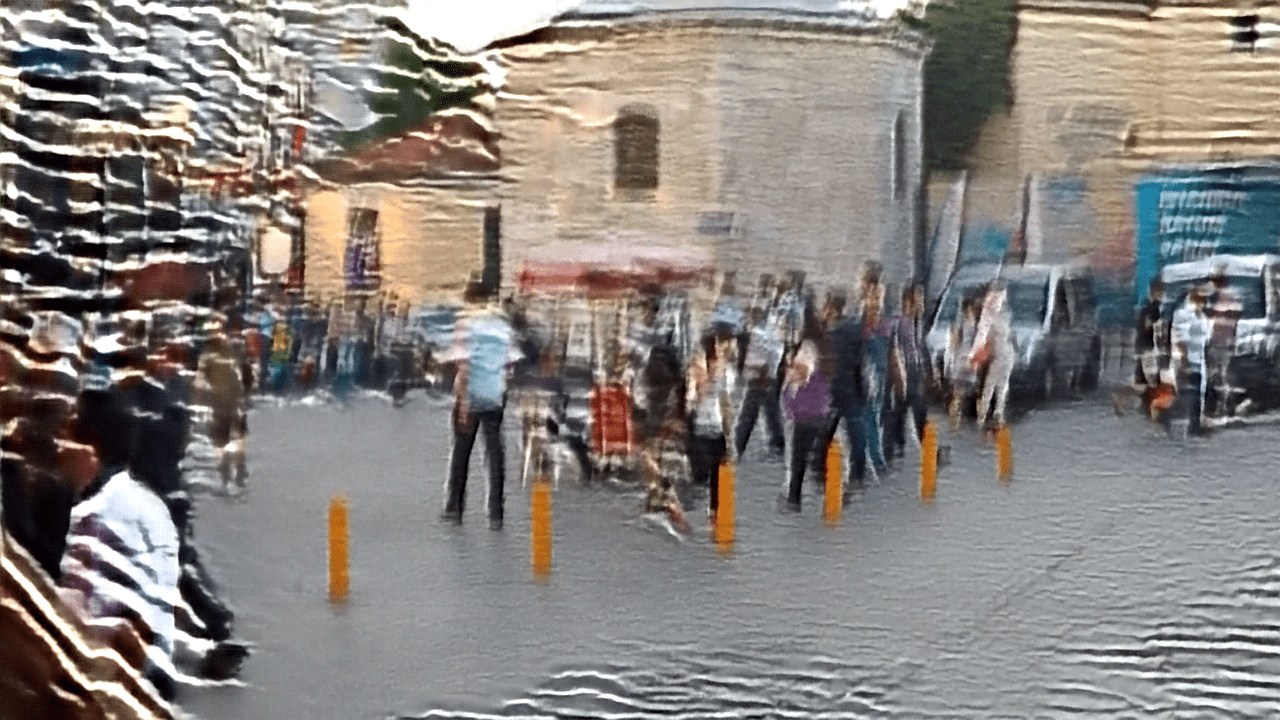}\\
  
\end{tabular}
}
\caption{Images reconstructed by different models after \textbf{CosPGD attack}. See Figure~\ref{fig:cospgd_attack} (Appendix~\ref{appendix:results}) to compare over all considered models.}
\label{fig:mini_cospgd_attack}
\end{figure*}

At first, one might overlook this shortcoming, however, when considering safety-critical  real-world applications like those in the medical domain for deblurring MRI images, or in autonomous driving, such shortcomings could be very hazardous. 
This is further highlighted in Figure~\ref{fig:mini_cospgd_attack} as we observe that both the Restormer and the Baseline network introduce ringing artifacts in the reconstructed images, however, NAFNet introduces very strong aliasing and color mixing that gets worse as the attack strength increases.
While aliasing and color artifacts are significantly reduced with adversarial training (please refer to Figure~\ref{fig:mini_cospgd_attack}), the reconstructions of NAFNet and the Baseline  network are still affected by residual ringing artifacts.
Interestingly, the quality of images reconstructed by Restormer after adversarial training is significantly better, as indicated by its performance in terms of PSNR and SSIM in  Table~\ref{tab:adv_perf}. 
At a low amount of adversarial attack iterations, the artifacts present in the images reconstructed by Restormer are negligible.
To ascertain that these observations are not specific to the adversarial attack itself, we visualize the images reconstructed after PGD attack in Figure~\ref{fig:pgd_attack} and observe a similar phenomenon.  
This accentuates the strength of the architectural design of Restormer and casts doubts over that of the networks proposed by \cite{chen2022simple}.

\section{Analysis and Discussion}
\label{sec:discussion}
Following we discuss the design choices made in NAFNet and the Baseline network that constrain the performance of the network against adversarial attacks, despite employing adequate defense techniques.

\subsection{Analyzing Intermediate networks}
\label{subsec:discuss:intermediate}
First, we study the \emph{Intermediate network} to ascertain if the spectral artifacts introduced by NAFNet in its reconstructed images were due to replacing a non-linear activation function with a \emph{Simple Gate}.
This is because the channel-wise multiplication would best explain the color mixing artifact and the inherent wrong subsampling during this operation and would account for the accentuated aliasing artifacts. 
Further to understand the influence of the non-linear activation, we also train the  Intermediate network with ReLU activation, referred to as \emph{Intermediate + ReLU}.

We report the findings on the Intermediate networks in Table~\ref{tab:intermediate}.
Here we observe that the Intermediate network performs marginally worse than even NAFNet, especially under adversarial attacks. 
Additionally, in Figure~\ref{fig:cospgd_attack}, we visualize the images reconstructed by the Intermediate network.
Firstly, the clean images~(unperturbed) have not been deblurred significantly.
Secondly, even under mild adversarial attacks, the quality of the reconstructed images is abysmal.
We observe severe checkboard patterns, aliasing, and color mixing in all images reconstructed by the Intermediate network under adversarial attack.
Thus, to better understand the performance of the Intermediate network in comparison to the Baseline network and NAFNet, we perform significantly weaker adversarial attacks. 
To this effect, we use the CosPGD attack but with $\epsilon\approx\frac{2}{255}$, and consider attack iterations $\in$ \{1, 3, 5\}.
We again use $\alpha=0.01$.

\begin{figure}[htb]
    \centering 
    \scalebox{0.83}{
    \begin{tabular}{@{}c@{\hspace{0.1cm}}c@{\hspace{0.1cm}}c@{}c@{\hspace{0.1cm}}c@{\hspace{0.1cm}}c@{}}
    \multicolumn{3}{c}{\footnotesize \textbf{MODEL}}  & 1 iterations & 3 iterations & 5 iterations \\  
  \rotatebox{90}{\textbf{Baseline}} &  & &
  \includegraphics[width=0.32\linewidth]{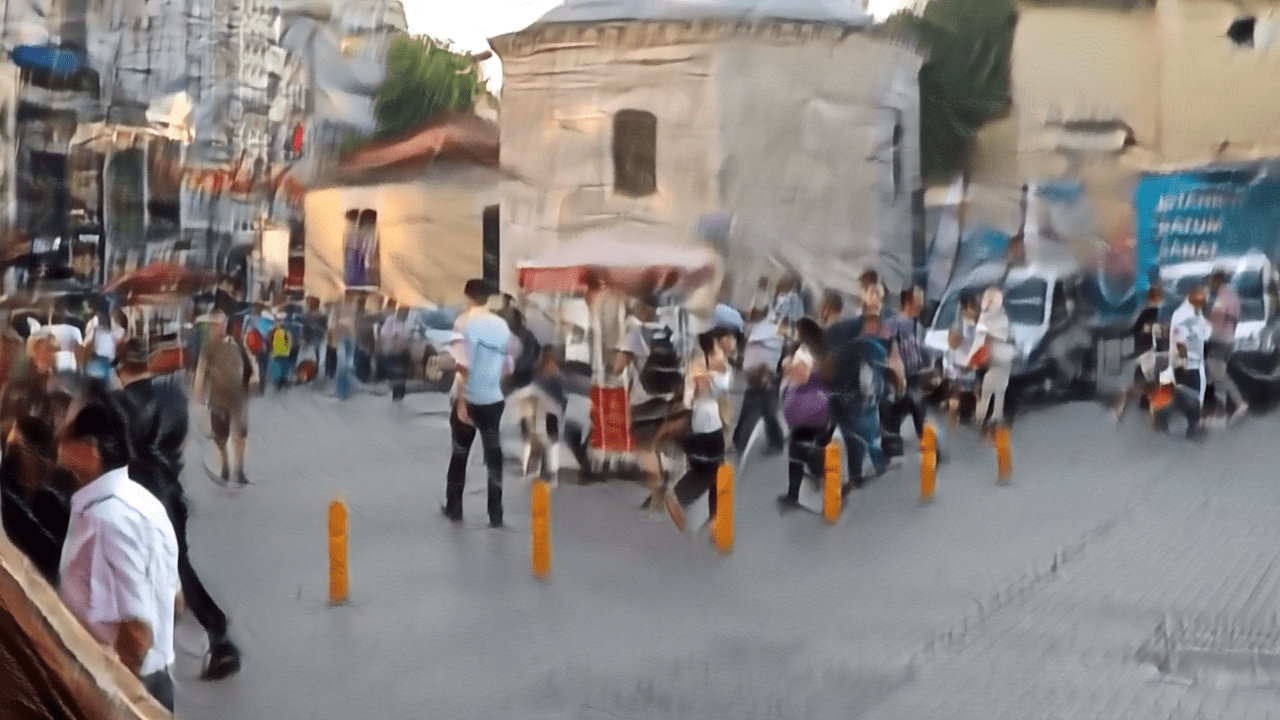}&
  \includegraphics[width=0.32\linewidth]{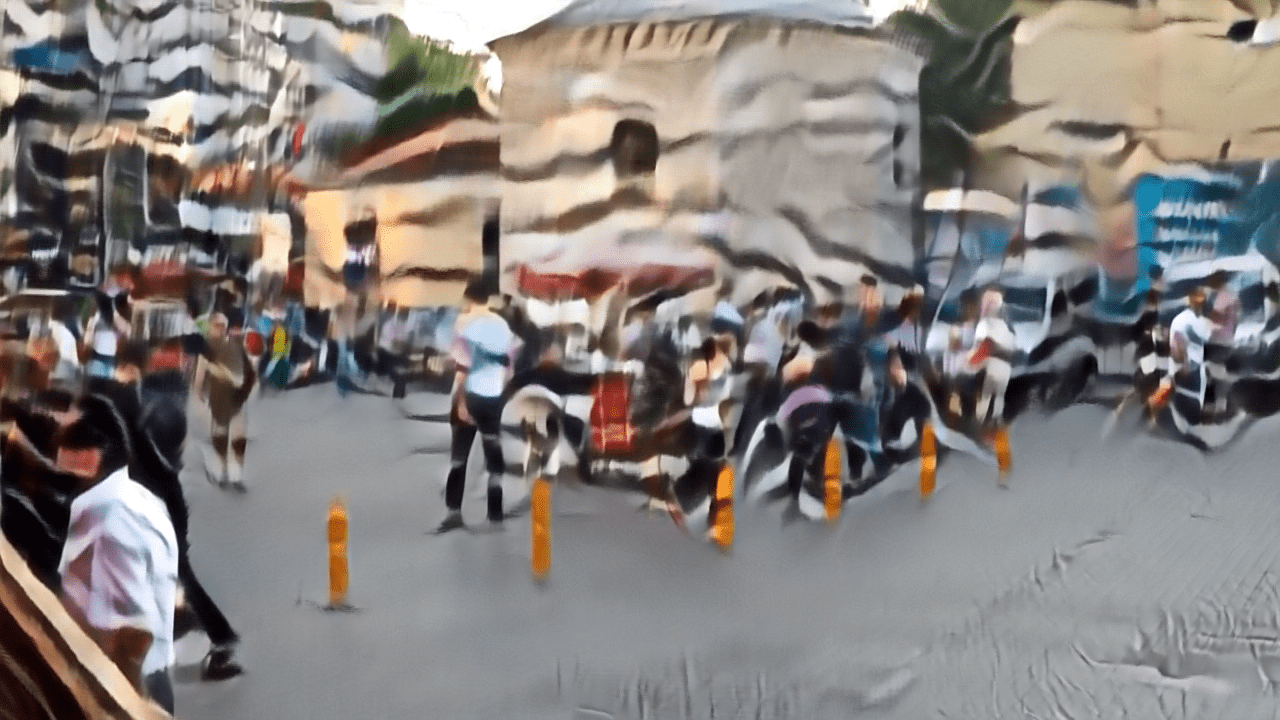}&
  \includegraphics[width=0.32\linewidth]{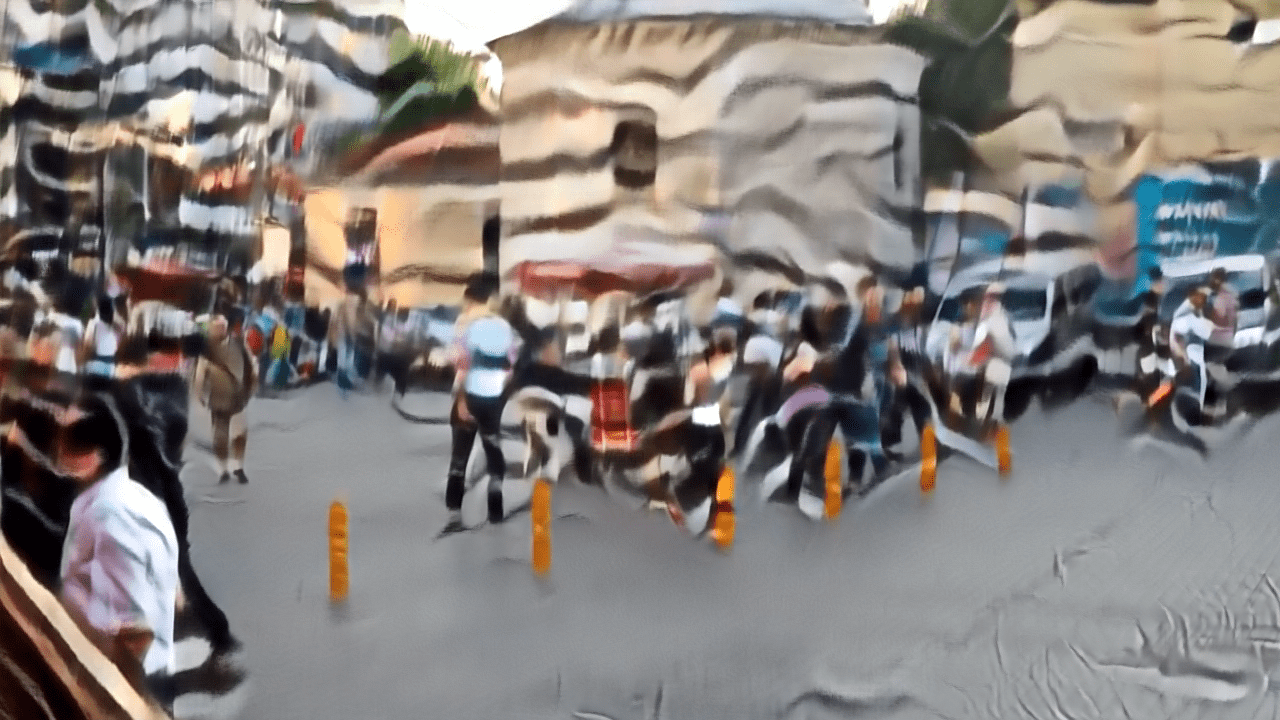}\\
  
  \rotatebox{90}{\textbf{NAFNet}} &   & &
  \includegraphics[width=0.32\linewidth]{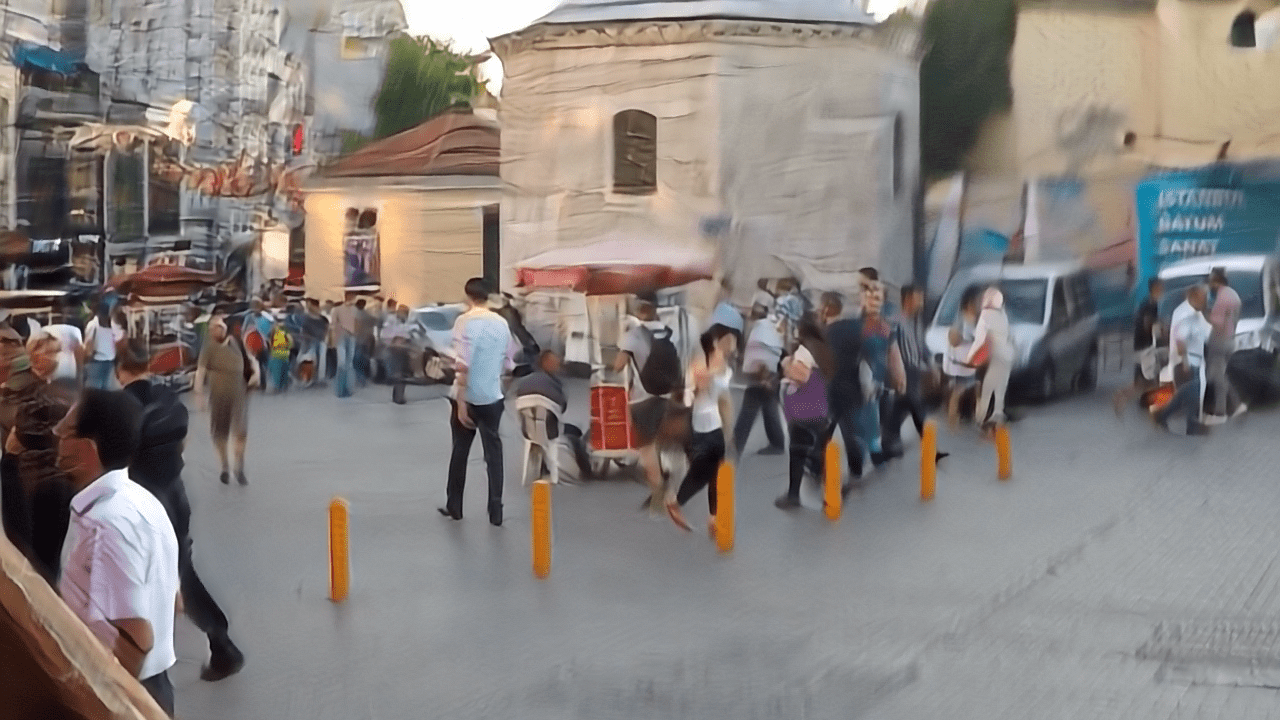}&
  \includegraphics[width=0.32\linewidth]{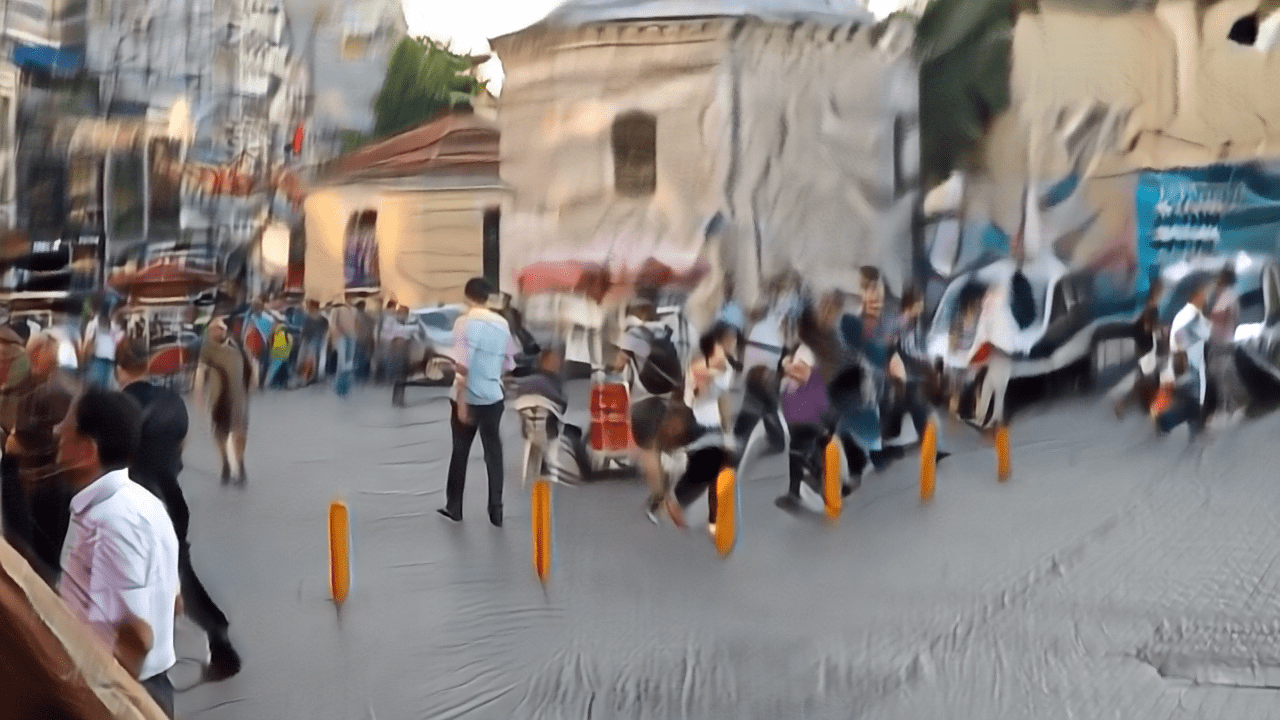}&
  \includegraphics[width=0.32\linewidth]{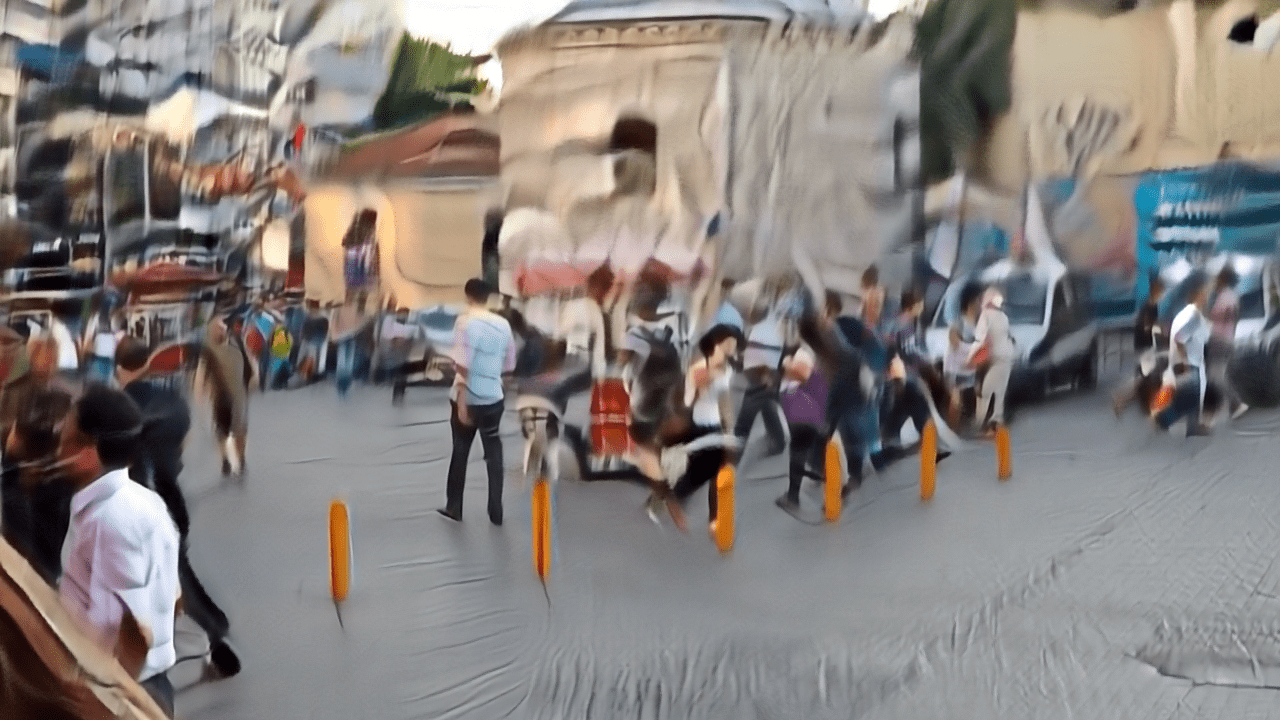}\\

  \rotatebox{90}{\footnotesize \textbf{Intermediate}} &  & &
  \includegraphics[width=0.32\linewidth]{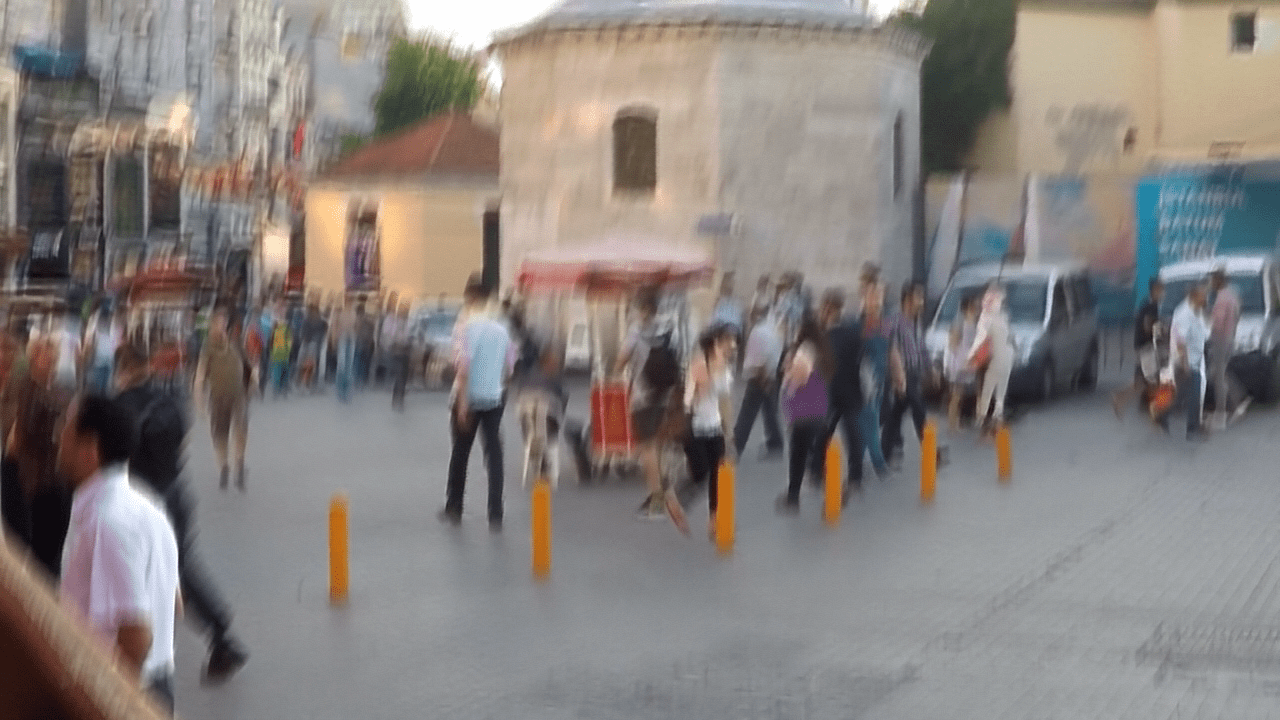}&
  \includegraphics[width=0.32\linewidth]{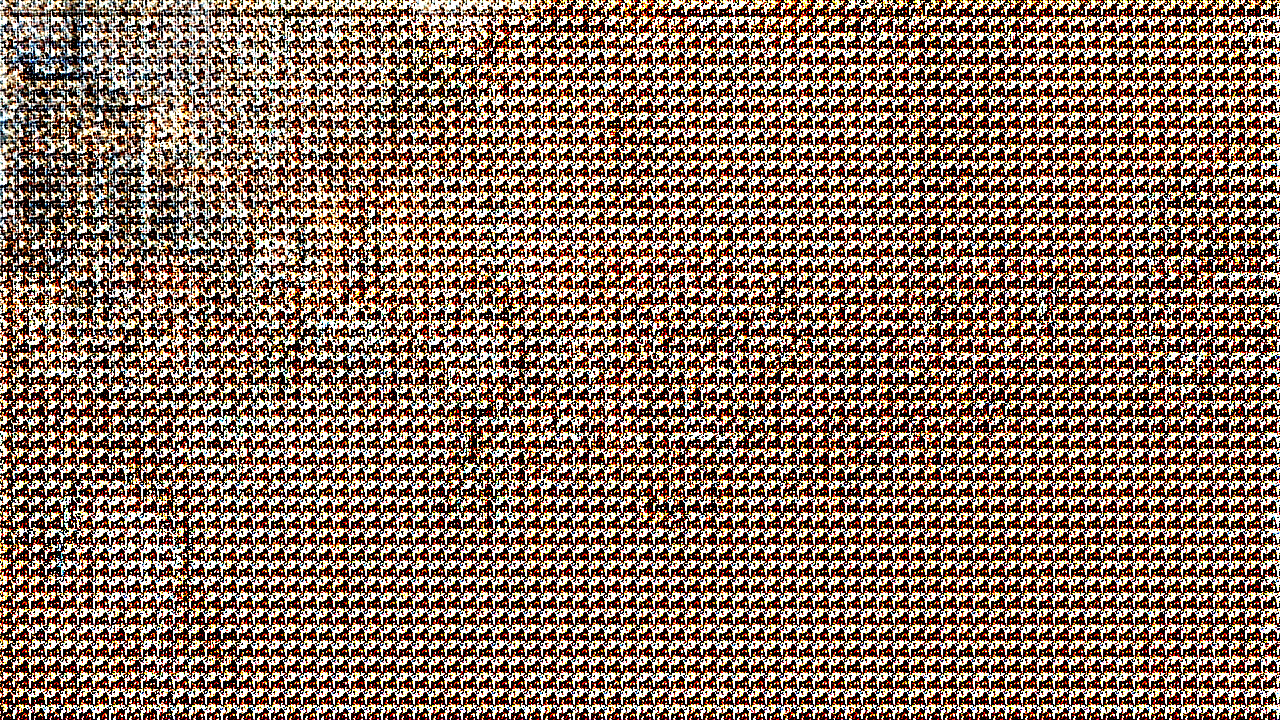}&
  \includegraphics[width=0.32\linewidth]{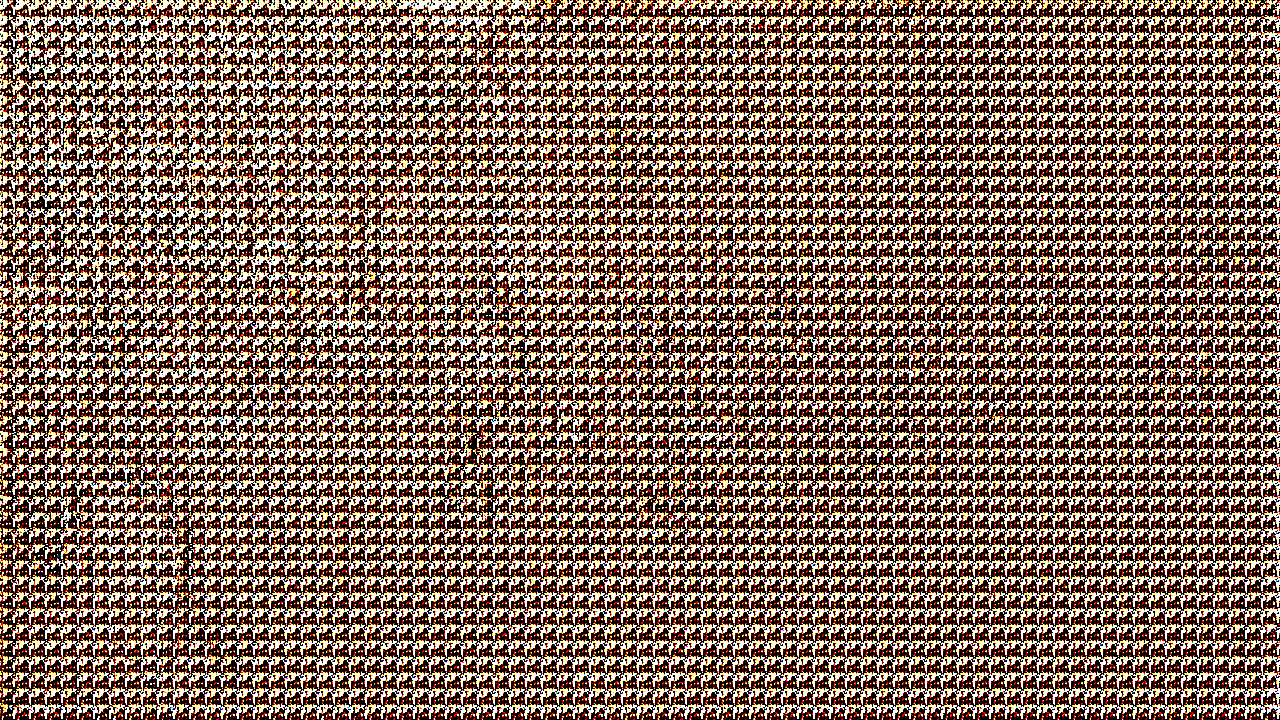}\\

  \rotatebox{90}{\footnotesize \textbf{Intermediate}} & \rotatebox{90}{~ \textbf{+ ADV}} & &
  \includegraphics[width=0.32\linewidth]{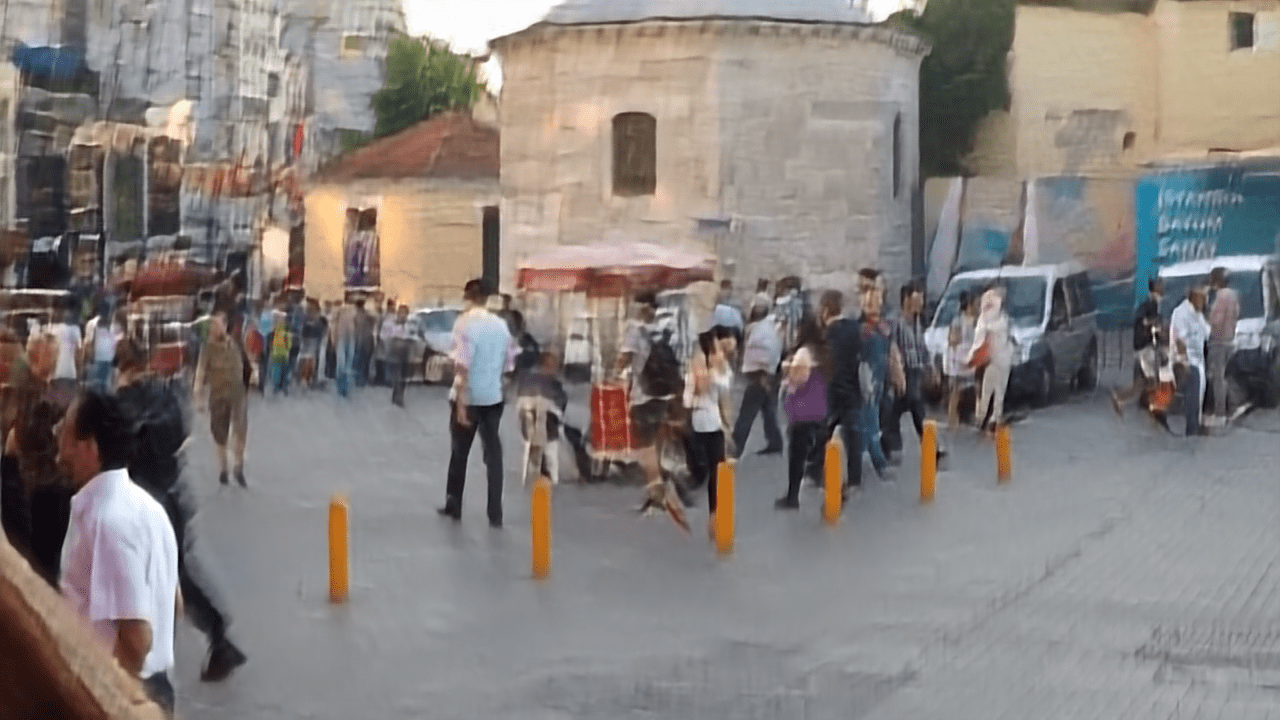}&
  \includegraphics[width=0.32\linewidth]{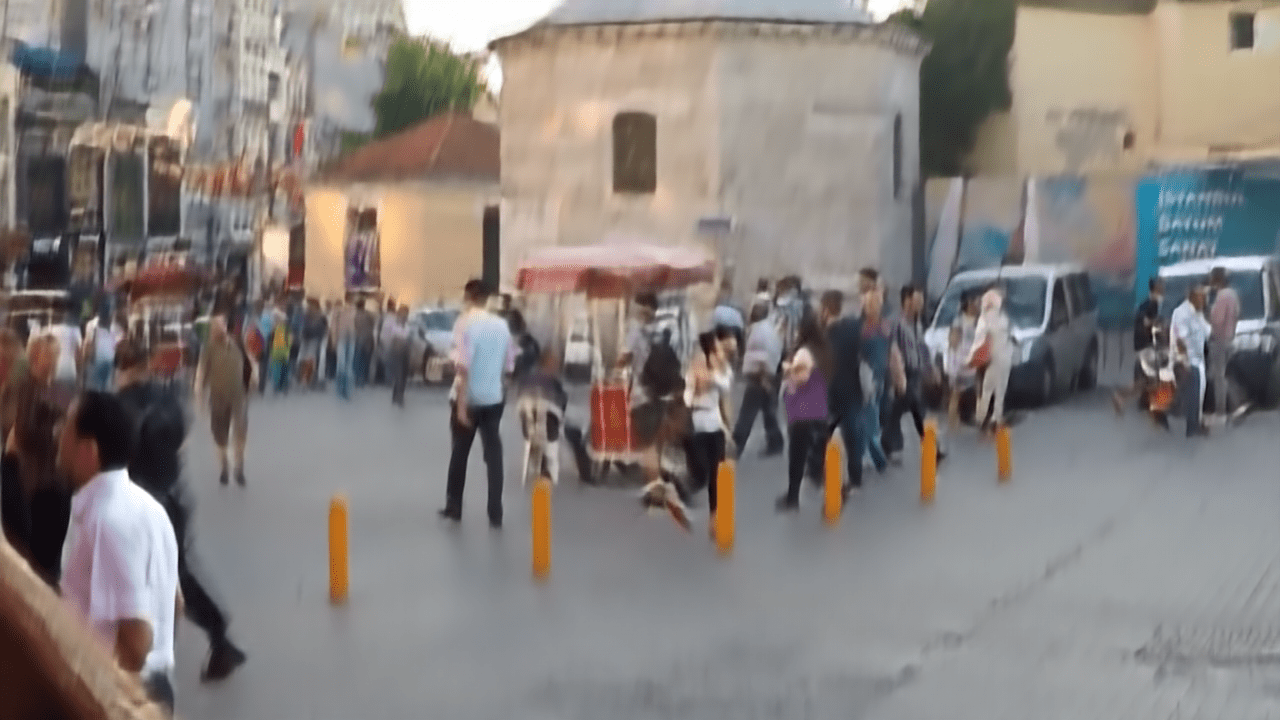}&
  \includegraphics[width=0.32\linewidth]{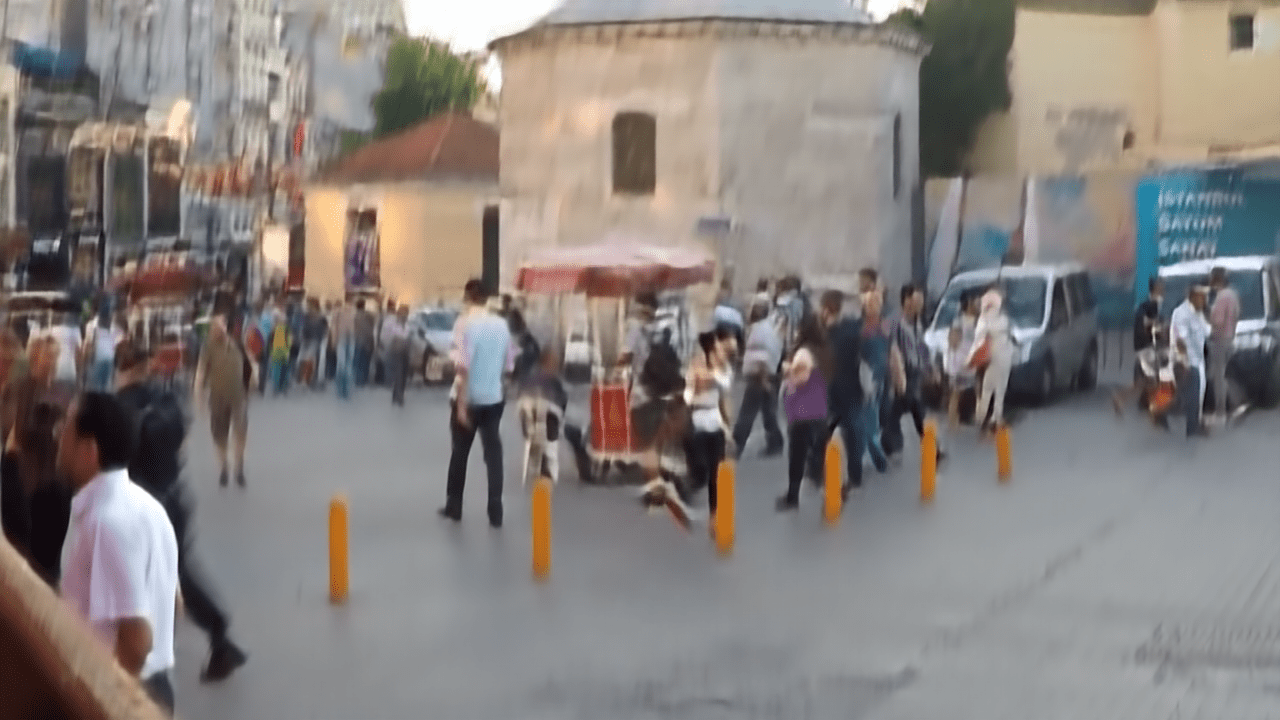}\\

  \rotatebox{90}{\footnotesize \textbf{Intermediate}} & \rotatebox{90}{~ \textbf{+ ReLU}} & &
  \includegraphics[width=0.32\linewidth]{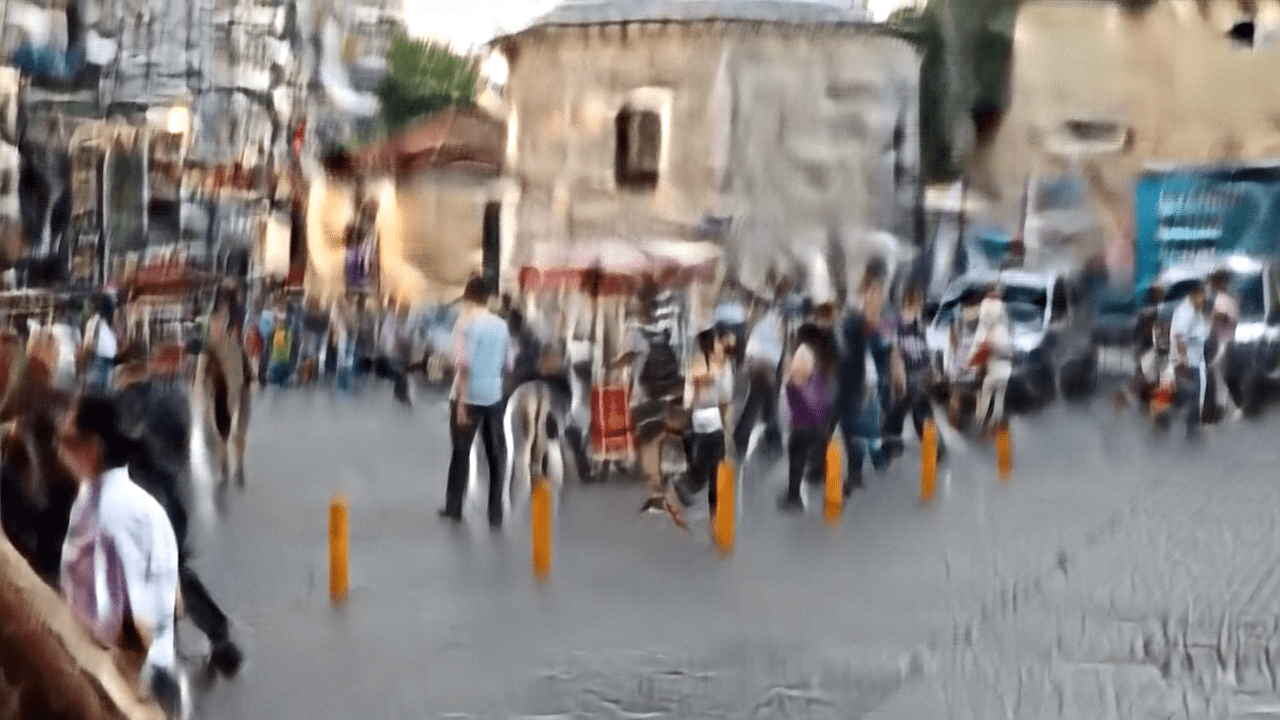}&
  \includegraphics[width=0.32\linewidth]{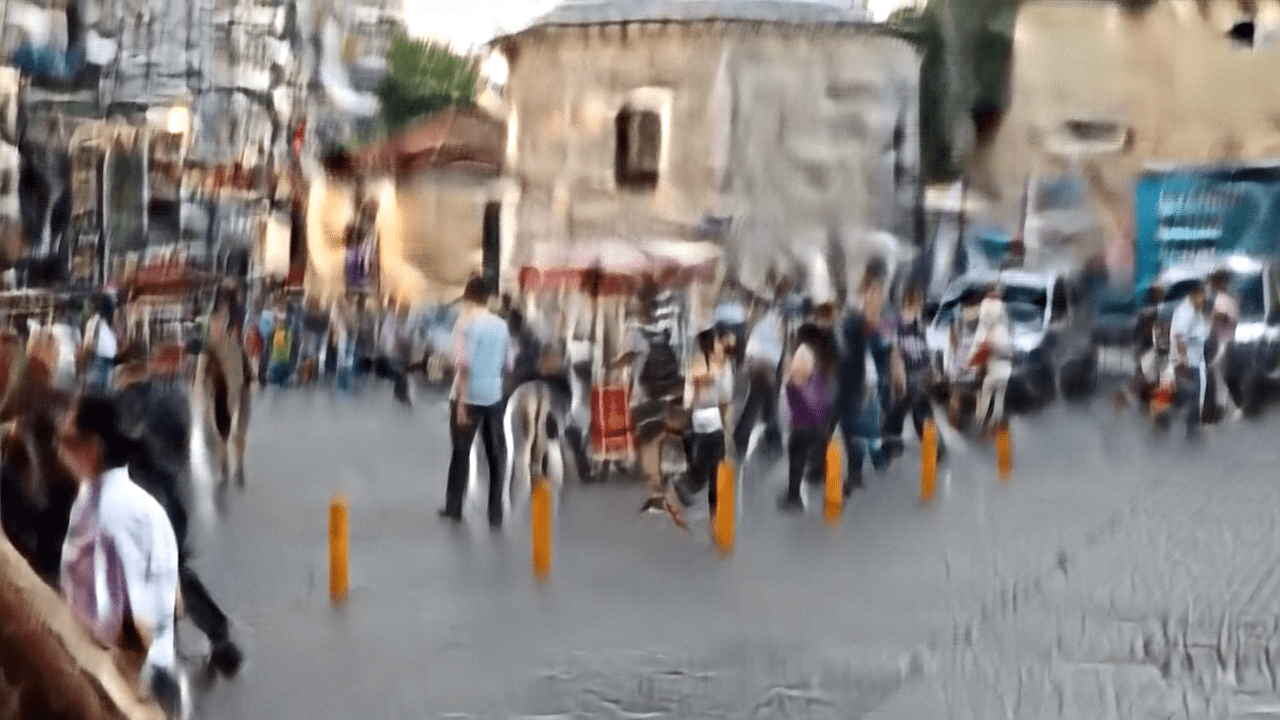}&
  \includegraphics[width=0.32\linewidth]{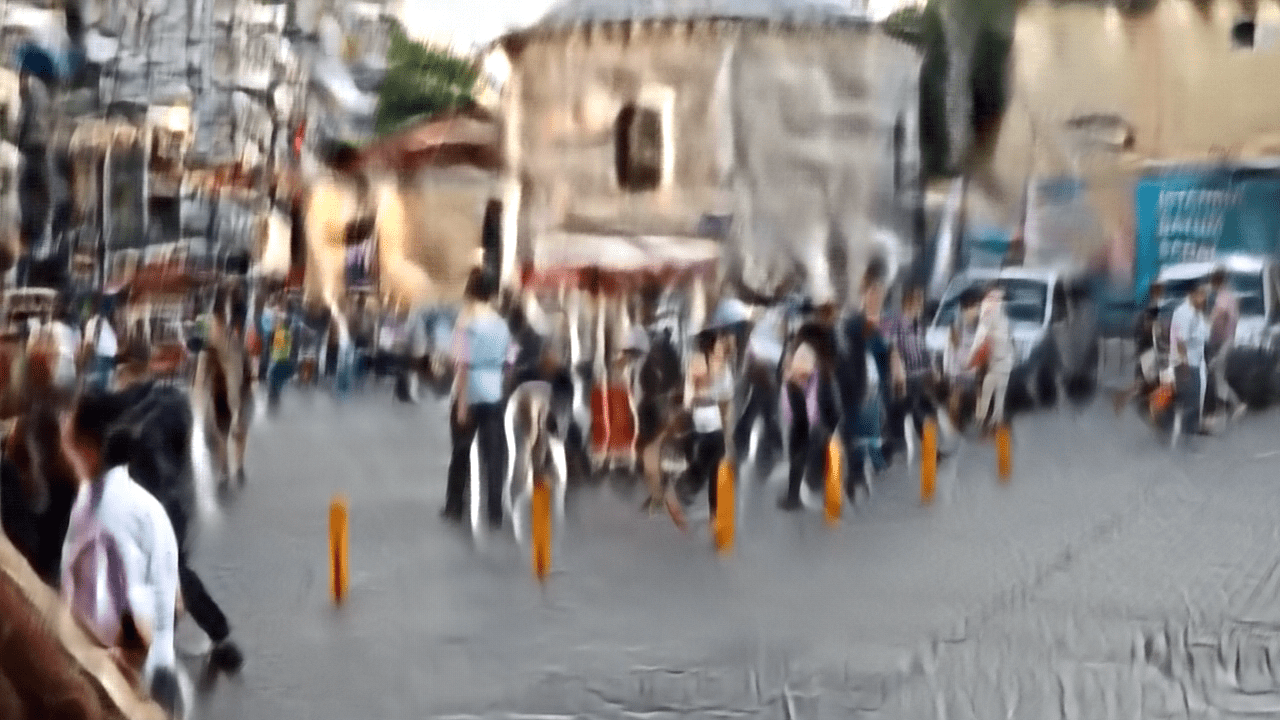}\\

  \rotatebox{90}{\footnotesize \textbf{Intermediate}} & \rotatebox{90}{~ \textbf{+ ReLU}} & \rotatebox{90}{~ \textbf{+ ADV}} &
  \includegraphics[width=0.32\linewidth]{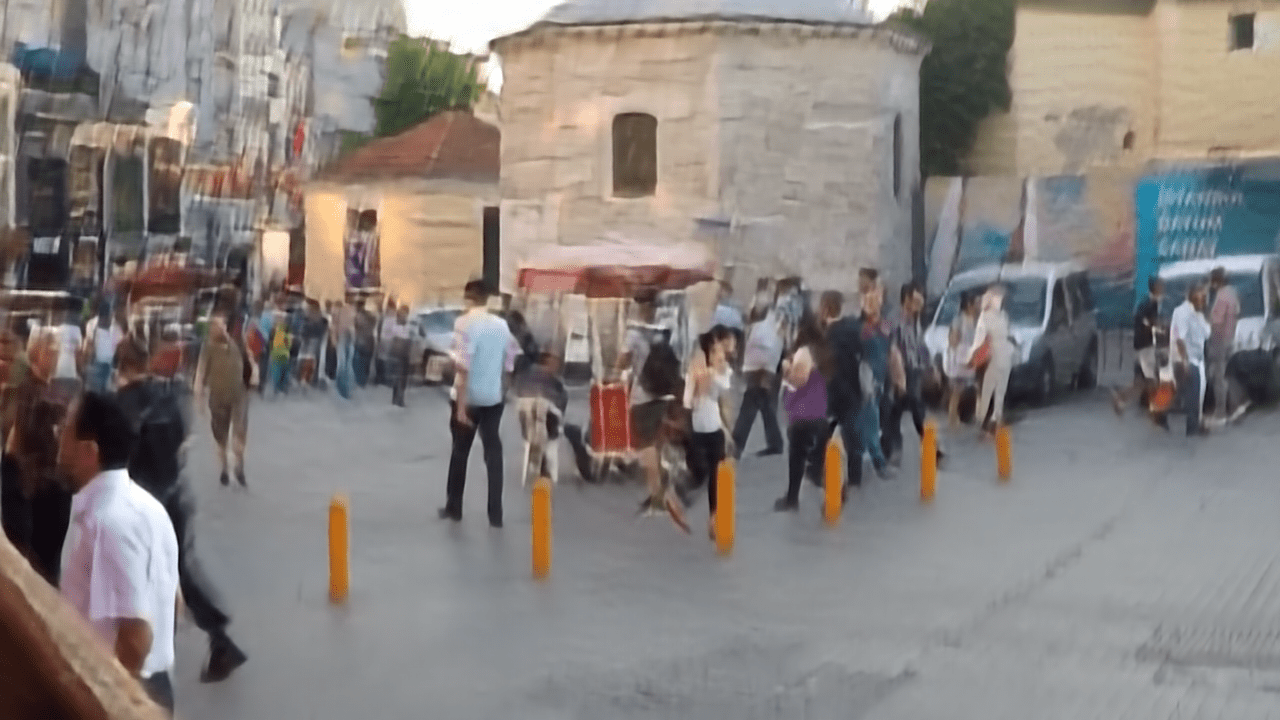}&
  \includegraphics[width=0.32\linewidth]{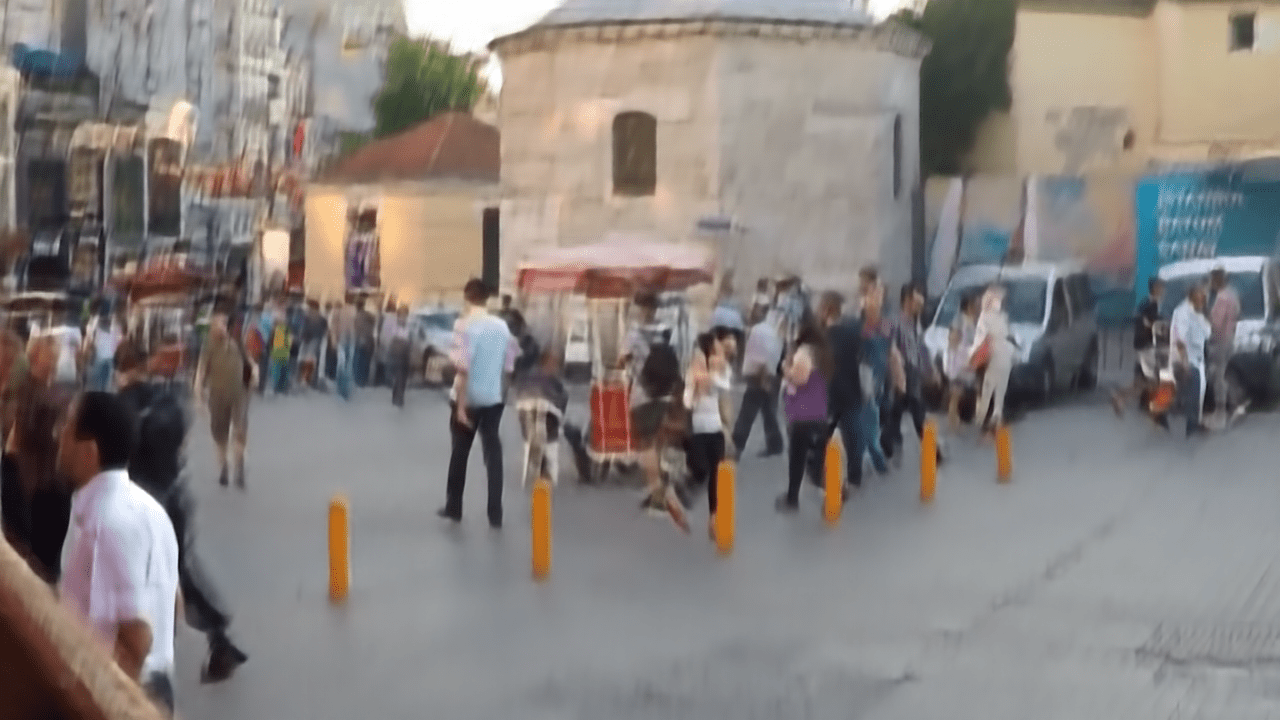}&
  \includegraphics[width=0.32\linewidth]{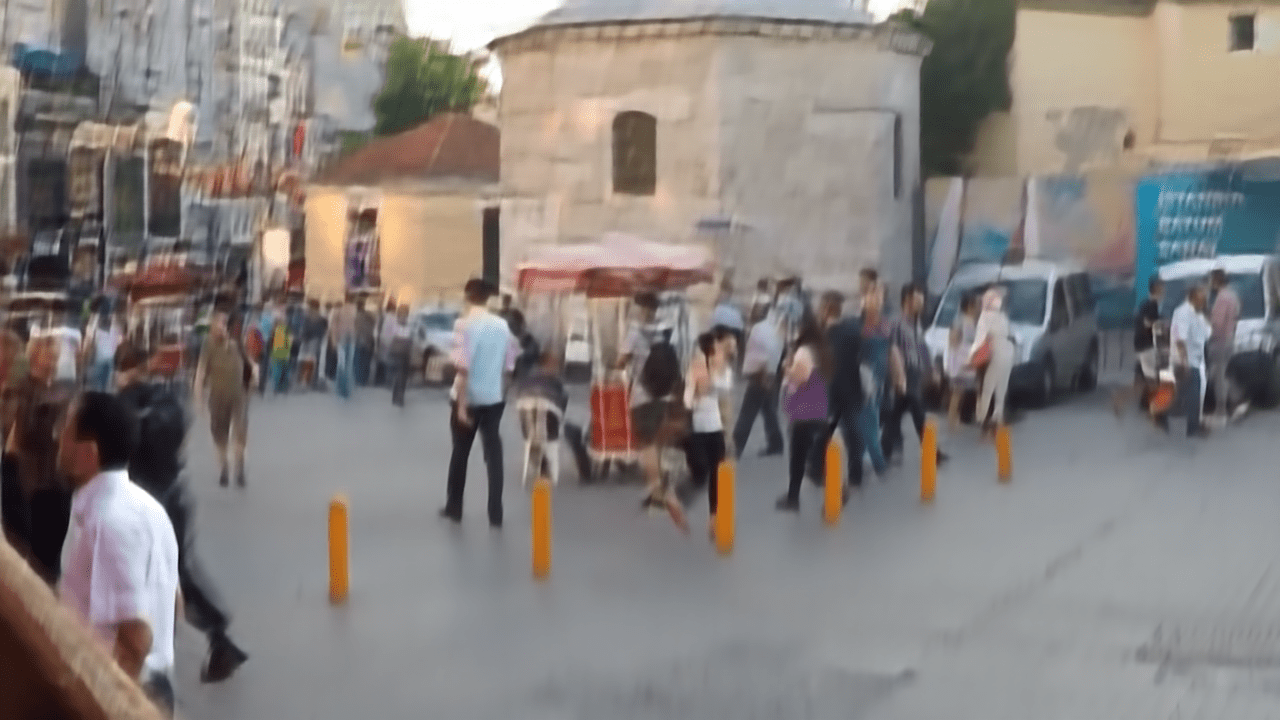}\\

\end{tabular}
}
\caption{Comparing images reconstructed by different models after \textbf{CosPGD attack} at $\epsilon\approx\frac{2}{255}$. 
Thus the attack strength is significantly weaker.}
\label{fig:weak_cospgd}
\end{figure}

\begin{table}[h]
    \centering
    \caption{Comparison of performance of the Baseline network, NAFNet, and Intermediate networks against significantly weak CosPGD attack. For this comparison we use $\epsilon\approx\frac{2}{255}$ and $\alpha=0.01$ and consider fewer attack steps i.e.    iterations $\in$\{1, 3, 5\} }
    \scalebox{0.73}{
    \begin{tabular}{@{}l|cc|cc|cc@{}}
    \toprule
    \multirow{3}{*}{Architecture} &  \multicolumn{6}{c}{CosPGD} \\
    & \multicolumn{2}{c|}{1 attack itrs} & \multicolumn{2}{c|}{3 attack itrs} & \multicolumn{2}{c}{5 attack itrs}\\
    & PSNR & SSIM & PSNR & SSIM & PSNR & SSIM \\
    \toprule
        Baseline & 21.38  & 0.7520  & 17.19 & 0.6356  & 16.99  &  0.6316  \\
        NAFNet & 22.54  & 0.7883 & 18.80 & 0.6948 & 18.46 &  0.6835  \\
        \midrule
        Intermediate & 25.14 & 0.8410 & 10.37 & 0.2940 & 8.56 & 0.1812  \\
        ~~~~~ + ADV & 25.47 & 0.8555 & 25.16 & 0.8501 & 25.32 & 0.8555 \\
        \midrule
        Intermediate + ReLU & 23.96 & 0.8112 & 20.96 & 0.7458 & 21.5777 & 0.7594  \\
        ~~~~~ + ADV & 26.11 & 0.8616 & 25.10 & 0.8459 & 24.86 & 0.8413 \\
    \bottomrule
    \end{tabular}    
    }
    \label{tab:lower_eps}
\end{table}
We report the performance of the Intermediate networks in Table~\ref{tab:lower_eps}. 
Interestingly, we observe that after one adversarial attack iteration, the Intermediate network is significantly outperforming both the Baseline network and NAFNet.
However, the Intermediate network is unable to retain this superior performance, and its performance significantly drops as we increase the attack strength~(attack iterations).
Additionally, in Figure~\ref{fig:weak_cospgd} we observe the introduction of the same spectral artifacts for the Intermediate network as those observed in Figure
\ref{fig:cospgd_attack} and Figure~\ref{fig:pgd_attack} (please refer to Section~\ref{appendix:results}).
The intensity of the spectral artifacts increases as we increase the attack strength.
This phenomenon is similar to the performance of NAFNet, which performs admirably on clean samples and under weak adversarial attacks but begins to perform significantly worse as the attack strength increases.
This indicates that even smoothed activation functions in the NAFNet architecture instead of Simple Gate produce strong spectral artifacts in the reconstructed images.

This is in striking contrast to using a non-smooth non-linear activation function, ReLU.
Interestingly, we observe that \emph{Intermediate+ReLU} is significantly more robust, and the degradation in its performance with attack strength is significantly lower than all considered networks, including Restormer. 
In Figures~\ref{fig:cospgd_attack},~\ref{fig:pgd_attack}~\&\ref{fig:weak_cospgd} we observe that the images reconstructed by \emph{Intermediate+ReLU}, while blurry, have significantly fewer artifacts for reasonable values of $\epsilon$.

\begin{figure}[htb]
    \centering 
    \begin{tabular}{@{}c@{\hspace{0.1cm}}c@{}}
    \includegraphics[width=0.49\linewidth]{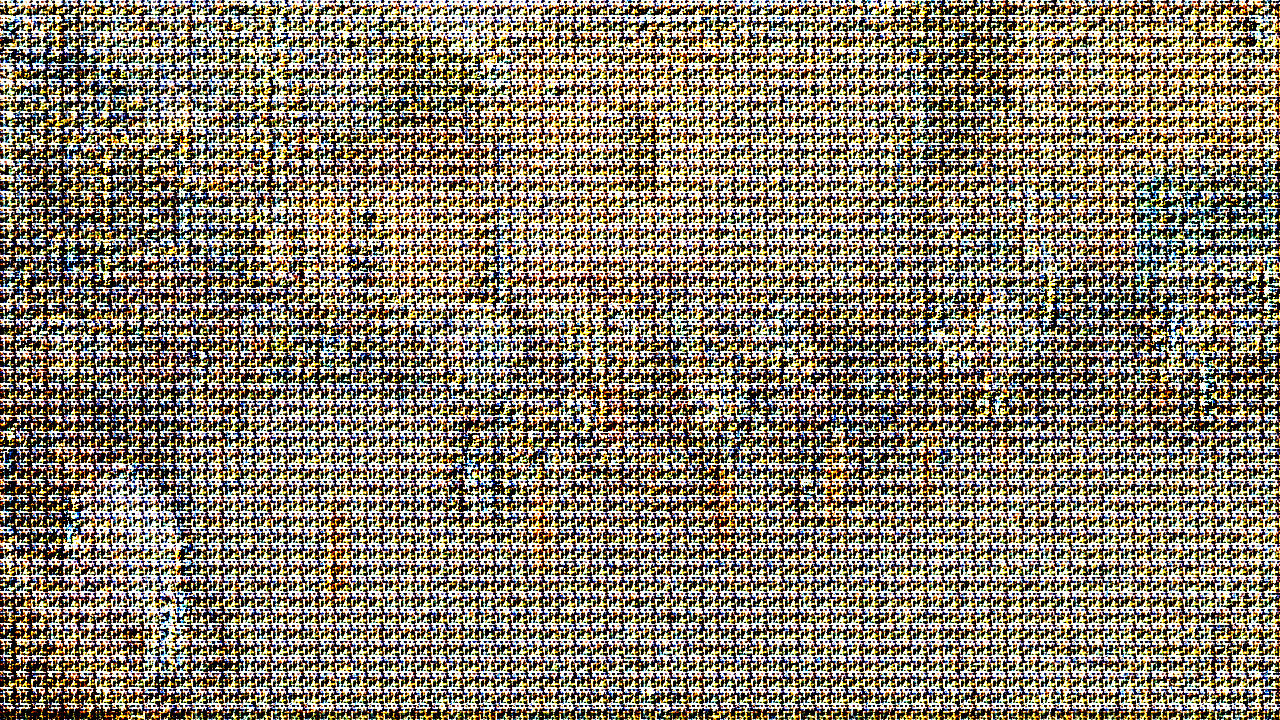} &
    \includegraphics[width=0.49\linewidth]{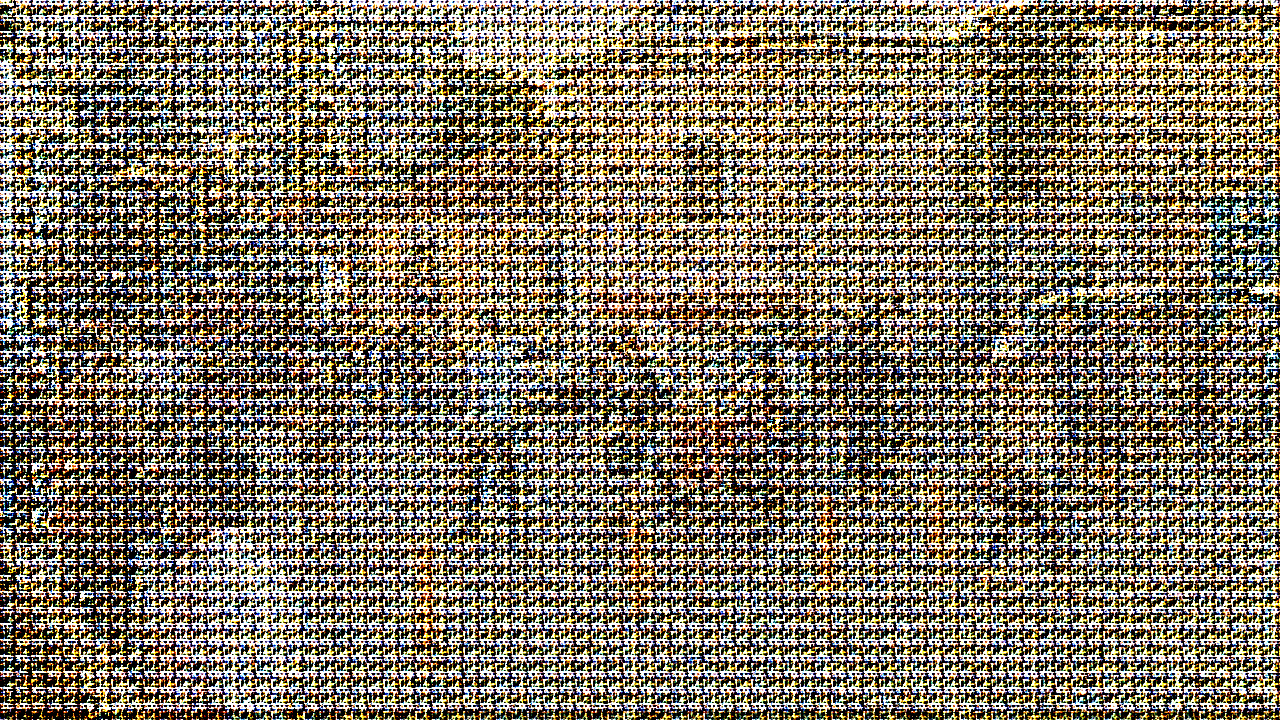} \\
\end{tabular}
\caption{Two different randomly chosen images reconstructed by \emph{Intermediate + ReLU} after 5 iterations of CosPGD attack with significantly higher $\epsilon\approx\frac{20}{255}$. We observe strong spectral artifacts similar to \emph{Intermediate network} in the recovered images.}
\label{fig:relu_higher_eps}
\end{figure}
Under adversarial attacks, the reconstructed images do not have spectral artifacts similar to \emph{Intermediate network} or NAFNet, but more similar to Restormer and the Baseline.
It is only at severely higher $\epsilon\approx\frac{20}{255}$ that spectral artifacts similar to those produced by \emph{Intermediate network} appear in the reconstructed images from \emph{Intermediate+ReLU}.
Thus, the smoothening of feature maps by the conjunction of Simplified Channel Attention and GeLU, and Simple Gate could be attributed to the introduction of some peculiar spectral artifacts and loss in robustness.
Using a non-smoothed non-linear activation function like ReLU appears to be an effective mitigation technique.

Additionally, as reported in Table~\ref{tab:adv_perf} we observe the adversarial robustness of both the \emph{Intermediate network} and \emph{Intermediate+ReLU} significantly increases after FGSM training, and is comparable to Restormer.
This significant improvement in adversarial performance is also visible at lower $\epsilon$ attacks, please refer to Table~\ref{tab:lower_eps} and visually shown in Figure~\ref{fig:weak_cospgd}.
Thus, as observed before, adversarial training is a fix to reduce artifacts, even for the \emph{Intermediate network}.

\subsection{Superiority of Restormer}
\label{subsec:discuss:restormer}
In their work, \cite{chen2022simple} attempt to reduce model complexity while retaining the performance of the Restormer.
However, as shown in our work this significantly degrades the generalization ability of the consequent models. 
As larger models tend to have a better trade-off between robustness and accuracy \cite{hendrycks2019benchmarking, hoffmann2021towards}, the reduced model capacity in the Baseline and NAFNet could contribute to the reduced robustness. 
While reducing model complexity is certainly important and desirable, to maintain robustness it requires a more careful and systematic pruning of networks \cite{ye2019adversarial,NEURIPS2020_e3a72c79,hoffmann2021towards} than simply dropping components. 
Apart from the model's complexity in terms of the number of parameters, the attention mechanism itself could be crucial for robustness.

While the Restormer uses a multi-headed self-attention mechanism, both the Baseline network and NAFNet use variants of channel-attention (NAFNet uses the simplified channel-attention proposed by \cite{chen2022simple}).
As shown by \cite{NEURIPS2021_e19347e1}, the self-attention module of vision Transformers significantly aids the Transformer based models to improve their robustness.
Additionally, it helps the model better utilize defense strategies such as additional training, distillation, etc.
A similar phenomenon is observed in Table~\ref{tab:adv_perf}, as Restormer, a vision transformer based model with a multi-headed self-attention module is able to better utilize adversarial training compared to the Baseline network and NAFNet.

\paragraph{Limitations. } Adversarial training and design choices like the use of smoothed or non-smoothed activation functions against using Simple Gates certainly have a significant impact on the performance of the considered image restoration models.
However, these still is a considerable gap in the clean performance of the considered models.
While the fixes work in increasing adversarial robustness and removal of spectral artifacts the images are far from ideal restoration.
As observed, the restored images after the fixes are significantly blurry.
This is a limitation of this work, as this work was focused on removal of spectral artifacts and better adversarial robustness.

This work is a step towards finding a fix and not an absolute fix.
Exploring methods other than adversarial training for increasing adversarial robustness and removal of spectral artifacts could be an interesting future work direction.

\section{Conclusion}
\label{sec:conclusion}
We raise concerns and awareness regarding the generalization ability of deep learning models.
Despite recent methods outperforming baselines for various vision tasks, for a method to have a significant contribution to real-world applications, it must be reliable and robust.
Thus in this work, we highlight this shortcoming of recently proposed Transformer based image restoration models.
While the models proposed by \cite{chen2022simple} perform satisfactorily for image deblurring on non-perturbed samples, they fail to generalize when slight adversarial perturbations are added to the blurred images.
We acknowledge that the reduction in model complexity compared to Restormer is a step in the right direction, however, in this case, it comes at the expense of model robustness.
Therefore, we additionally employ adversarial training in an attempt to fix this shortcoming while also improving the quality of the reconstructed images.
We observe that adversarial training is able to reduce the spectral artifacts and also results in significant improvements in adversarial robustness of the image restoration models.
However, the extent of the improvement varied with the architectural design decisions.
Thus lastly, we investigate the design decisions that might lead to the occurrence of spectral artifacts and loss in robustness for the considered methods and find a an interesting ablation concerning the type of activation functions used when downsampling.

\clearpage
{\small
\bibliographystyle{ieee_fullname}
\bibliography{egbib}
}

\clearpage
\appendix
\clearpage
\setcounter{page}{1}
\setcounter{figure}{0}
\renewcommand{\thefigure}{A\arabic{figure}}
\setcounter{table}{0}
\renewcommand{\thetable}{A\arabic{table}}

{ \onecolumn
    \centering
    \Large
    \textbf{On the unreasonable vulnerability of transformers for image restoration \\
– and an easy fix} \\
    \vspace{0.5em}Supplementary Material \\
    \vspace{1.0em}
}
\begin{multicols}{2}
Following we provide additional visual and quantitative results.

\section{Additional Results}
\label{appendix:results}
\begin{figure*}[htb]
    \centering 
\scalebox{0.92}{
    \begin{tabular}{@{}c@{\hspace{0.1cm}}c@{\hspace{0.1cm}}c@{\hspace{0.1cm}}c@{\hspace{0.1cm}}c@{\hspace{0.1cm}}c@{}}
    \multicolumn{2}{c}{MODEL} & NO ATTACK & 5 iterations & 10 iterations & 20 iterations\\
  \rotatebox{90}{\textbf{Restormer}} & & 
  \includegraphics[width=0.23\textwidth]{restormer_no_attack_GOPR0384_11_00-000002.png}&
  \includegraphics[width=0.23\textwidth]{restormer_cospgd_5_GOPR0384_11_00-000002.png}&
  \includegraphics[width=0.23\textwidth]{restormer_cospgd_10_GOPR0384_11_00-000002.png}&
  \includegraphics[width=0.23\textwidth]{restormer_cospgd_20_GOPR0384_11_00-000002.png}\\
  
  \rotatebox{90}{\textbf{Baseline}} & & 
  \includegraphics[width=0.23\textwidth]{baseline_no_attack_GOPR0384_11_00-000002.png}&
  \includegraphics[width=0.23\textwidth]{baseline_cospgd_5_GOPR0384_11_00-000002.png}&
  \includegraphics[width=0.23\textwidth]{baseline_cospgd_10_GOPR0384_11_00-000002.png}&
  \includegraphics[width=0.23\textwidth]{baseline_cospgd_20_GOPR0384_11_00-000002.png}\\

  \rotatebox{90}{\textbf{Intermediate}} & & 
  \includegraphics[width=0.23\textwidth]{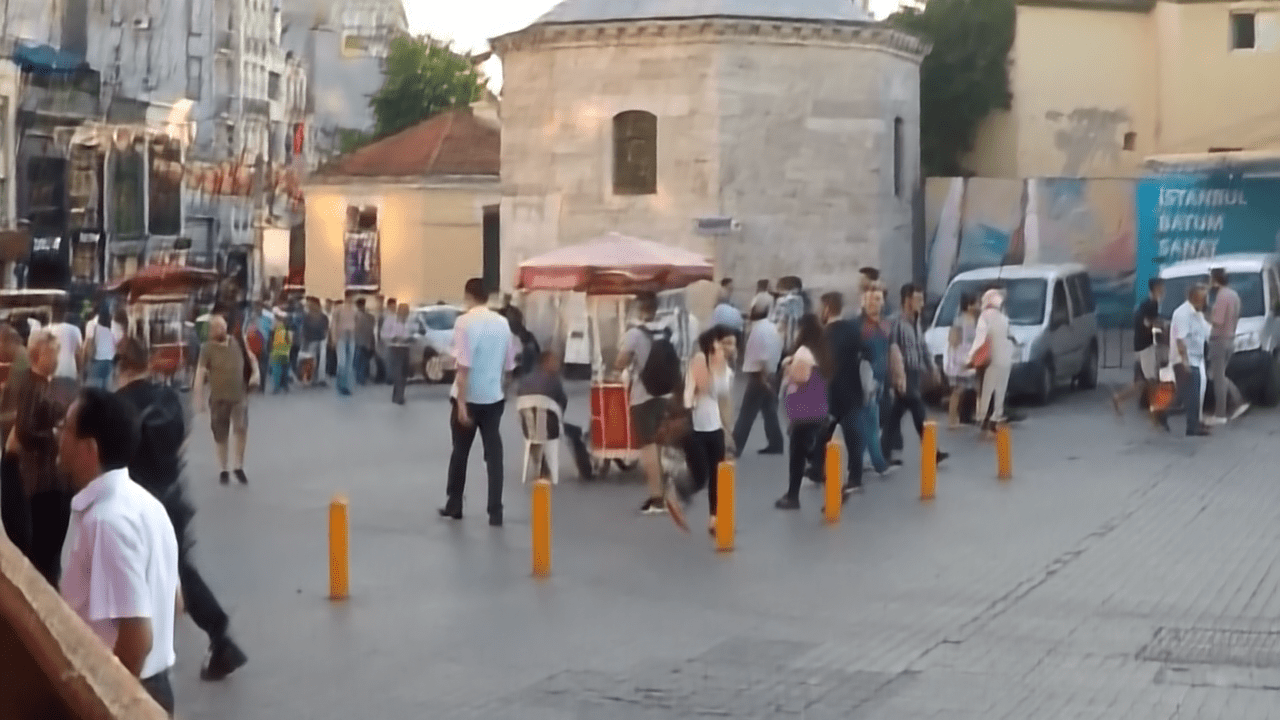}&
  \includegraphics[width=0.23\textwidth]{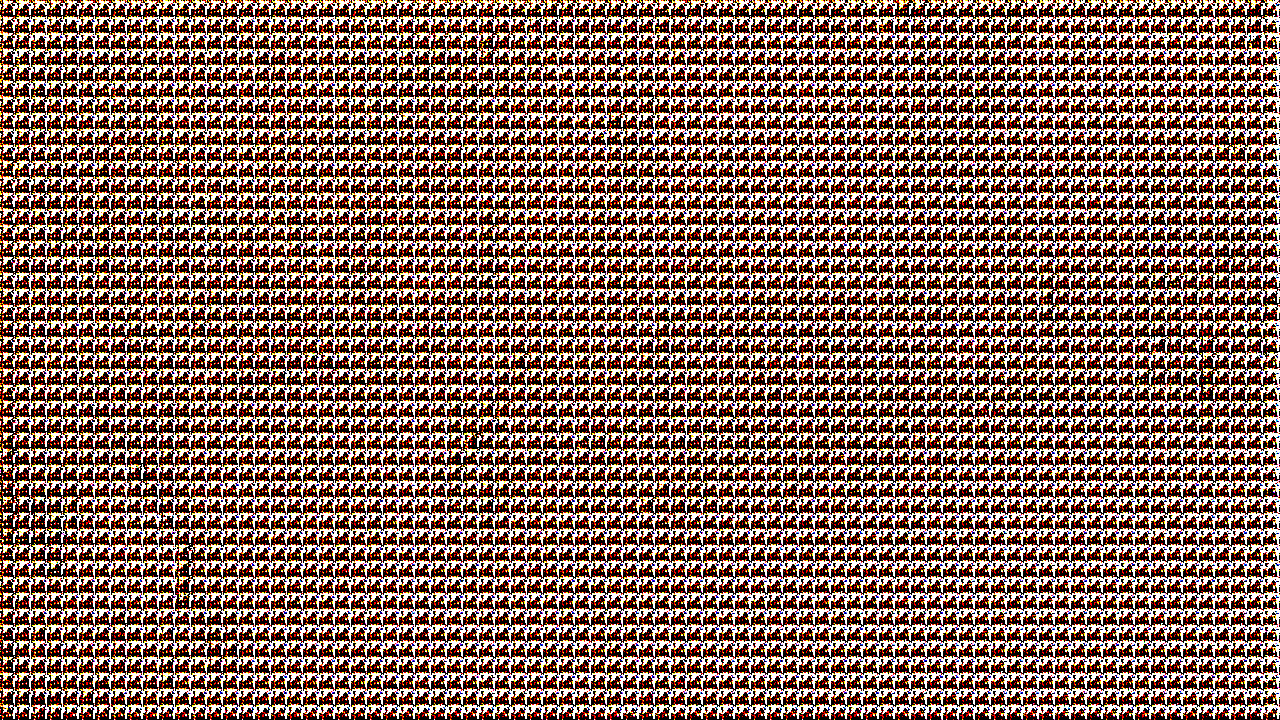}&
  \includegraphics[width=0.23\textwidth]{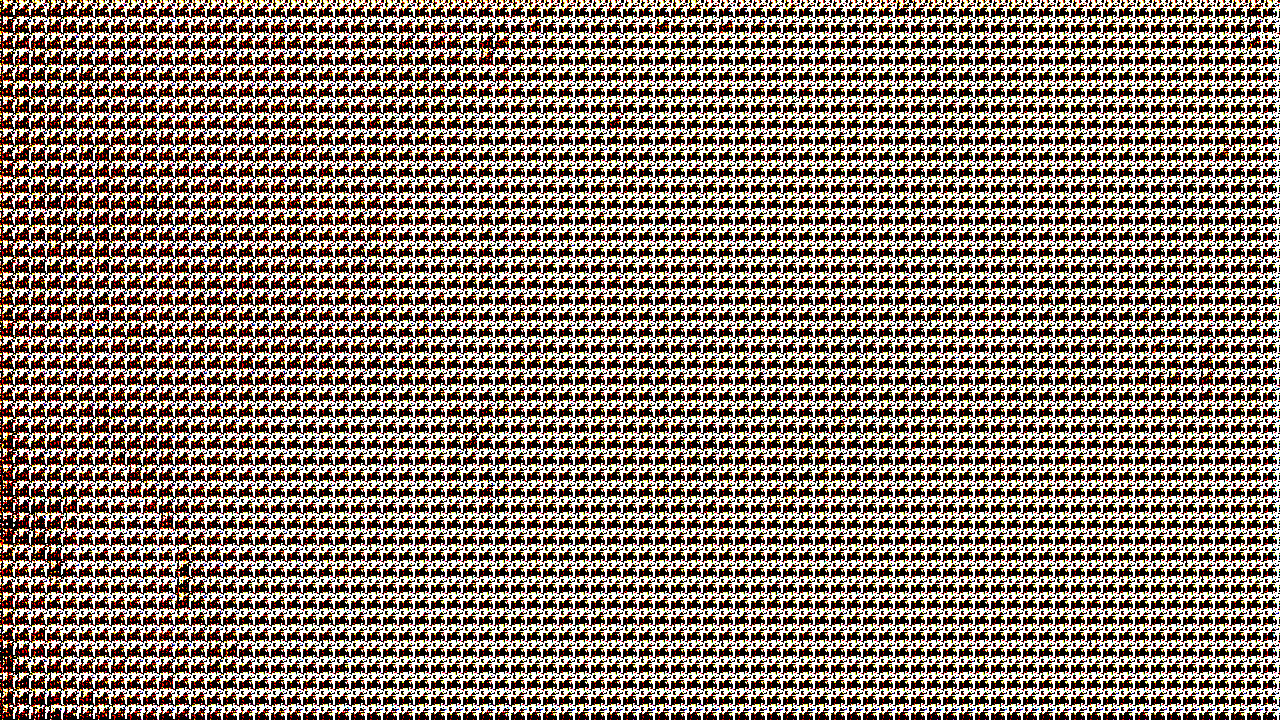}&
  \includegraphics[width=0.23\textwidth]{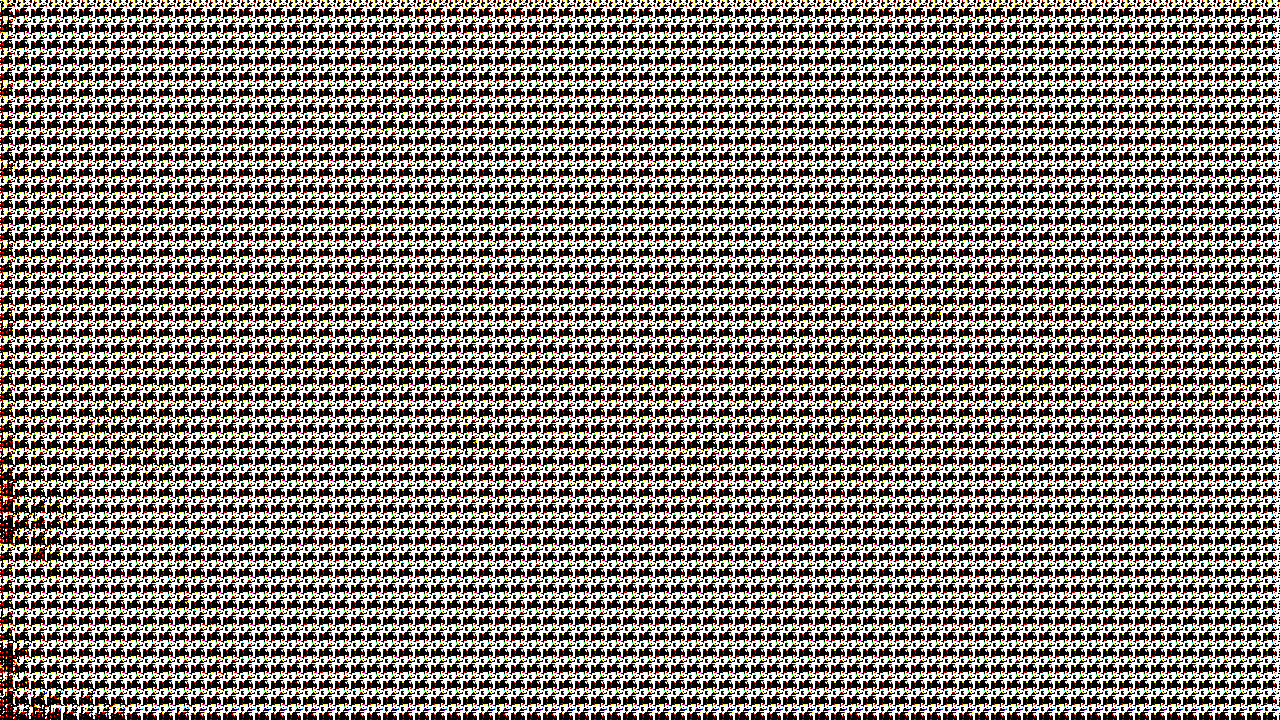}\\

    \rotatebox{90}{\textbf{Intermediate}} & \rotatebox{90}{~~\textbf{+ADV}} & 
  \includegraphics[width=0.23\textwidth]{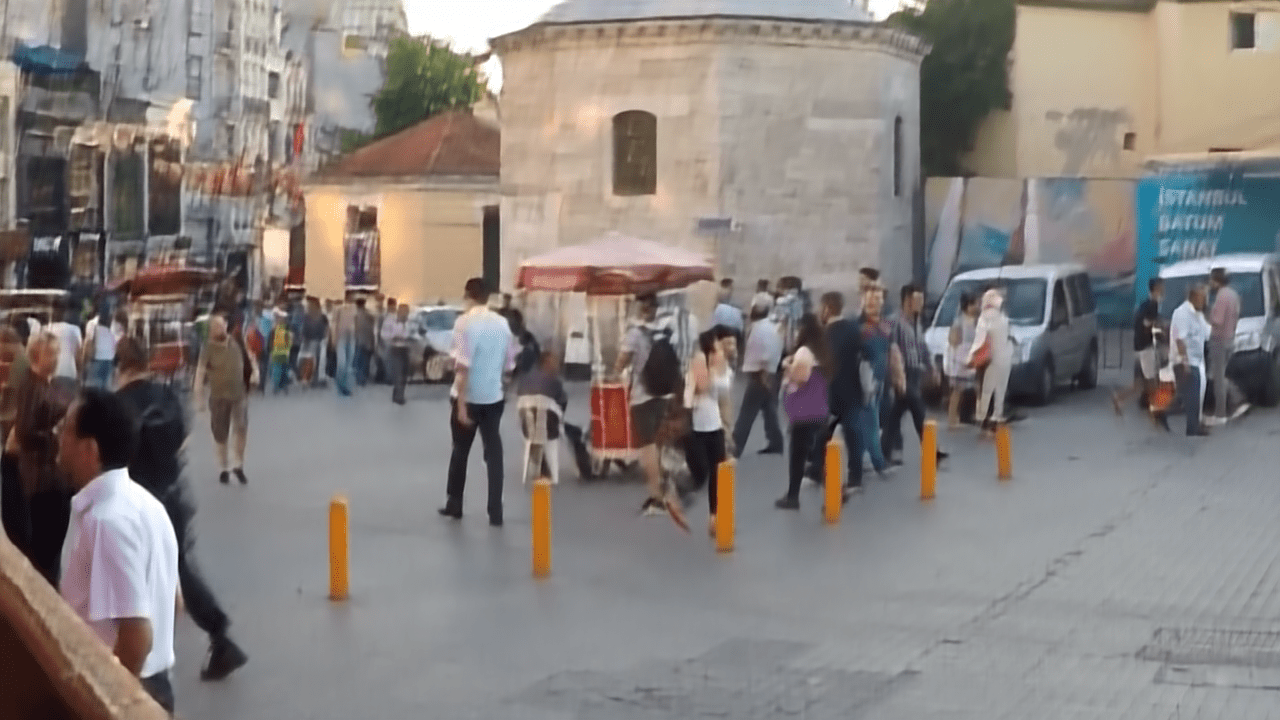}&
  \includegraphics[width=0.23\textwidth]{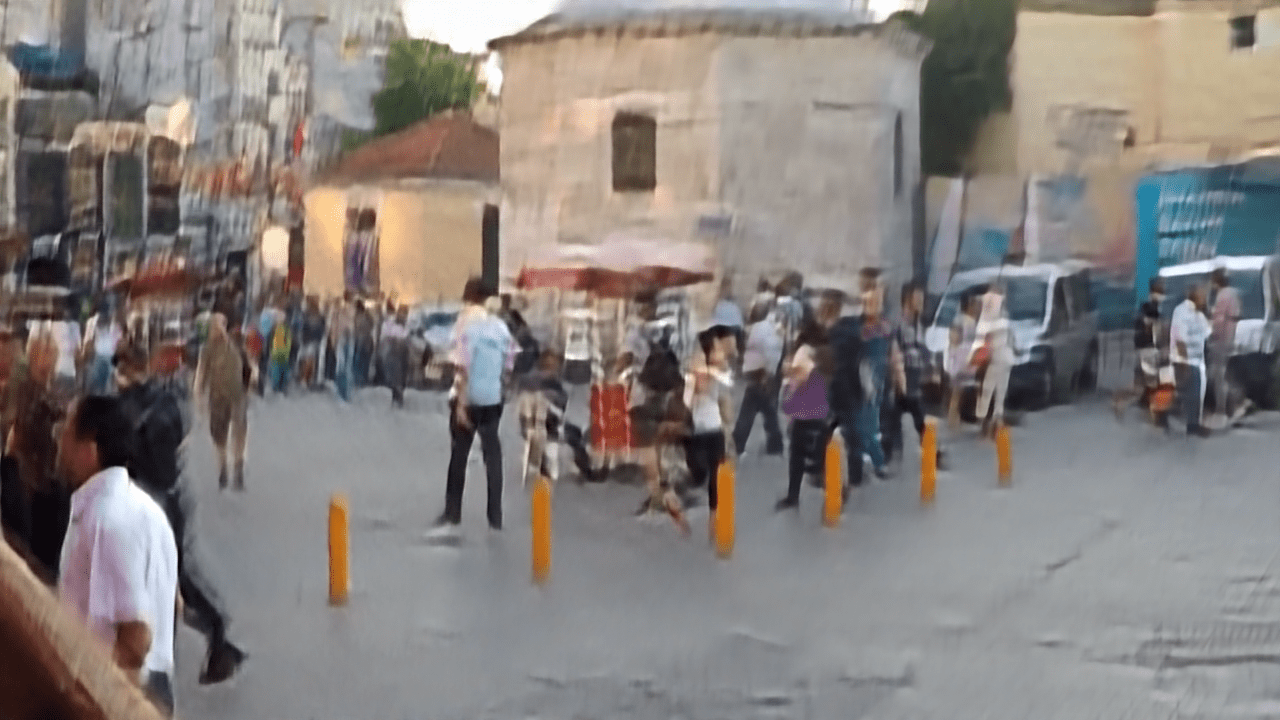}&
  \includegraphics[width=0.23\textwidth]{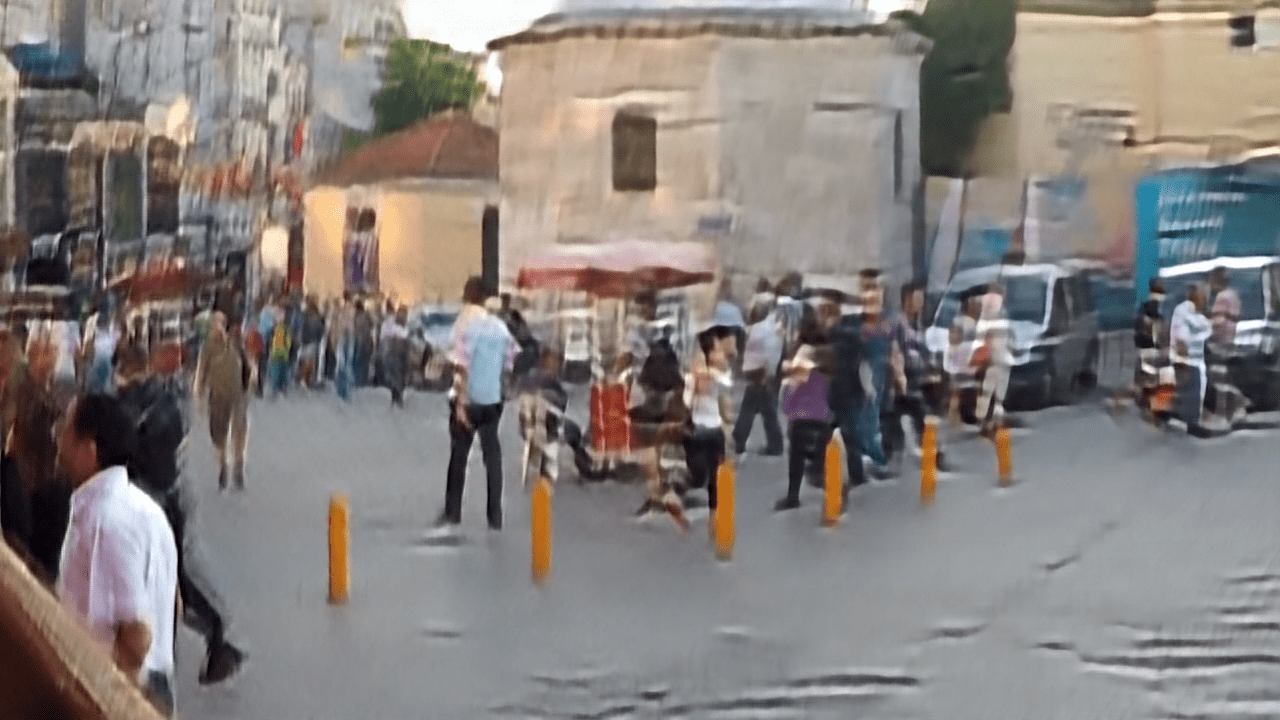}&
  \includegraphics[width=0.23\textwidth]{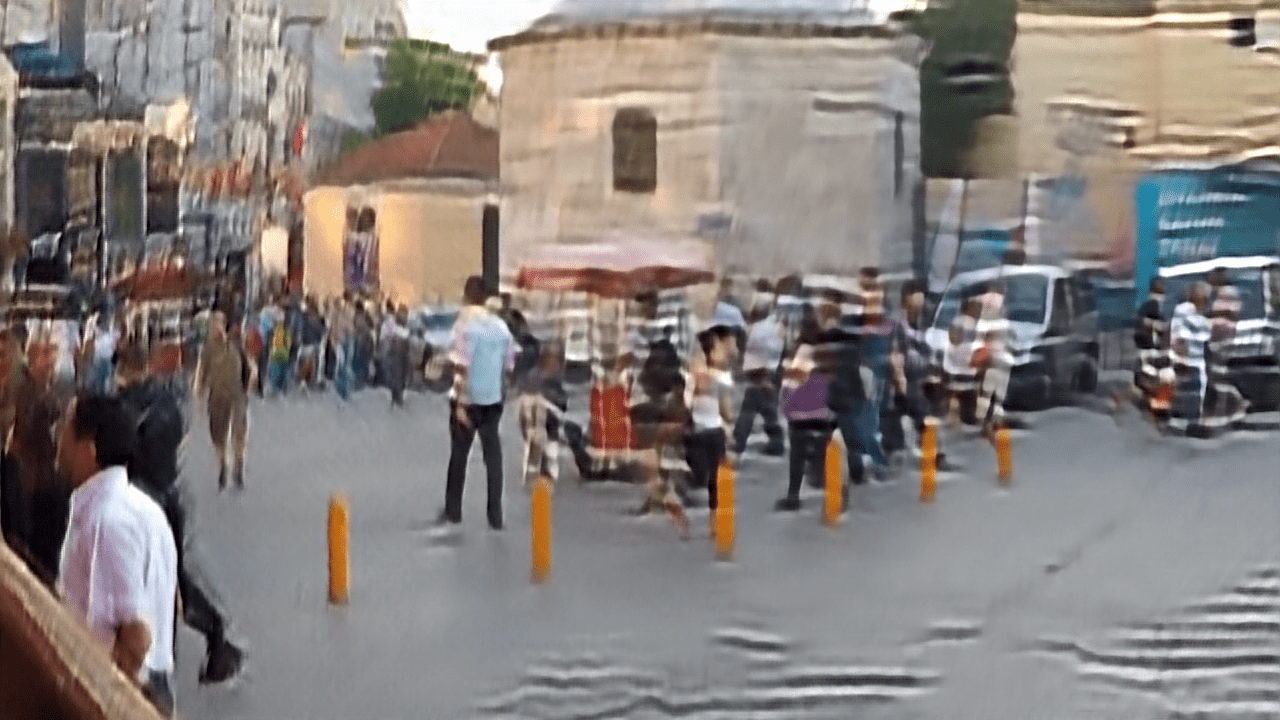}\\

  \rotatebox{90}{\textbf{Intermediate}} & \rotatebox{90}{~~\textbf{+ReLU}} & 
  \includegraphics[width=0.23\textwidth]{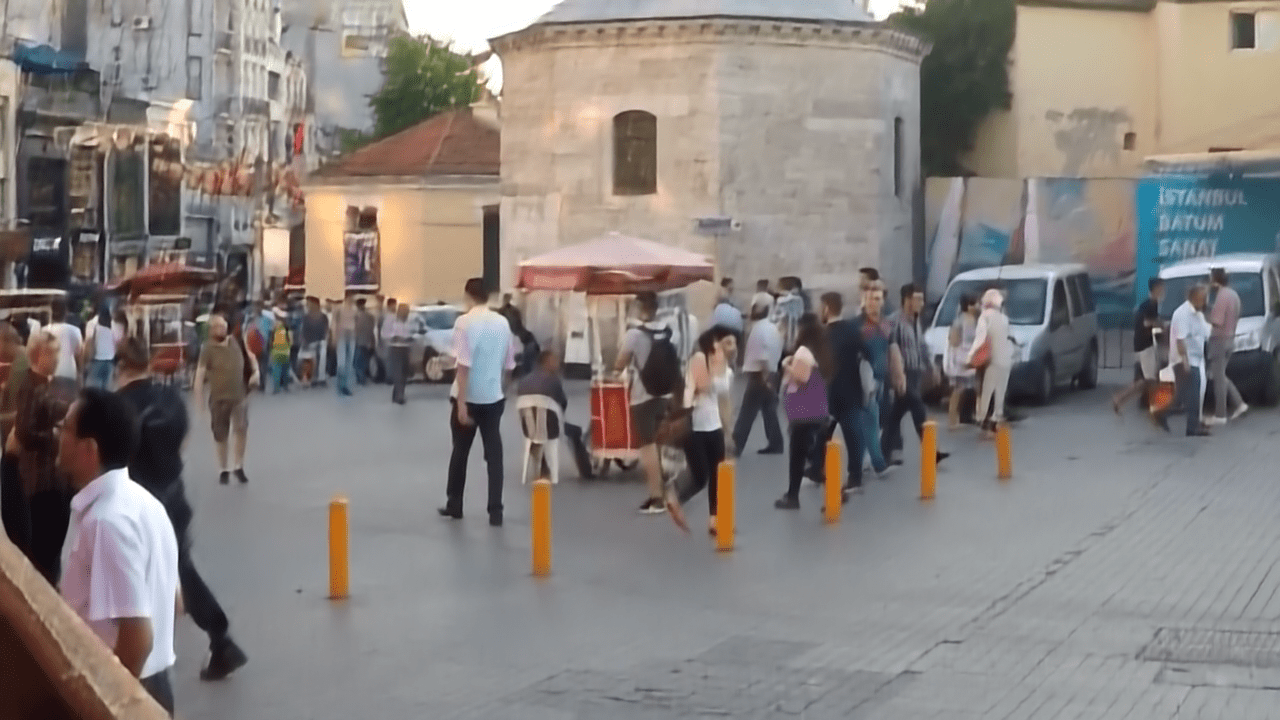}&
  \includegraphics[width=0.23\textwidth]{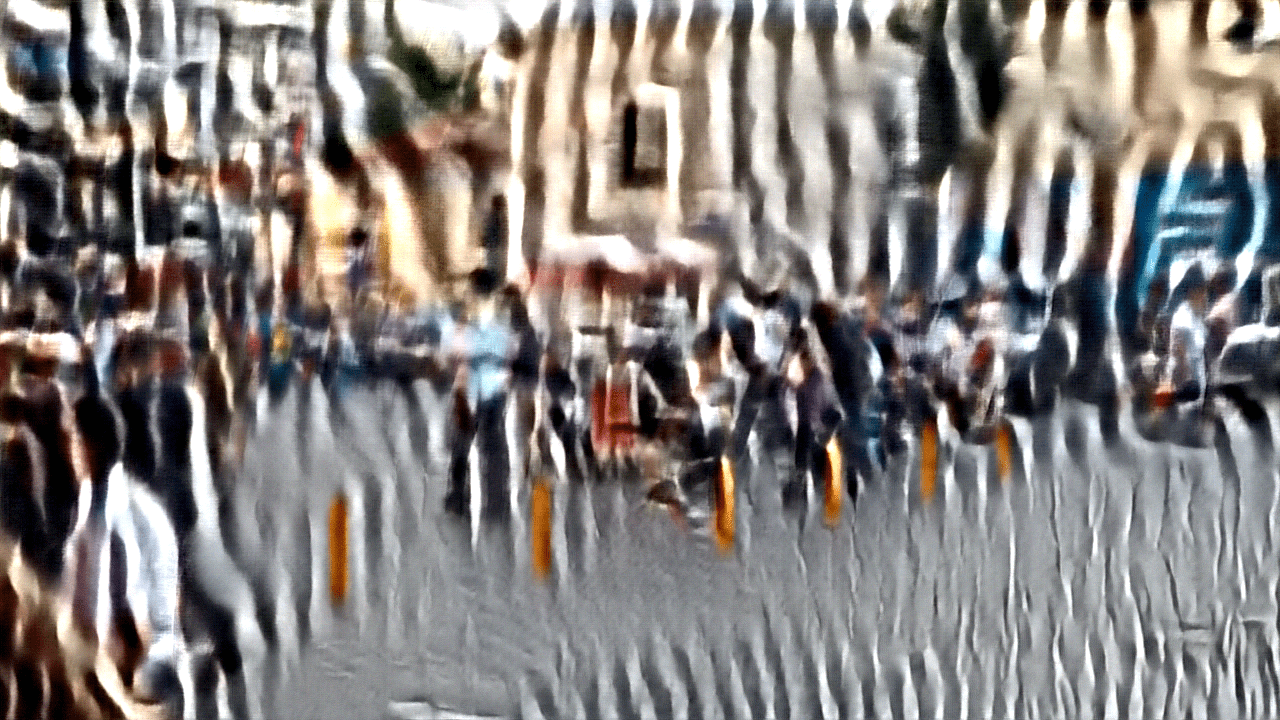}&
  \includegraphics[width=0.23\textwidth]{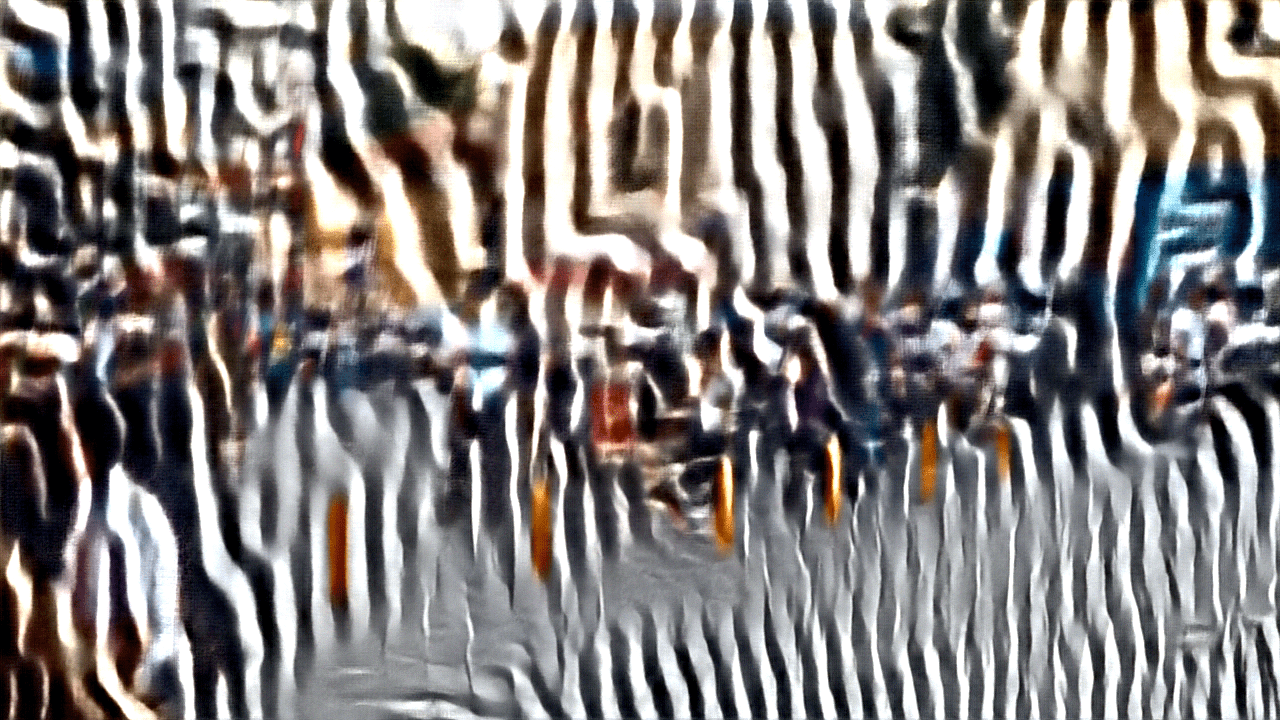}&
  \includegraphics[width=0.23\textwidth]{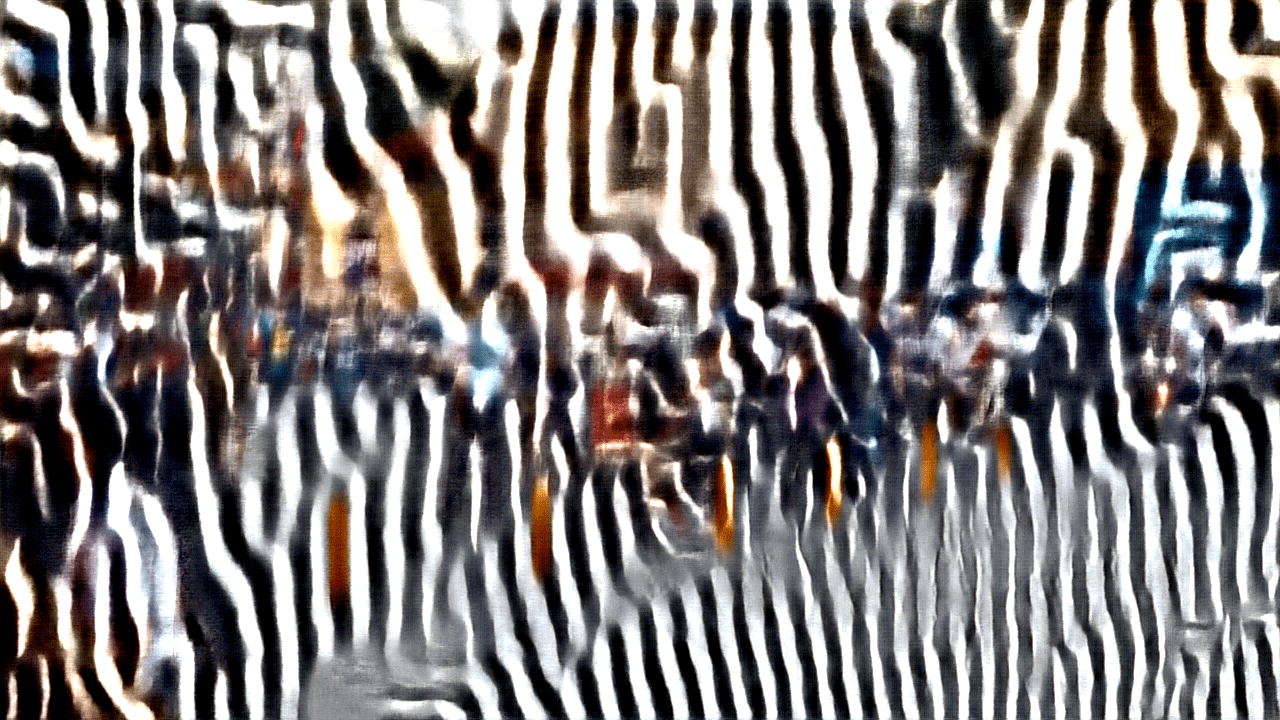}\\

  \rotatebox{90}{\textbf{Intermediate}} & \rotatebox{90}{~\textbf{+ReLU +ADV}} & 
  \includegraphics[width=0.23\textwidth]{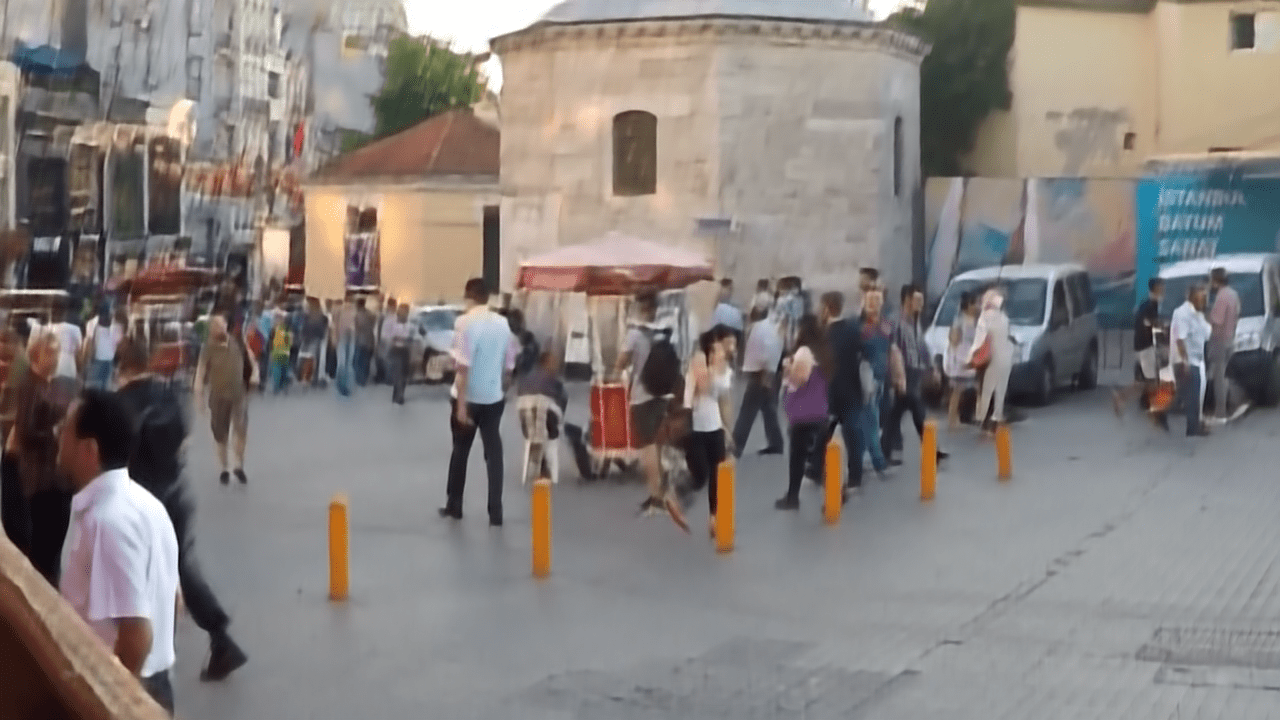}&
  \includegraphics[width=0.23\textwidth]{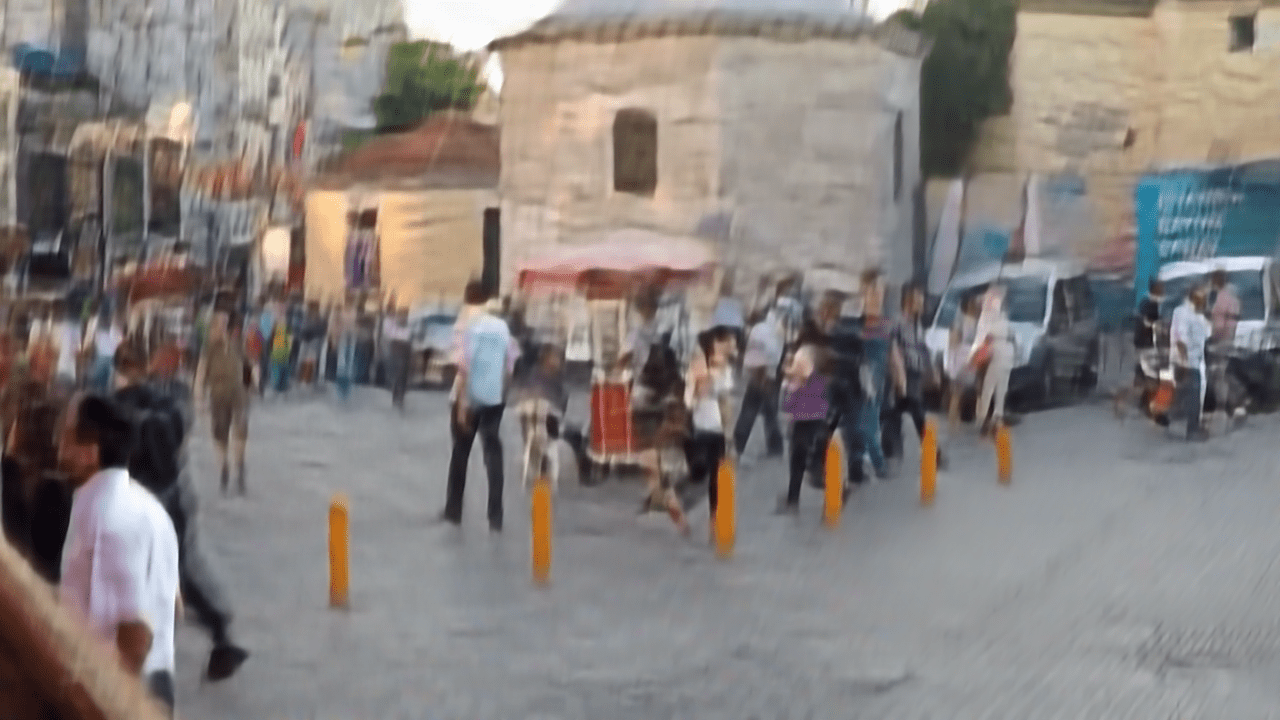}&
  \includegraphics[width=0.23\textwidth]{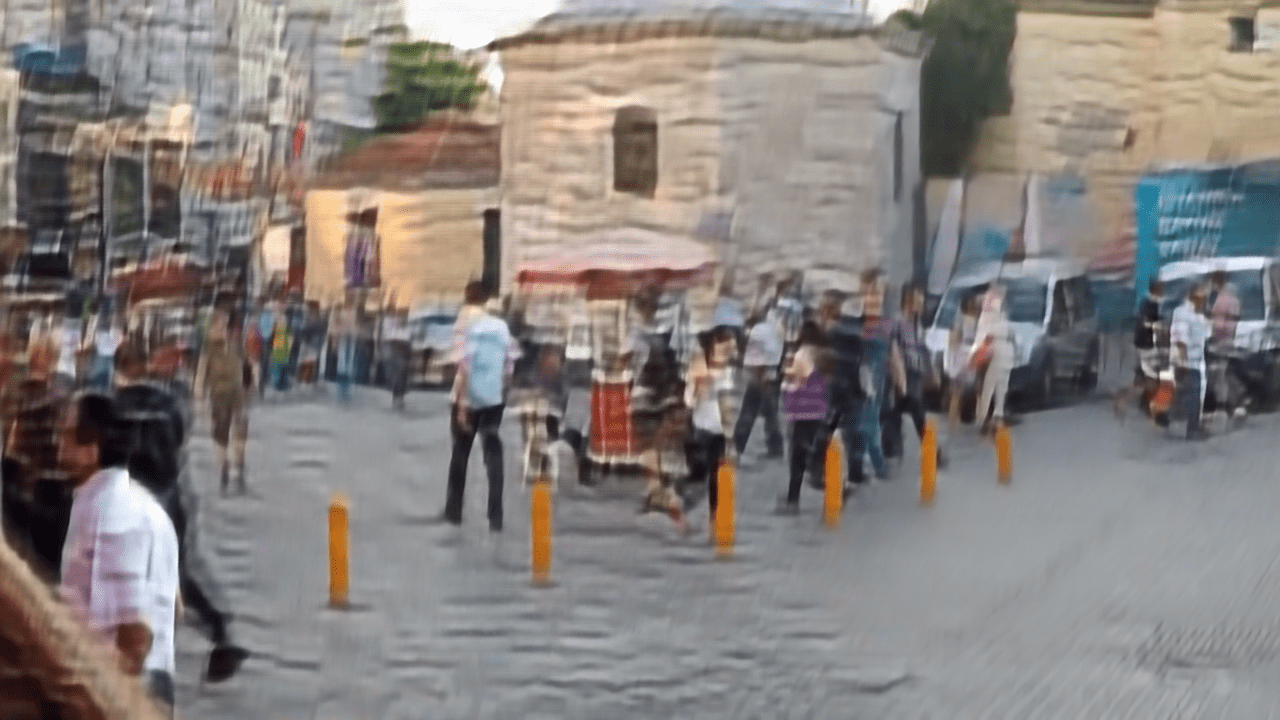}&
  \includegraphics[width=0.23\textwidth]{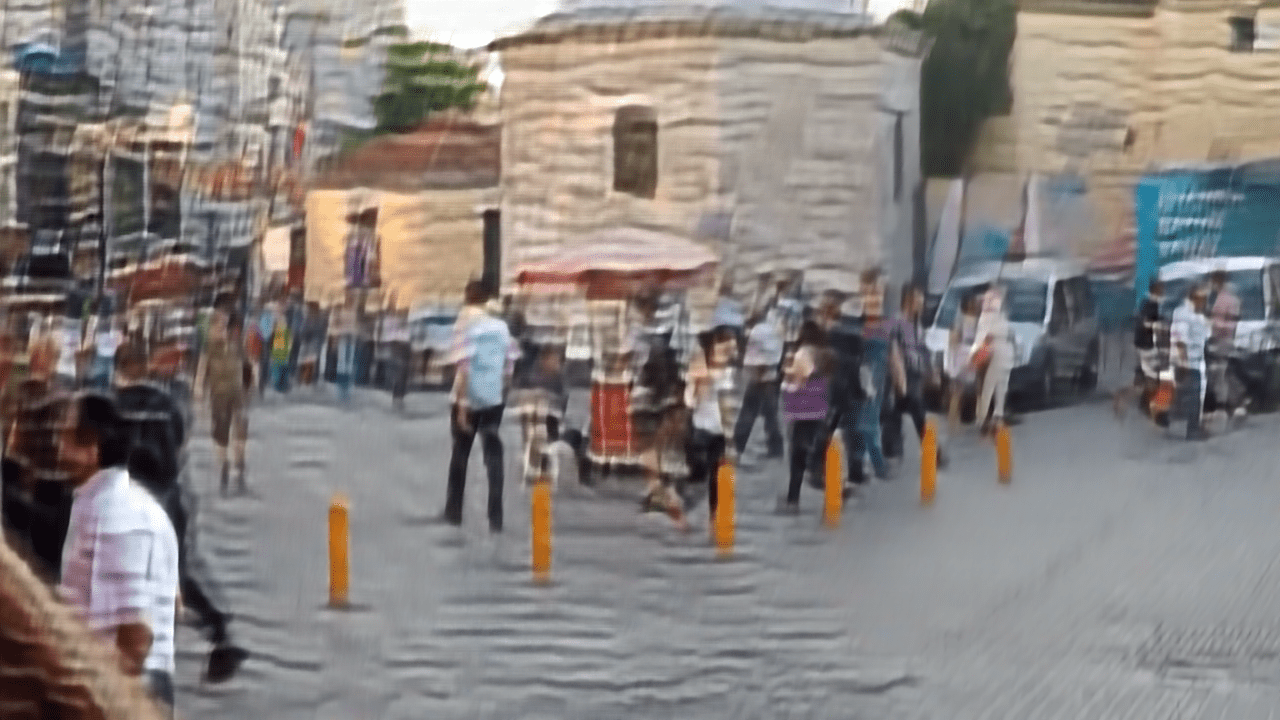}\\
  
  \rotatebox{90}{\textbf{NAFNet}} & & 
  \includegraphics[width=0.23\textwidth]{nafnet_no_attack_GOPR0384_11_00-000002.png}&
  \includegraphics[width=0.23\textwidth]{nafnet_cospgd_5_GOPR0384_11_00-000002.png}&
  \includegraphics[width=0.23\textwidth]{nafnet_cospgd_10_GOPR0384_11_00-000002.png}&
  \includegraphics[width=0.23\textwidth]{nafnet_cospgd_20_GOPR0384_11_00-000002.png}\\
  \rotatebox{90}{\textbf{Restormer}} & \rotatebox{90}{~~\textbf{+ADV}} & 
  \includegraphics[width=0.23\textwidth]{restormer_adv_no_attack_GOPR0384_11_00-000002.png}&
  \includegraphics[width=0.23\textwidth]{restormer_adv_cospgd_5_GOPR0384_11_00-000002.png}&
  \includegraphics[width=0.23\textwidth]{restormer_adv_cospgd_10_GOPR0384_11_00-000002.png}&
  \includegraphics[width=0.23\textwidth]{restormer_adv_cospgd_20_GOPR0384_11_00-000002.png}\\
  
  \rotatebox{90}{\textbf{Baseline}} & \rotatebox{90}{~~\textbf{+ADV}} & 
  \includegraphics[width=0.23\textwidth]{baseline_adv_no_attack_GOPR0384_11_00-000002.png}&
  \includegraphics[width=0.23\textwidth]{baseline_adv_cospgd_5_GOPR0384_11_00-000002.png}&
  \includegraphics[width=0.23\textwidth]{baseline_adv_cospgd_10_GOPR0384_11_00-000002.png}&
  \includegraphics[width=0.23\textwidth]{baseline_adv_cospgd_20_GOPR0384_11_00-000002.png}\\

  \rotatebox{90}{\textbf{NAFNet}} & \rotatebox{90}{~~\textbf{+ADV}} & 
  \includegraphics[width=0.23\textwidth]{nafnet_adv_no_attack_GOPR0384_11_00-000002.png}&
  \includegraphics[width=0.23\textwidth]{nafnet_adv_cospgd_5_GOPR0384_11_00-000002.png}&
  \includegraphics[width=0.23\textwidth]{nafnet_adv_cospgd_10_GOPR0384_11_00-000002.png}&
  \includegraphics[width=0.23\textwidth]{nafnet_adv_cospgd_20_GOPR0384_11_00-000002.png}\\

\end{tabular}
}
\caption{Comparing images reconstructed by all models after \textbf{CosPGD attack}}
\label{fig:cospgd_attack}
\end{figure*}

\begin{figure*}[htb]
    \centering 
\scalebox{0.92}{
    \begin{tabular}{@{}c@{\hspace{0.1cm}}c@{\hspace{0.1cm}}c@{\hspace{0.1cm}}c@{\hspace{0.1cm}}c@{\hspace{0.1cm}}c@{}}
    \multicolumn{2}{c}{MODEL} & NO ATTACK & 5 iterations & 10 iterations & 20 iterations\\
  \rotatebox{90}{\textbf{Restormer}} & & 
  \includegraphics[width=0.23\textwidth]{restormer_no_attack_GOPR0384_11_00-000002.png}&
  \includegraphics[width=0.23\textwidth]{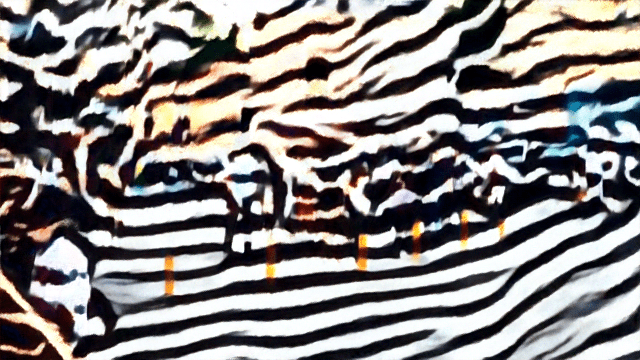}&
  \includegraphics[width=0.23\textwidth]{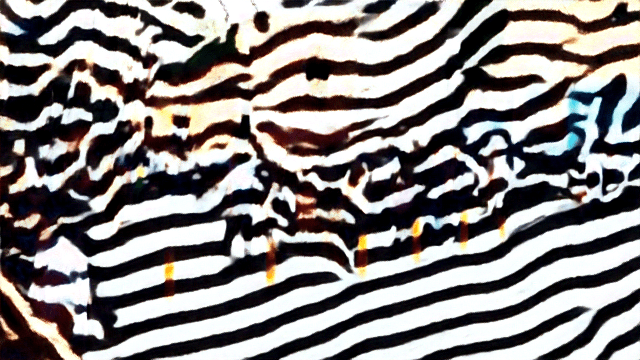}&
  \includegraphics[width=0.23\textwidth]{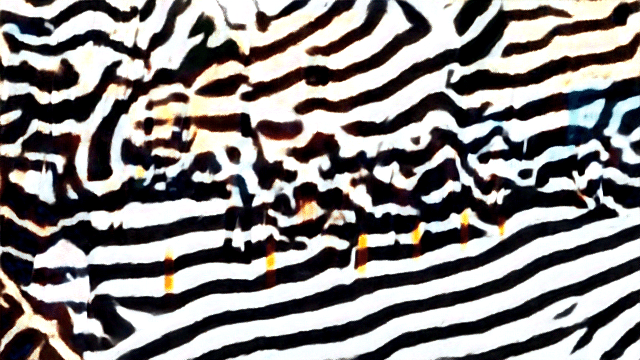}\\
  
  \rotatebox{90}{\textbf{Baseline}} & & 
  \includegraphics[width=0.23\textwidth]{baseline_no_attack_GOPR0384_11_00-000002.png}&
  \includegraphics[width=0.23\textwidth]{baseline_cospgd_5_GOPR0384_11_00-000002.png}&
  \includegraphics[width=0.23\textwidth]{baseline_cospgd_10_GOPR0384_11_00-000002.png}&
  \includegraphics[width=0.23\textwidth]{baseline_cospgd_20_GOPR0384_11_00-000002.png}\\

  \rotatebox{90}{\textbf{Intermediate}} & & 
  \includegraphics[width=0.23\textwidth]{intermediate_no_attack_GOPR0384_11_00-000002.png}&
  \includegraphics[width=0.23\textwidth]{intermediate_cospgd_5_GOPR0384_11_00-000002.png}&
  \includegraphics[width=0.23\textwidth]{intermediate_cospgd_10_GOPR0384_11_00-000002.png}&
  \includegraphics[width=0.23\textwidth]{intermediate_cospgd_20_GOPR0384_11_00-000002.png}\\

    \rotatebox{90}{\textbf{Intermediate}} & \rotatebox{90}{~~\textbf{+ADV}} & 
  \includegraphics[width=0.23\textwidth]{intermediate_gelu_adv_no_attack_GOPR0384_11_00-000002.png}&
  \includegraphics[width=0.23\textwidth]{intermediate_gelu_adv_cospgd_5_GOPR0384_11_00-000002.png}&
  \includegraphics[width=0.23\textwidth]{intermediate_gelu_adv_cospgd_10_GOPR0384_11_00-000002.png}&
  \includegraphics[width=0.23\textwidth]{intermediate_gelu_adv_cospgd_20_GOPR0384_11_00-000002.png}\\

  \rotatebox{90}{\textbf{Intermediate}} & \rotatebox{90}{~~\textbf{+ReLU}} & 
  \includegraphics[width=0.23\textwidth]{intermediate_relu_no_attack_GOPR0384_11_00-000002.png}&
  \includegraphics[width=0.23\textwidth]{intermediate_relu_cospgd_5_GOPR0384_11_00-000002.png}&
  \includegraphics[width=0.23\textwidth]{intermediate_relu_cospgd_10_GOPR0384_11_00-000002.png}&
  \includegraphics[width=0.23\textwidth]{intermediate_relu_cospgd_20_GOPR0384_11_00-000002.png}\\

    \rotatebox{90}{\textbf{Intermediate}} & \rotatebox{90}{~\textbf{+ReLU +ADV}} & 
  \includegraphics[width=0.23\textwidth]{intermediate_relu_adv_no_attack_GOPR0384_11_00-000002.png}&
  \includegraphics[width=0.23\textwidth]{intermediate_relu_adv_cospgd_5_GOPR0384_11_00-000002.png}&
  \includegraphics[width=0.23\textwidth]{intermediate_relu_adv_cospgd_10_GOPR0384_11_00-000002.png}&
  \includegraphics[width=0.23\textwidth]{intermediate_relu_adv_cospgd_20_GOPR0384_11_00-000002.png}\\
  
  \rotatebox{90}{\textbf{NAFNet}} & & 
  \includegraphics[width=0.23\textwidth]{nafnet_no_attack_GOPR0384_11_00-000002.png}&
  \includegraphics[width=0.23\textwidth]{nafnet_cospgd_5_GOPR0384_11_00-000002.png}&
  \includegraphics[width=0.23\textwidth]{nafnet_cospgd_10_GOPR0384_11_00-000002.png}&
  \includegraphics[width=0.23\textwidth]{nafnet_cospgd_20_GOPR0384_11_00-000002.png}\\

  \rotatebox{90}{\textbf{Restormer}} & \rotatebox{90}{~~\textbf{+ADV}} & 
  \includegraphics[width=0.23\textwidth]{restormer_adv_no_attack_GOPR0384_11_00-000002.png}&
  \includegraphics[width=0.23\textwidth]{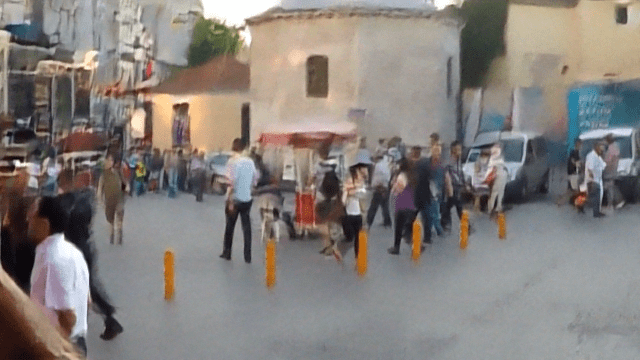}&
  \includegraphics[width=0.23\textwidth]{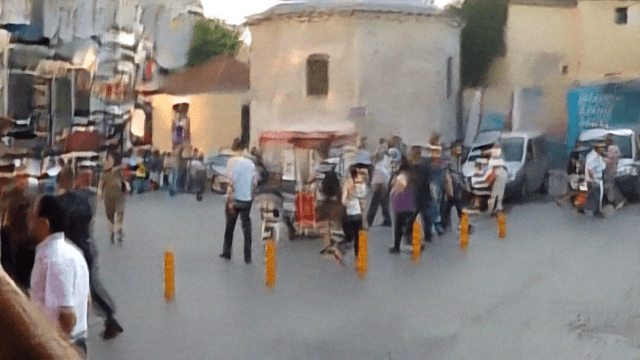}&
  \includegraphics[width=0.23\textwidth]{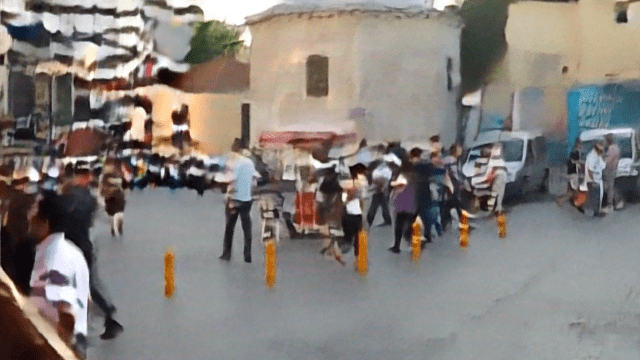}\\
  
  \rotatebox{90}{\textbf{Baseline}} & \rotatebox{90}{~~\textbf{+ADV}} & 
  \includegraphics[width=0.23\textwidth]{baseline_adv_no_attack_GOPR0384_11_00-000002.png}&
  \includegraphics[width=0.23\textwidth]{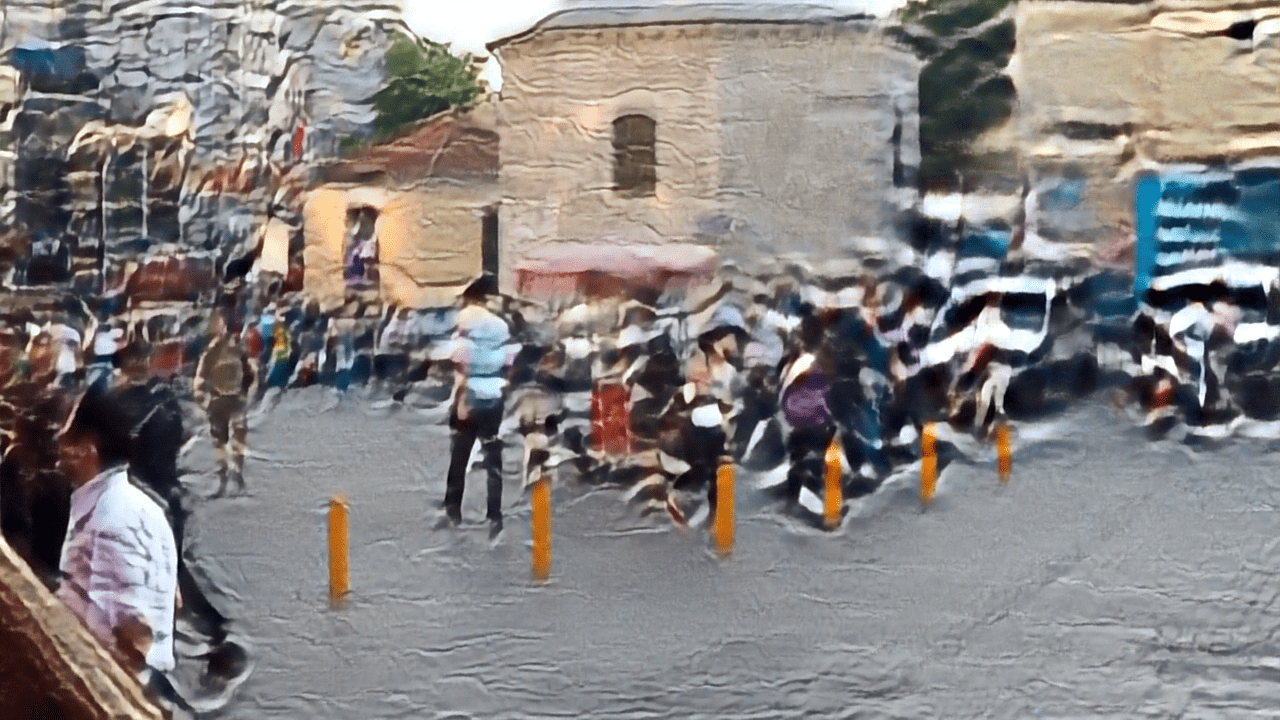}&
  \includegraphics[width=0.23\textwidth]{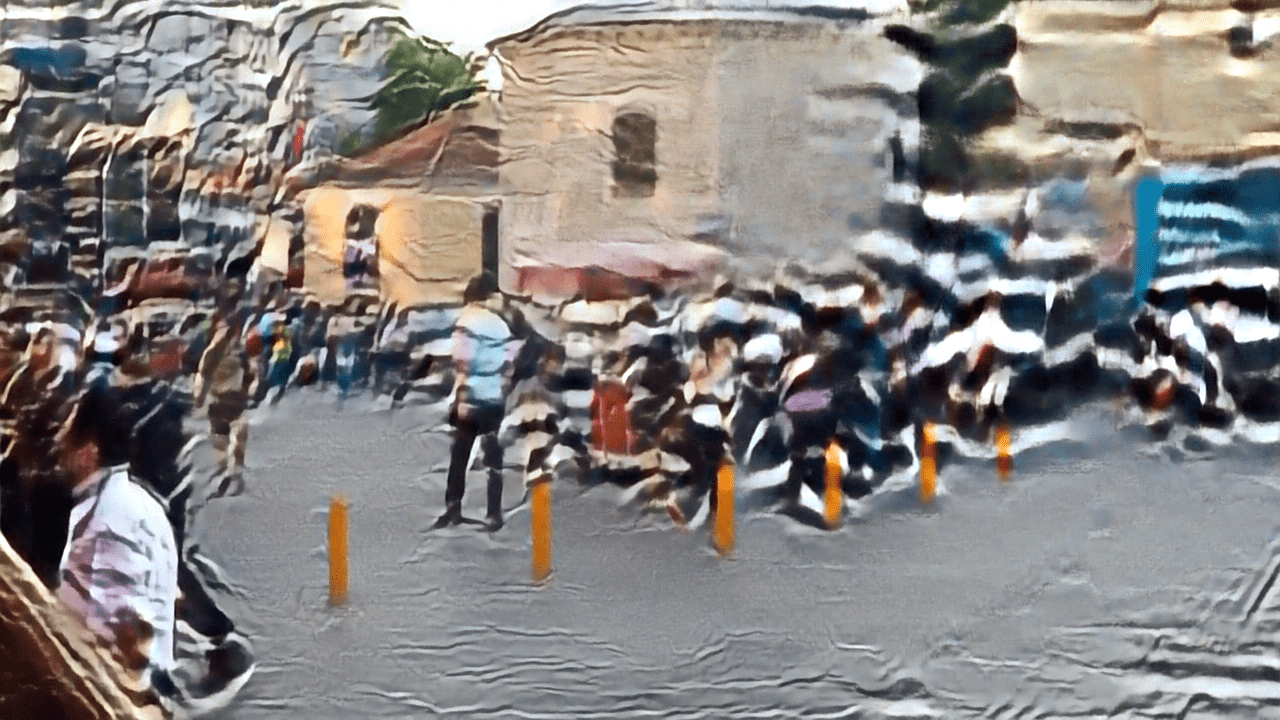}&
  \includegraphics[width=0.23\textwidth]{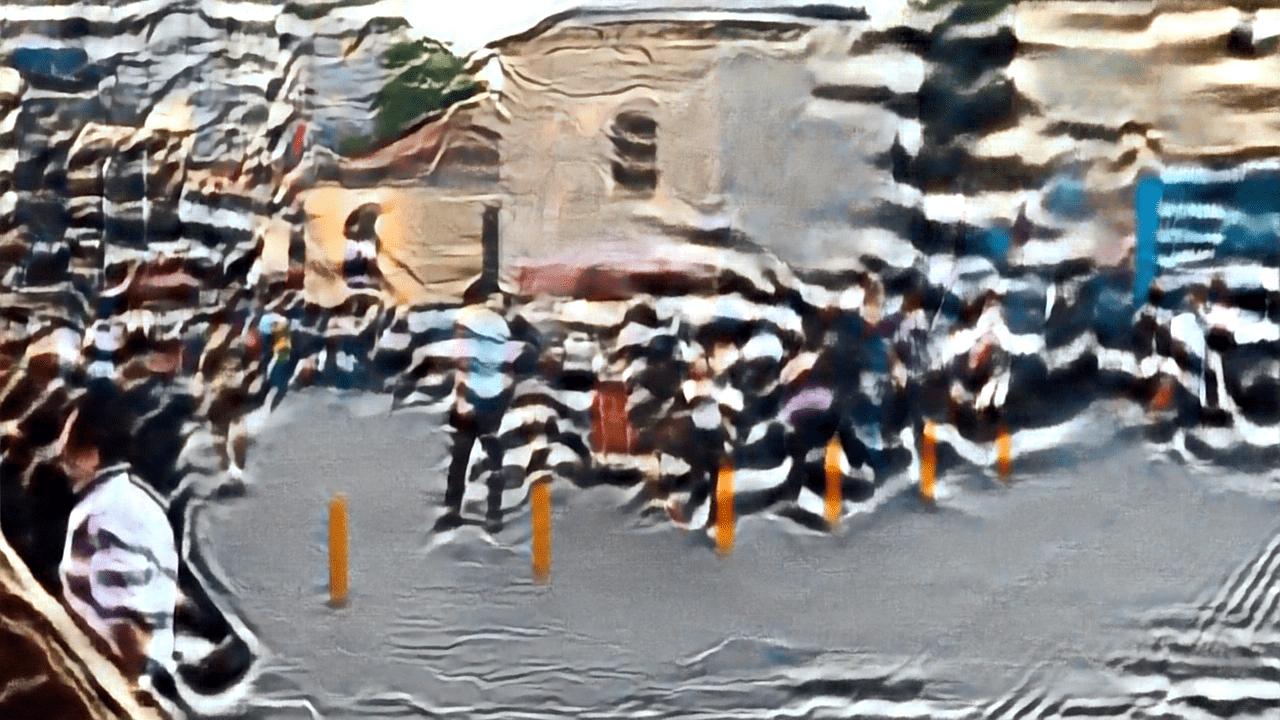}\\

  \rotatebox{90}{\textbf{NAFNet}} & \rotatebox{90}{~~\textbf{+ADV}} & 
  \includegraphics[width=0.23\textwidth]{nafnet_adv_no_attack_GOPR0384_11_00-000002.png}&
  \includegraphics[width=0.23\textwidth]{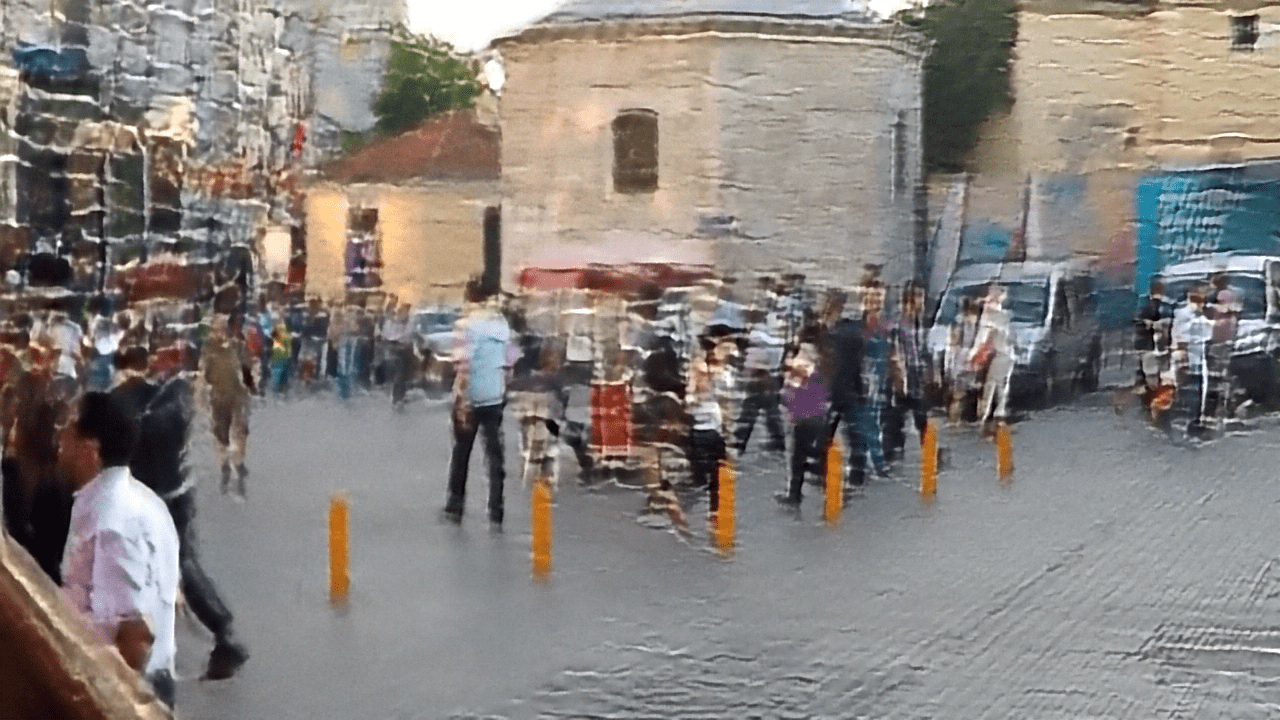}&
  \includegraphics[width=0.23\textwidth]{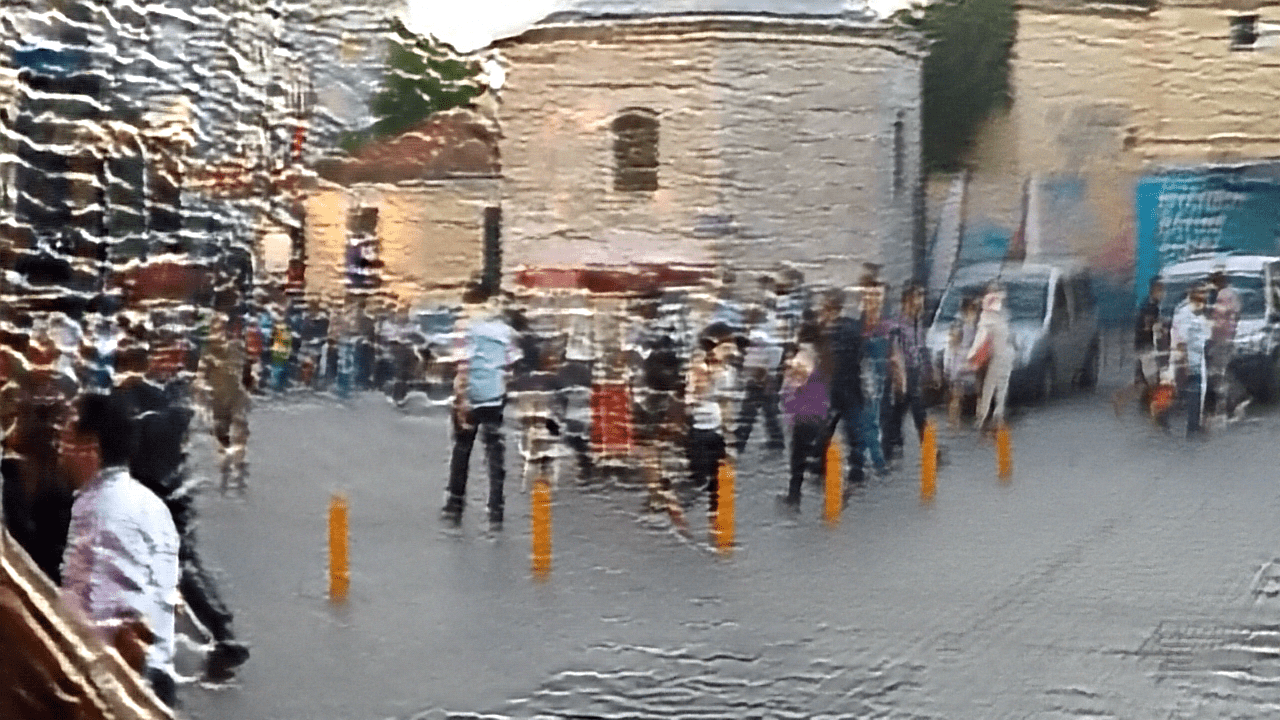}&
  \includegraphics[width=0.23\textwidth]{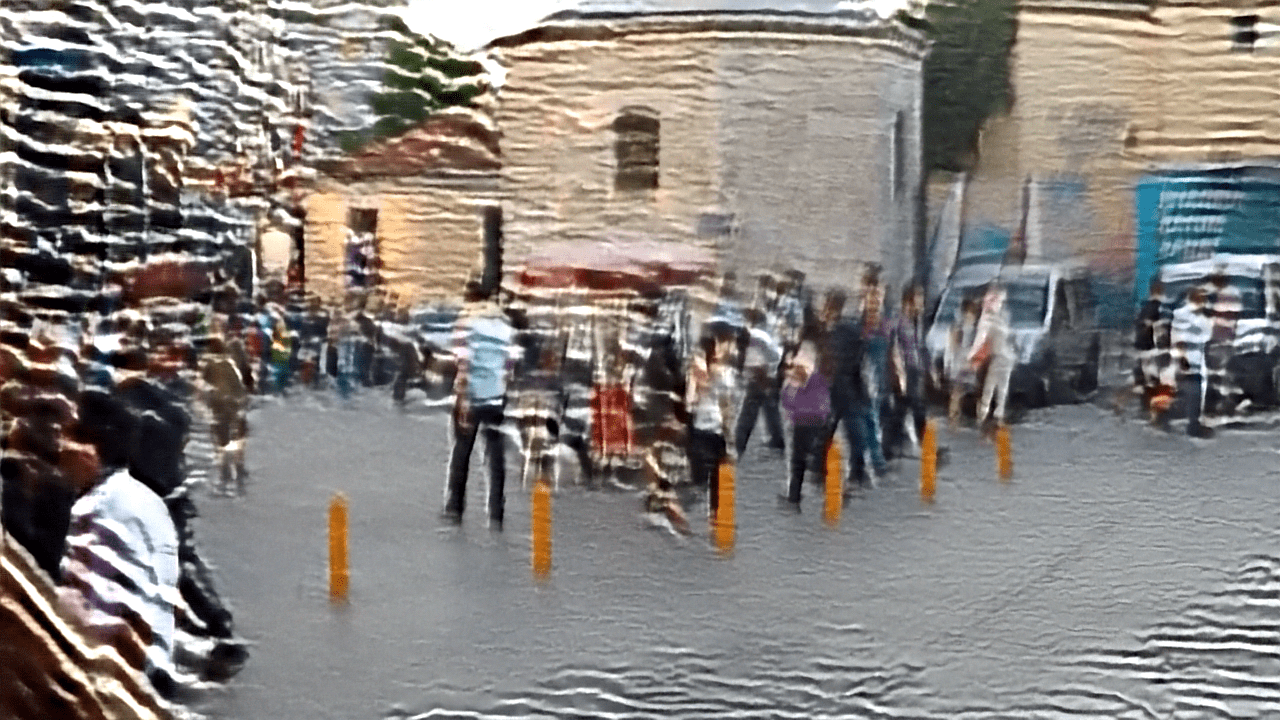}\\
  
\end{tabular}
}
\caption{Comparing images reconstructed by all models after \textbf{PGD attack}}
\label{fig:pgd_attack}
\end{figure*}

We provide sample reconstructed images from all considered networks under adversarial attacks.
Figure~\ref{fig:cospgd_attack} shows reconstructed images from GoPro test dataset~\cite{gopro} after the CosPGD attack~\cite{agnihotri2023cospgd} on the models.
Whereas Figure.~\ref{fig:pgd_attack} shows reconstructed images from GoPro test dataset~\cite{gopro} after the PGD attack~\cite{pgd} on the models.

\subsection{Intermediate networks}
In Table~\ref{tab:intermediate} we report the performance of the \emph{Intermediate network} and \emph{Intermediate + ReLU}.
Please note, the performance of the Intermediate network on the clean (unperturbed) samples is marginally lower than that reported by \cite{chen2022simple}.
As \cite{chen2022simple} does not provide the code, pre-trained weights, or training configuration for this intermediate step between the Baseline network and NAFNet, our implementation is limited to the best of our understanding.

\end{multicols}

\begin{table*}[h]    
\centering
    \caption{Comparison of performance of the considered \emph{Intermediate network} on clean test images and against CosPGD and PGD attacks with various attack strengths.
    The adversarial attacks were performed at $\epsilon\approx\frac{8}{255}$.}
    \scalebox{0.75}{
    
    \begin{tabular}{@{}l|cc|cc|cc|cc|cc|cc|cc@{}}
    \toprule
    \multirow{3}{*}{Architecture} & \multicolumn{2}{c|}{Clean} & \multicolumn{6}{c|}{CosPGD} & \multicolumn{6}{c}{PGD} \\
    
    & \multirow{2}{*}{PSNR} & \multirow{2}{*}{SSIM} & \multicolumn{2}{c|}{5 attack itrs } & \multicolumn{2}{c|}{10 attack itrs } & \multicolumn{2}{c|}{20 attack itrs } & \multicolumn{2}{c|}{5 attack itrs } & \multicolumn{2}{c|}{10 attack itrs } & \multicolumn{2}{c}{20 attack itrs } \\
    & & & PSNR & SSIM & PSNR & SSIM & PSNR & SSIM & PSNR & SSIM & PSNR & SSIM & PSNR & SSIM \\
    \toprule
         \textbf{Intermediate} & 29.93 & 0.9289 & 6.0224 & 0.0509 & 5.8166 & 0.0366 & 5.7199 & 0.0315 & 6.0225 & 0.0509 & 5.8158 & 0.0365 & 5.7173 & 0.0314 \\
         \textbf{~~~~ + ADV} & 29.00  & 0.9154 & 24.02 & 0.8213 & 22.01 & 0.7775 & 20.15 & 0.7286 & 24.02 & 0.8213 & 21.98 & 0.7770 & 20.15 & 0.7286 \\
         \midrule
         \textbf{Intermediate + ReLU} & 30.39 & 0.9349 & 13.87 & 0.4093 & 11.63 & 0.3128 & 10.29 & 0.2538 & 13.87 & 0.4094 & 11.62 & 0.3127 & 10.29 & 0.2542 \\   
           
        \textbf{~~~~ + ADV} & 28.49 & 0.9072 & 23.90 & 0.8046 & 22.46 & 0.7637 & 21.85 & 0.7484 & 23.91 & 0.8046 & 22.47 & 0.7638 & 21.84 & 0.7481 \\ 
    \bottomrule
    \end{tabular}    
    }
    \label{tab:intermediate}
\end{table*}

\end{document}